%% file: main.tex
\documentclass[runningheads]{llncs}

\usepackage{eccv}

\usepackage{eccvabbrv}

\usepackage{graphicx}
\usepackage{booktabs}
\usepackage{array}
\usepackage{multirow}
\usepackage{amsmath}	
\usepackage{amssymb}	

\usepackage{caption}
\usepackage{xcolor}
\usepackage{float}
\usepackage{placeins}
\usepackage{color, colortbl}
\usepackage{stfloats}
\usepackage{enumitem}
\usepackage{tabularx}
\usepackage{xstring}
\usepackage{multirow}
\usepackage{xspace}
\usepackage{url}
\usepackage{subcaption}
\usepackage{xcolor}
\usepackage[hang,flushmargin]{footmisc}
\usepackage{bbm}
\usepackage{listings}
\usepackage{longtable}
\usepackage[accsupp]{axessibility}  %

\newcommand{\R}[1]{{%
    \textbf{%
        \ifstrequal{#1}{1}{\textcolor{red}{R#1}}{%
        \ifstrequal{#1}{2}{\textcolor{blue}{R#1}}{%
        \ifstrequal{#1}{3}{\textcolor{magenta}{R#1}}{%
        \ifstrequal{#1}{4}{\textcolor{teal}{R#1}}{%
                           \textcolor{cyan}{R#1}%
        }}}}%
    }%
}}

\usepackage{xr-hyper}

\makeatletter
\newcommand*{\addFileDependency}[1]{
  \typeout{(#1)}
  \@addtofilelist{#1}
  \IfFileExists{#1}{}{\typeout{No file #1.}}
}

\makeatother

\newlength\savewidth\newcommand\shline{\noalign{\global\savewidth\arrayrulewidth
		\global\arrayrulewidth 1pt}\hline\noalign{\global\arrayrulewidth\savewidth}}
\newcommand{\tablestyle}[2]{\setlength{\tabcolsep}{#1}\renewcommand{\arraystretch}{#2}\centering\footnotesize}
\renewcommand{\paragraph}[1]{\vspace{1.25mm}\noindent\textbf{#1}}

\usepackage{xspace}
\makeatletter
\DeclareRobustCommand\onedot{\futurelet\@let@token\@onedot}
\def\@onedot{\ifx\@let@token.\else.\null\fi\xspace}
\def\eg{\emph{e.g}\onedot} 
\def\ie{\emph{i.e}\onedot} 
 
 \def\vs{\emph{vs}\onedot}

\g@addto@macro{\endtabular}{\rowfont{}}%
\makeatother
\newcommand{\rowfonttype}{}%
\newcommand{\rowfont}[1]{%
\gdef\rowfonttype{#1}#1\ignorespaces%
}

\definecolor{hr}{gray}{0.7}  %
\definecolor{dt}{HTML}{ADCAD8}  %

\newcolumntype{*}{>{\global\let\currentrowstyle\relax}}
\newcolumntype{^}{>{\currentrowstyle}}
\newcommand{\rowstyle}[1]{\gdef\currentrowstyle{#1}#1\ignorespaces}
\newcolumntype{H}{>{\setbox0=\hbox\bgroup}c<{\egroup}@{}}
\newcolumntype{Z}{>{\setbox0=\hbox\bgroup}c<{\egroup}@{\hspace*{-\tabcolsep}}}

\makeatother

\usepackage{pifont}

\definecolor{deemph}{gray}{0.6}

\definecolor{baselinecolor}{gray}{.9}

\usepackage[pagebackref,breaklinks,colorlinks,citecolor=eccvblue]{hyperref}

\usepackage{orcidlink}

\begin{document}

\title{A Semantic Space is Worth 256 Language Descriptions: 
Make Stronger Segmentation Models with Descriptive Properties}

\titlerunning{ProLab: Property-level Label Space}

\def\authorBlock{
    Junfei Xiao\textsuperscript{1} \quad
    Ziqi Zhou \textsuperscript{2}\quad
    Wenxuan Li\textsuperscript{1} \quad
    Shiyi Lan\textsuperscript{3} \quad 
    Jieru Mei\textsuperscript{1} \quad   \\ 
    Zhiding Yu\textsuperscript{3} \quad
    Bingchen Zhao\textsuperscript{4} \quad 
   Alan Yuille\textsuperscript{1} \quad Yuyin Zhou\textsuperscript{2} \quad Cihang Xie\textsuperscript{2}
}

\institute{\textsuperscript{1}Johns Hopkins University \quad 
   \textsuperscript{2}UCSC \quad \textsuperscript{3}NVIDIA \quad \textsuperscript{4}University of Edinburgh   \\}
\author{\authorBlock}

\authorrunning{Xiao et al.}

\maketitle

\input{00_abstract}

\input{01_intro}

\input{02_related}

\input{03_method}

\input{04_experiment}

\input{10_conclusion}

\section*{Acknowledgments}
This work was supported by ONR grant N00014-23-1-2641. We thank Chen Wei, Zongwei Zhou, and Yaoyao Liu for their valuable discussions.
\bibliographystyle{splncs04}
\bibliography{main}

\clearpage

\appendix
\section*{\centering{Appendix}}
This appendix contains more implementation details (\S\ref{appendix:impl}), evaluation with creative generated images (\S\ref{appendix:eval_ood_gen}), the descriptive properties retrieved from GPT-3.5 (\S\ref{appendix:properties}), and a set of category pairs with ``similar" properties (\S\ref{appendix:similar_categories}).

\section{Implementation Details}
\label{appendix:impl}
We use Pytorch as the deep learning framework and our code is built upon MMSegmentation~\cite{contributors2020mmsegmentation}. Details are provided below: optimizer and hyperparameters (\S\ref{appendix:optim_and_hyperparam}), training data pipeline (\S\ref{appendix:training_pipeline}), and testing data pipeline (\S\ref{appendix:testing_pipeline}).

\subsection{Optimizer and Hyper-parameters}
\label{appendix:optim_and_hyperparam}

\Cref{tab:imple_details} provides detailed information about the optimizer and hyperparameter settings. 

\begin{table}[ht]
\tablestyle{6pt}{1.02}
\small
\vspace{-5mm}
\begin{tabular}{cc}
Config & Setting \\
\shline
Optimizer & AdamW \cite{loshchilov2018decoupled} \\
Learning rate & 6e-5 \\
Weight decay & 0.01 \\
Optimizer momentum & $\beta_1, \beta_2{=}0.9, 0.999$ \\
Batch size & 16 \\
Learning rate schedule & Poly \cite{chen2017deeplab}\\
Warmup iters \cite{goyal2017accurate} & 1500 \\
\end{tabular}
\caption{\textbf{Optimizer \& hyper-parameters settings.}}
\vspace{-10mm}
\label{tab:imple_details}
\end{table}

\vspace{-5mm}
\subsection{Training Data Pipelines}
\label{appendix:training_pipeline}

We follow the training pipeline used in \cite{chen2023vision}. \Cref{tab:train_pip_natural} and \Cref{tab:train_pip_driving} provide the detailed data processing pipelines for training. A warmup training stage with one-hot category-level labels is performed when training large models (\ie, ViT-L). It is performed at the first 40K iterations, which leads to better performance.

\begin{table}[h]
\tablestyle{10pt}{1.1}
\small
\vspace{-5mm}
\begin{tabular}{c|c}
 Operation & Setting \\ 
\shline
Resize &  Scale: (2048, 512), Ratio: (0.5, 2.0) \\
RandomCrop &  Crop size: (512, 512) \\
RandomFlip & Prob: 0.5 \\
PhotoMetricDistortion & Default
\end{tabular}
\caption{\textbf{Training data pipeline for ADE20K, COCO-Stuff, and PascalContext.}}
\vspace{-10mm}
\label{tab:train_pip_natural} 
\end{table}

\begin{table}[h]
\tablestyle{8pt}{1.05}
\small
\vspace{-5mm}
\begin{tabular}{c|c}
 Operation & Setting \\ 
\shline
Resize &  Scale: (2048, 1024), Ratio: (0.5, 2.0) \\
RandomCrop &  Crop size: (768, 768) \\
RandomFlip & Prob: 0.5 \\
PhotoMetricDistortion & Default
\end{tabular}
\caption{\textbf{Training data pipeline for Cityscapes and BDD.}}
\vspace{-5mm}
\label{tab:train_pip_driving} 
\end{table}

\subsection{Testing Data Pipelines}
\label{appendix:testing_pipeline}
Sliding window strategy is used in testing. \Cref{tab:test_pip_natural} and \Cref{tab:test_pip_driving} provide the testing data pipelines.

\begin{table}[ht]
\tablestyle{16pt}{1.2}
\small
\begin{tabular}{c|c}
 Operation & Setting \\ 
\shline
 Crop Size & (512, 512) \\
Sliding Stride & (341, 341) \\
 Random     Flip & True
\end{tabular}
\caption{\textbf{Testing data pipeline for ADE20K, COCO-Stuff, and PascalContext.}}
\label{tab:test_pip_natural} 
\vspace{-10mm}
\end{table}

\begin{table}[ht]
\tablestyle{16pt}{1.2}
\small
\vspace{-10mm}
\begin{tabular}{c|c}
 Operation & Setting \\ 
\shline
 Crop Size & (768, 768) \\
Sliding Stride & (512, 512) \\
 Random     Flip & True
\end{tabular}
\caption{\textbf{Testing data pipeline for Cityscapes and BDD.}}
\label{tab:test_pip_driving} 
\vspace{-4mm}
\end{table}

\vspace{-10mm}
\section{Evaluation with Out-of-domain  Generations}
\label{appendix:eval_ood_gen}
\begin{figure*}[b]
\centering

\begin{subfigure}[b]{0.48\linewidth}
\includegraphics[width=\linewidth]{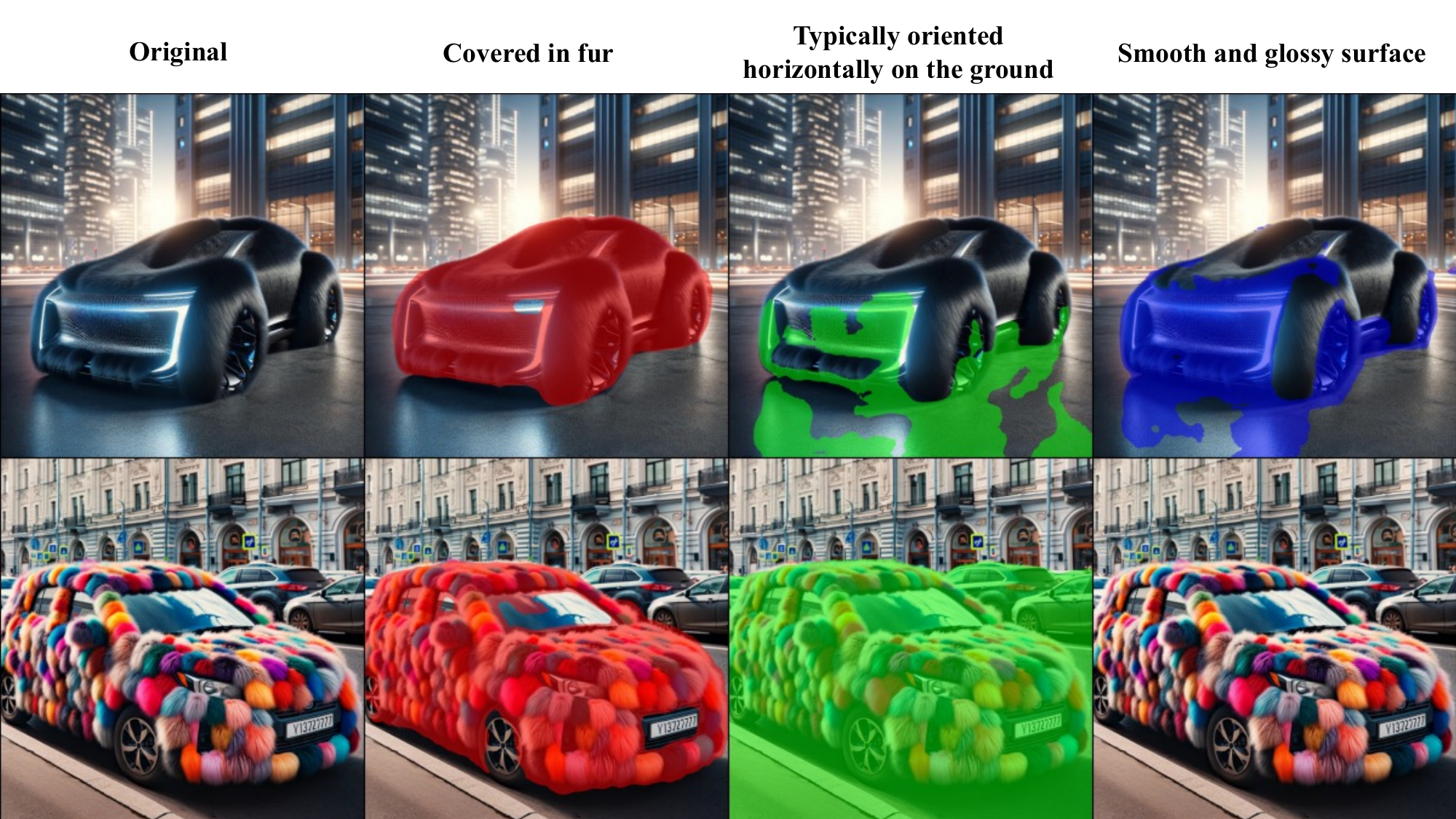}
\caption{\textbf{A car covered in fur}}
\label{fig:vis_1}
\end{subfigure}
\hfill
\begin{subfigure}[b]{0.48\linewidth}
\includegraphics[width=\linewidth]{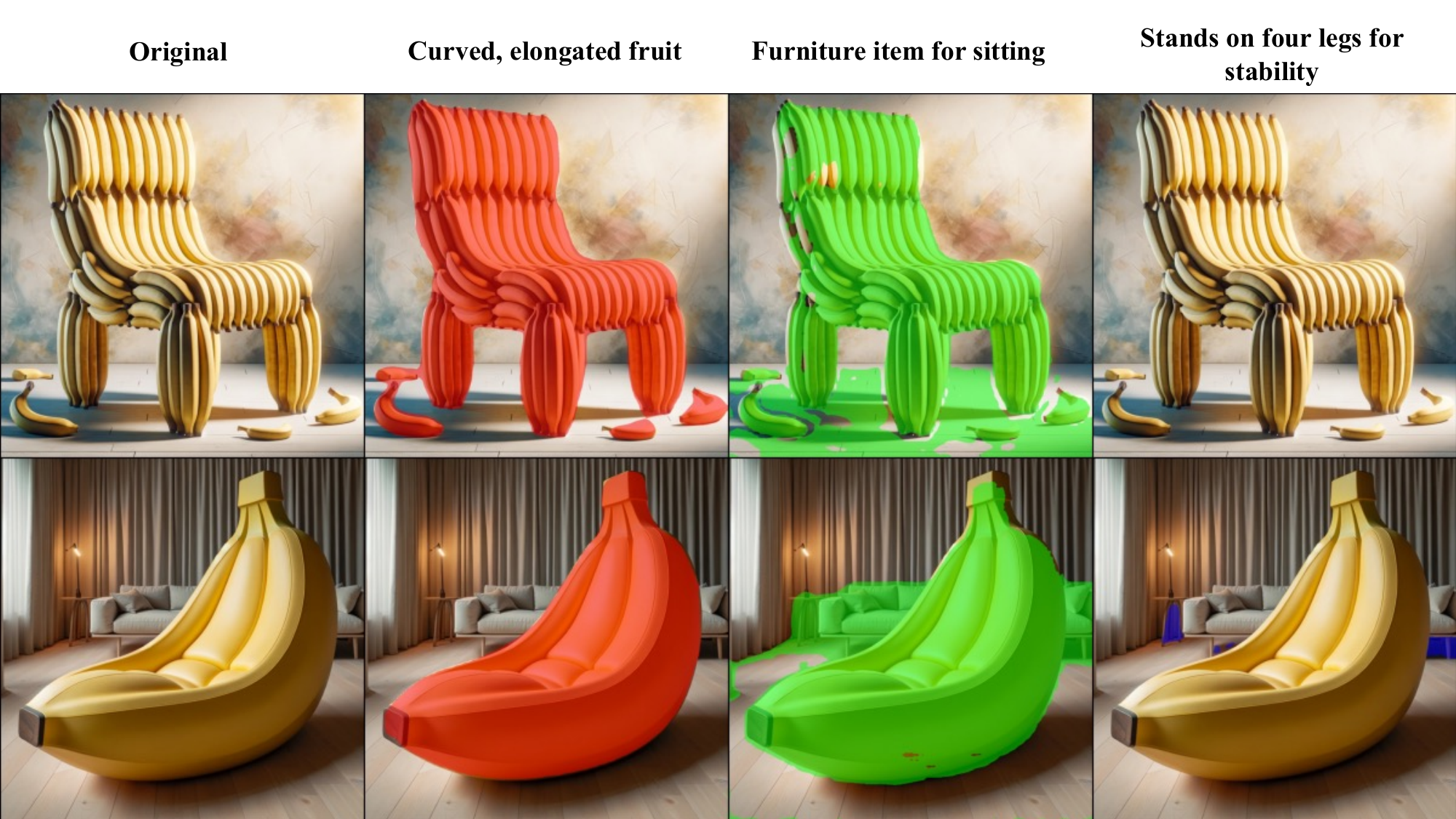}
\caption{\textbf{A banana chair}}
\label{fig:vis_2}
\end{subfigure}

\begin{subfigure}[b]{0.48\linewidth}
\includegraphics[width=\linewidth]{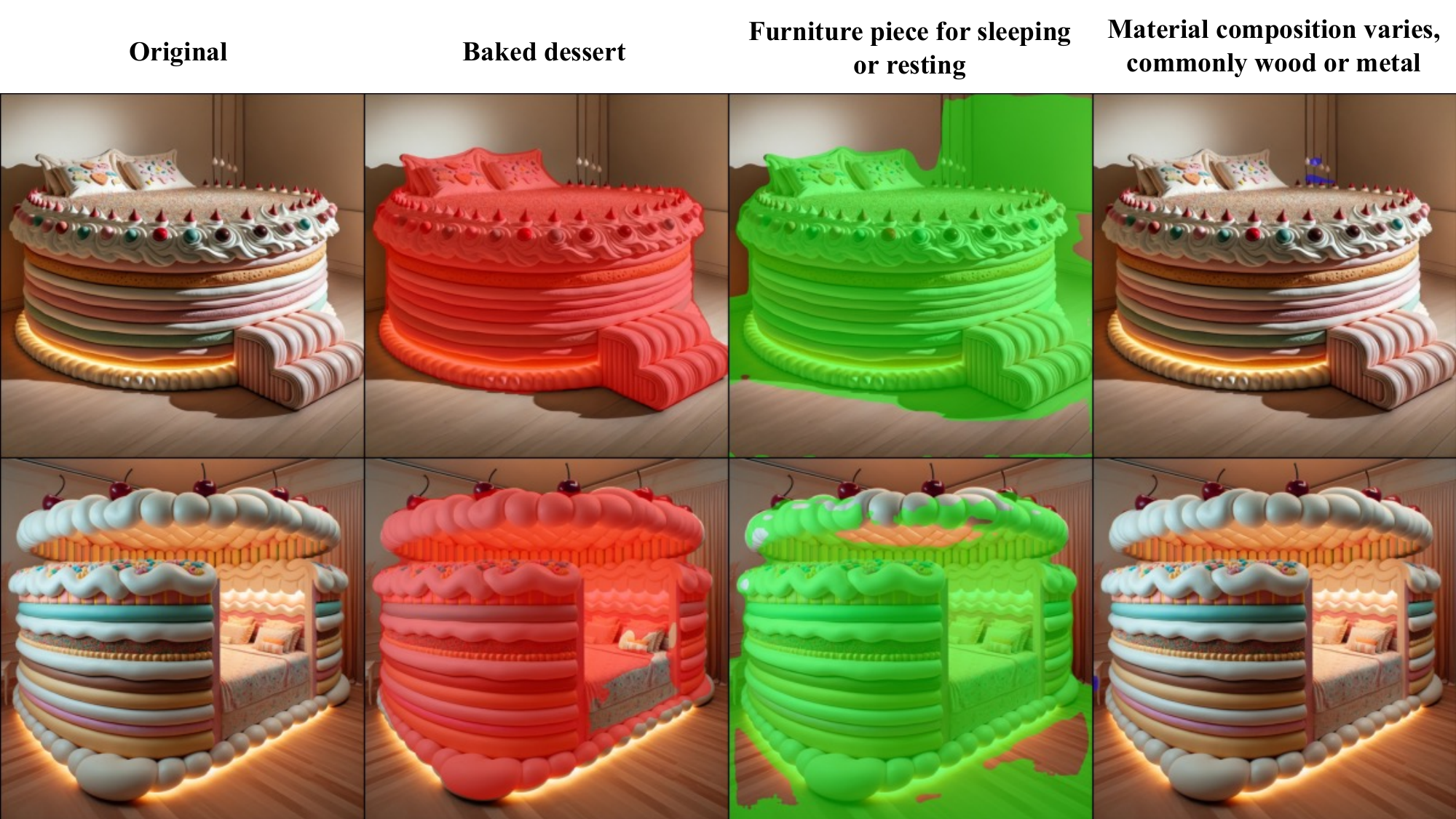}
\caption{\textbf{A cake bed}}
\label{fig:vis_3}
\end{subfigure}
\hfill
\begin{subfigure}[b]{0.48\linewidth}
\includegraphics[width=\linewidth]{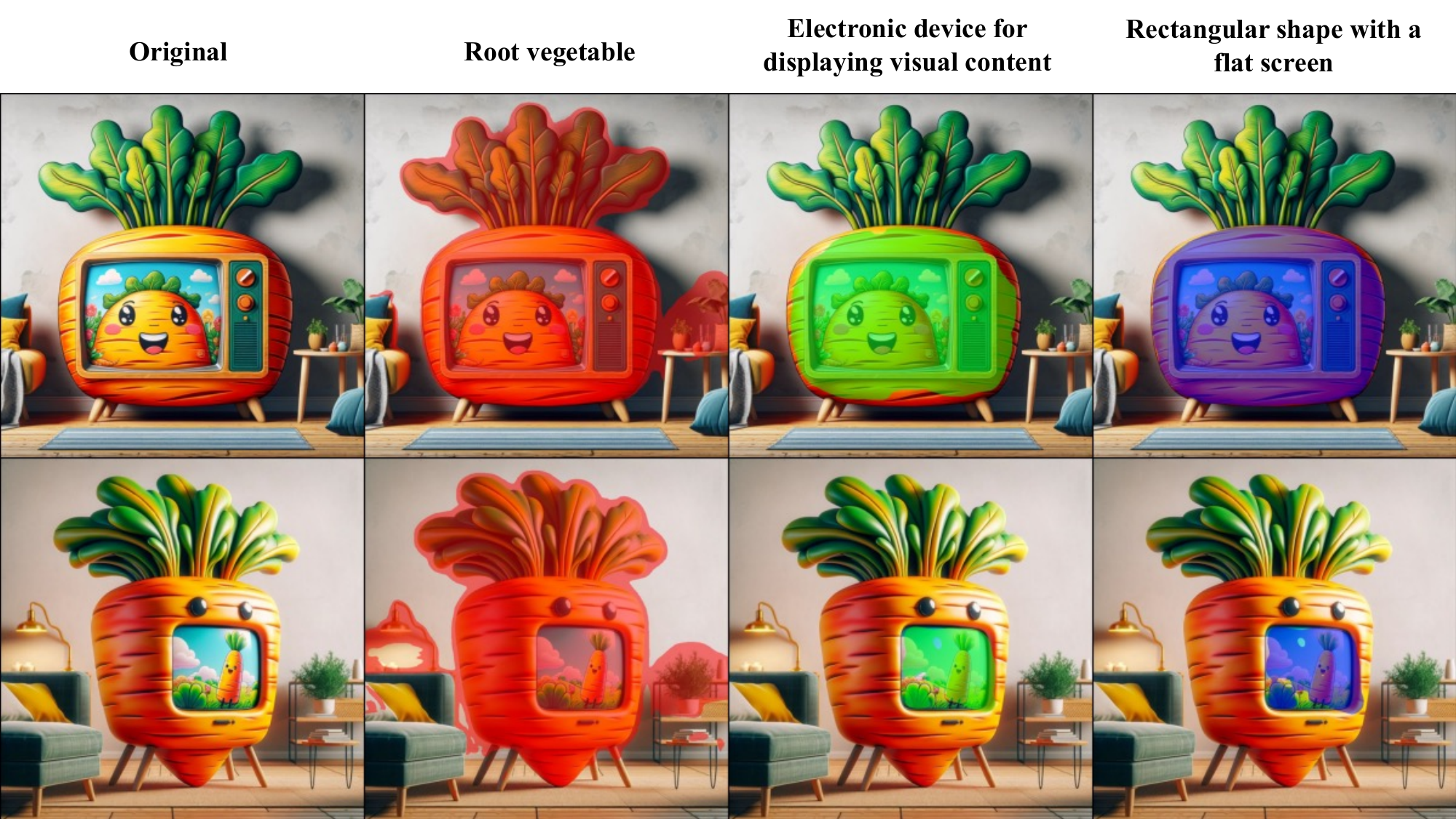}
\caption{\textbf{A carrot television}}
\label{fig:vis_4}
\end{subfigure}
\hfill

\begin{subfigure}[b]{0.48\linewidth}
\includegraphics[width=\linewidth]{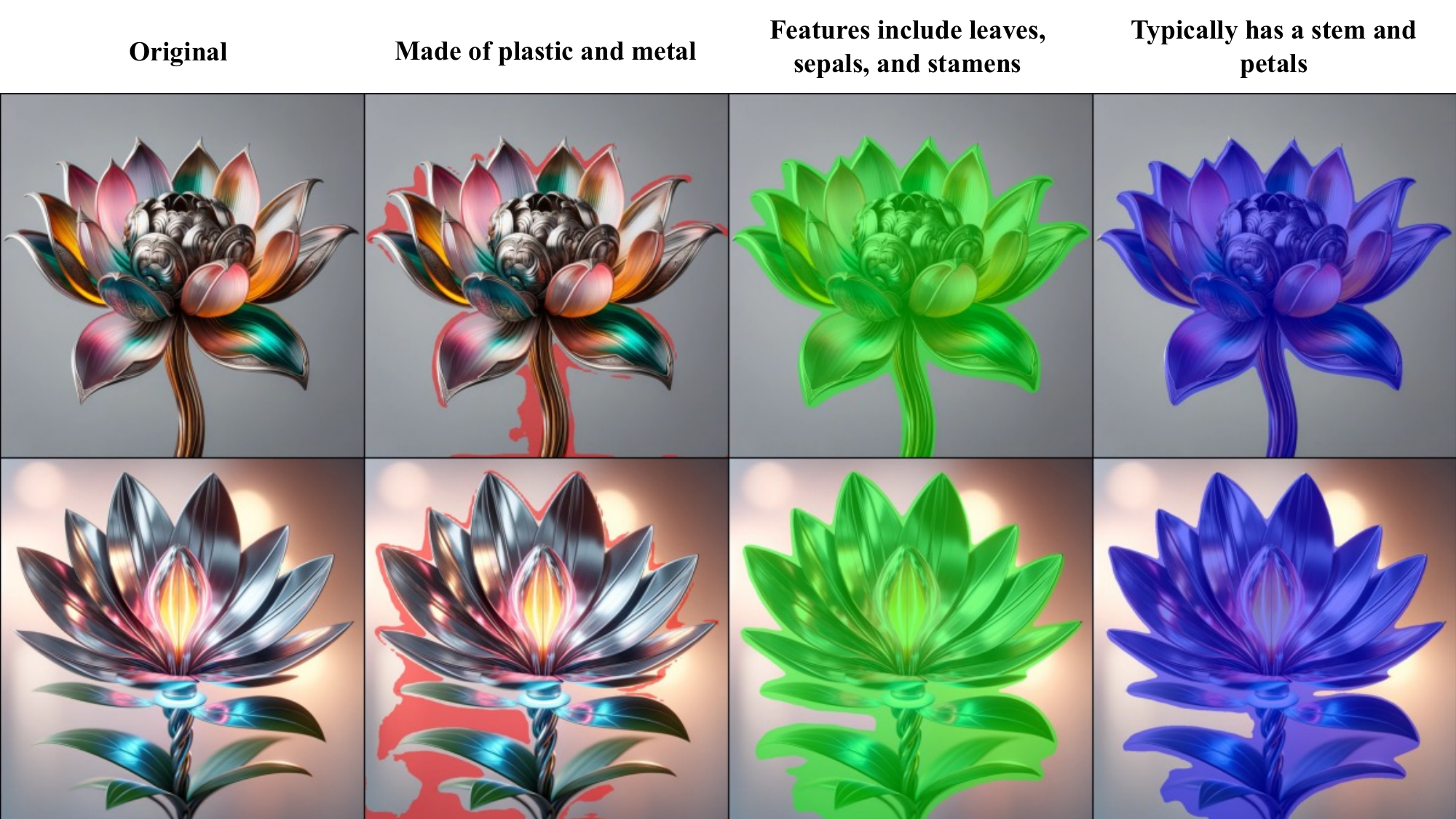}
\caption{\textbf{A flower made of plastic and metal}}
\label{fig:vis_5}
\end{subfigure}
\hfill
\begin{subfigure}[b]{0.48\linewidth}
\includegraphics[width=\linewidth]{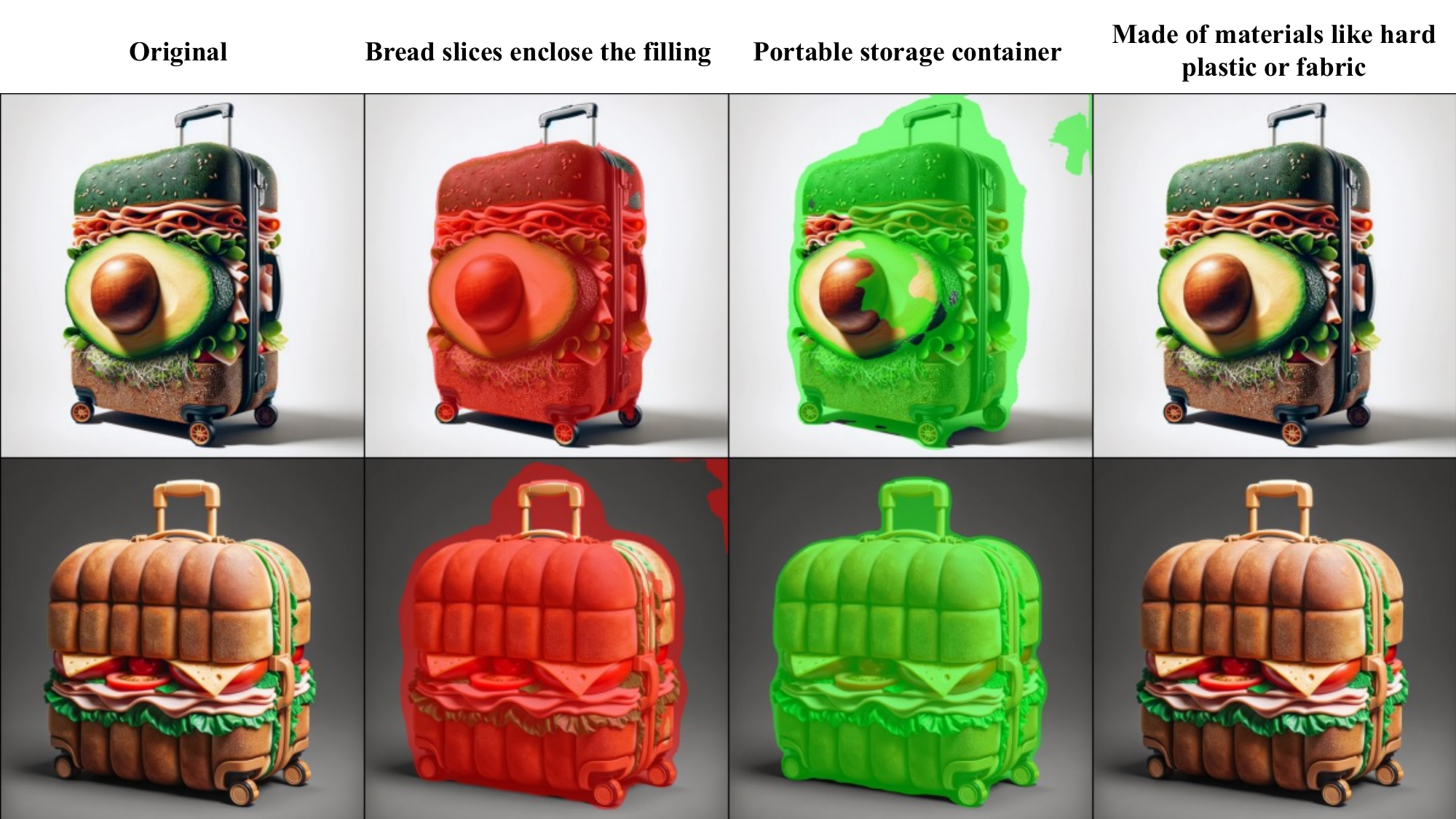}
\caption{\textbf{A sandwich suitcase}}
\label{fig:vis_6}
\end{subfigure}
\hfill

\begin{subfigure}[b]{0.48\linewidth}
\includegraphics[width=\linewidth]{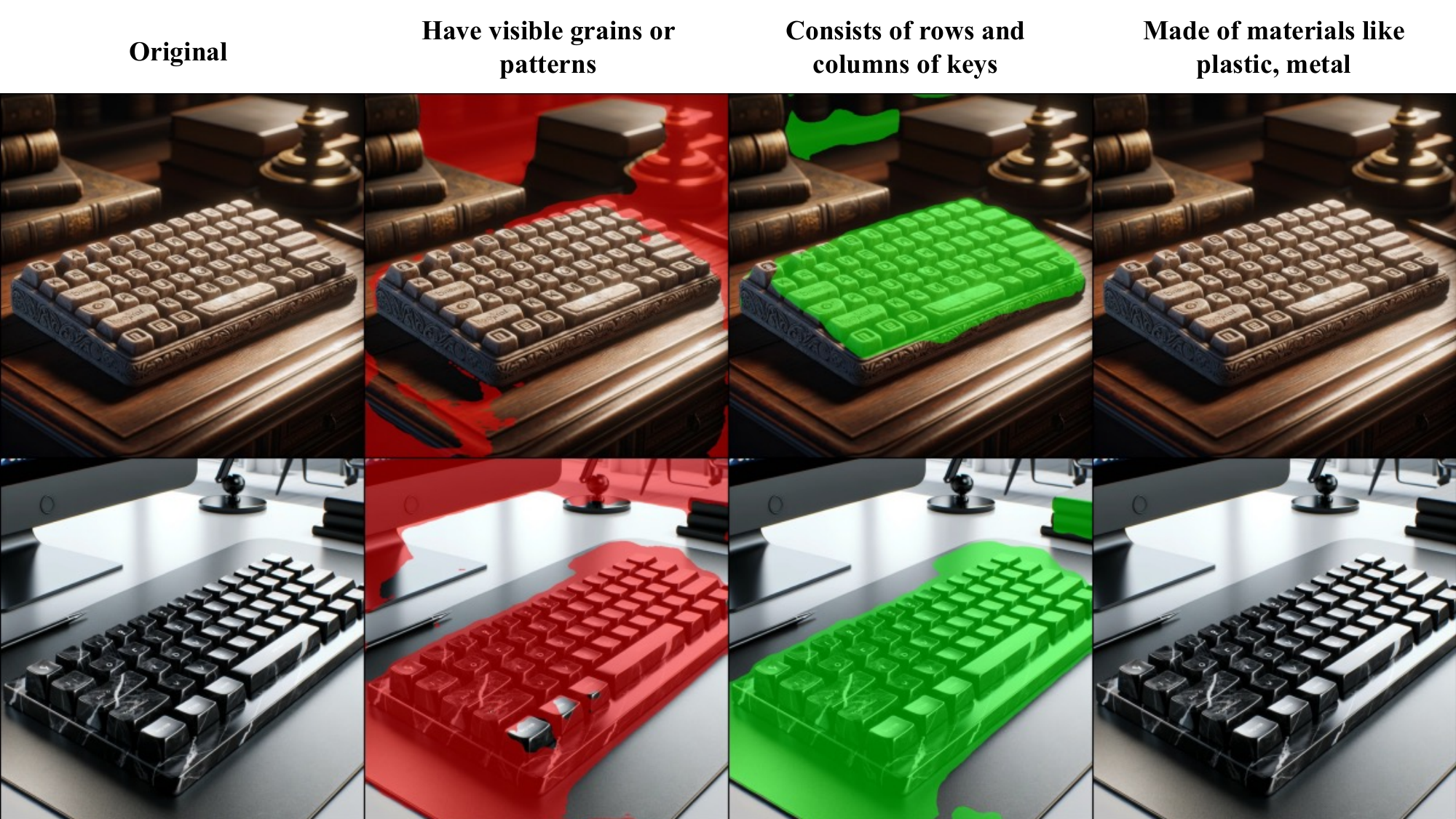}
\caption{\textbf{A keyboard made of stone}}
\label{fig:vis_7}
\end{subfigure}
\hfill
\begin{subfigure}[b]{0.48\linewidth}
\includegraphics[width=\linewidth]{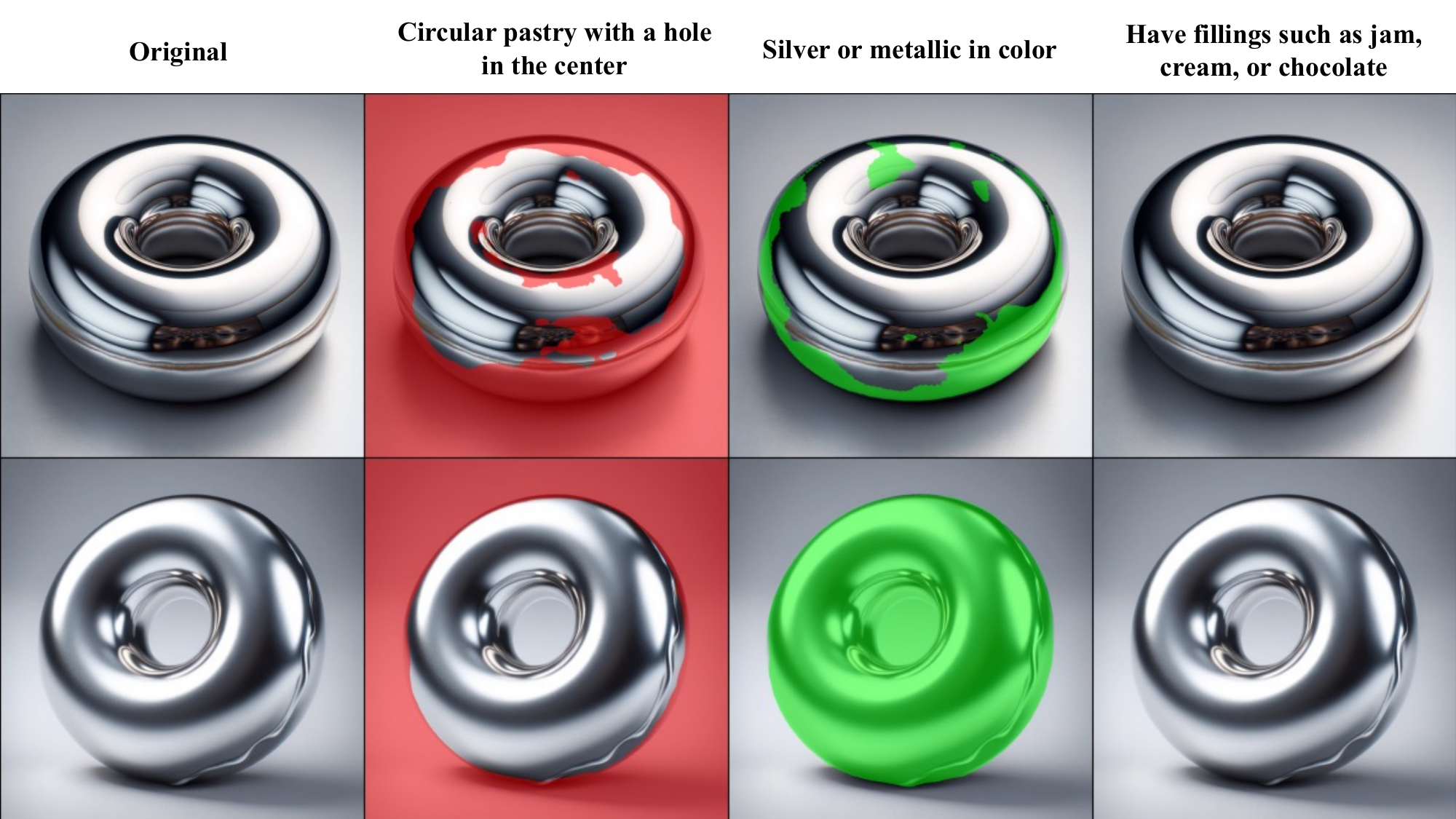}
\caption{\textbf{A silver donut}}
\label{fig:vis_8}
\end{subfigure}
\vspace{-2mm}
\caption{\textbf{Evaluation with out-of-domain generated images}. The images used in this evaluation are generated by DALL-E 3~\cite{betker2023dalle3} using a variety of creative prompts. Details of these prompts are provided in their respective captions.}
\label{fig:eval_ood_gen}
\vspace{-6mm}
\end{figure*}

Given our model's ability to segment objects based on descriptive properties, we generate a set of images using creative prompts (e.g., "a car covered in fur") to assess our model's performance with respect to the model's comprehension of various properties and its capability to produce reasonable results. Qualitative results are shown in \Cref{fig:eval_ood_gen}.

\vspace{-2mm}

\section{Descriptive Properties Details}
\label{appendix:properties}

This section is to provide the detailed 256 descriptive properties of COCO-Stuff (detailed in \Cref{tab:coco_property_1}--\ref{tab:coco_property_8}) and ADE20K (detailed in \Cref{tab:ade_property_1}--\ref{tab:ade_property_7}).

\section{Categories with  ``Similar" Properties}
\label{appendix:similar_categories}
\textbf{ProLab} could model the correlations of different categories based on the ratio of shared properties. To show that \textbf{ProLab}'s understanding of similar categories aligns well with human understanding, we collect a set of category pairs that are ``similar"  (the cosine similarity between their multi-hot property-level labels higher than 0.5). As shown in \Cref{tab:coco_shared_properties_1}-\ref{tab:ade_shared_properties_6}, categories sharing a lot of ``similar" properties are somehow consistent with human understanding.

\begin{table*}[t]
\centering
\tablestyle{3pt}{1.0}
\resizebox{\columnwidth}{!}{
\begin{tabular}{c|p{6cm}|p{10cm}}
\shline
Idx & Descriptive Property & Categories with this Property \\
\hline
1 & Comes in various sizes, shapes, designs, and variations & net, skis, bed, book, house, cupboard, bottle, cell phone, teddy bear, food-other, desk-stuff, baseball glove, bench, wall-other, tie, scissors, paper, roof, mirror-stuff, spoon, handbag, door-stuff, clock, counter, tent, parking meter, laptop, rug, tennis racket, wine glass, person, potted plant, cup, boat, suitcase, plastic, metal, skyscraper, furniture-other, clothes, railing, ceiling-other. \\
\hline
2 & Texture varies or can range from smooth to rough or different textures & book, bed, cloth, gravel, plant-other, textile-other, wall-brick, fruit, tree, dirt, wall-other, window-blind, curtain, solid-other, rock, person, potted plant, toothbrush, elephant, pizza, branch, floor-tile, chair, structural-other, clothes. \\
\hline
3 & Rectangular shape with varying dimensions & book, cupboard, snowboard, cell phone, cabinet, wall-brick, stop sign, microwave, wall-wood, oven, toaster, paper, wall-panel, waterdrops, refrigerator, remote, counter, parking meter, laptop, banner, car, suitcase, bus, skateboard. \\
\hline
4 & The orientation of the object can vary, including vertical, horizontal, diagonal, or inclined. & book, bed, grass, plant-other, stop sign, stairs, food-other, keyboard, river, curtain, train, mirror-stuff, clock, clouds, hair drier, car, potted plant, leaves, boat, branch, bus, skyscraper, tv, wall-tile. \\
\hline
5 & The shape of the object varies, but it is typically rectangular or has a rectangular shape. & cloth, table, sandwich, floor-wood, mat, pillow, cake, mirror-stuff, handbag, sink, cage, clock, tent, hair drier, bowl, platform, rug, window-other, toothbrush, floor-tile, skyscraper, chair, dining table, wall-tile. \\
\hline
6 & Made of various materials, including metal, wood, plastic, glass, ceramic, and composite materials & fire hydrant, fork, bottle, cell phone, cabinet, baseball bat, airplane, bench, scissors, oven, roof, spoon, cage, door-stuff, window-other, cup, truck, motorcycle, bus, fence, railing, bicycle. \\
\hline
7 & Location or context of usage & book, table, floor-other, shelf, desk-stuff, wall-other, curtain, cage, waterdrops, remote, clock, counter, solid-other, platform, laptop, wall-stone, banner, structural-other, couch, ceiling-other. \\
\hline
8 & Found in various environments & branch, tree, bush, dirt, flower, gravel, bench, plant-other, bird, vegetable, fruit, apple, dog, horse, fence, stairs, elephant, playingfield. \\
\hline
9 & Made of various materials & wall-panel, bed, branch, table, skateboard, solid-other, knife, cupboard, chair, floor-wood, furniture-other, floor-other, building-other, dining table, structural-other, shelf, wall-other, desk-stuff. \\
\hline
10 & Common colors include white, off-white, beige, gray, brown, green, various shades, clear, opaque, darker shades, metallic tones, black, neutral tones, vibrant patterns, tan, pastel shades. & wall-panel, door-stuff, gravel, floor-tile, cupboard, bottle, towel, wall-wood, window-blind, dining table, teddy bear, curtain, rug, tree, napkin, toilet, paper, wall-tile. \\
\hline
11 & Smooth or textured surface & wall-panel, suitcase, remote, sports ball, vase, plastic, skyscraper, bench, bottle, parking meter, cell phone, cabinet, dining table, toaster, bridge, wall-tile, desk-stuff. \\
\hline
12 & Smooth and glossy surface & sink, skis, refrigerator, microwave, bus, airplane, fork, frisbee, cup, scissors, oven, car, baseball bat, surfboard, toilet, floor-marble. \\
\hline
13 & Shows a spectrum of colors and shades & branch, grass, potted plant, moss, mountain, chair, floor-wood, plant-other, floor-other, salad, fruit, apple, leaves, vegetable, traffic light, hill. \\
\hline
14 & Common colors include black, white, silver, and red & clock, counter, metal, umbrella, textile-other, tie, cell phone, shelf, mouse, tv, oven, toaster, car, bicycle, desk-stuff. \\
\hline
15 & Vertical structure & wall-panel, door-stuff, skyscraper, parking meter, wall-other, wall-wood, structural-other, wall-brick, wall-concrete, stairs, wall-stone, window-other, wine glass, bridge, traffic light. \\
\hline
16 & Made from a variety of materials & sports ball, baseball glove, mat, textile-other, tie, blanket, towel, window-blind, curtain, rug, backpack, clothes, napkin, banner, handbag. \\
\hline
17 & Can have various patterns or designs & net, wall-panel, vase, floor-tile, mat, textile-other, tie, floor-other, wall-brick, rug, clothes, railing, wall-stone, paper, wall-tile. \\
\hline
18 & Horizontal surface or structure & table, carpet, ceiling-other, counter, pavement, bench, mat, platform, dining table, couch, railing, ground-other, banner, shelf. \\
\hline
19 & Comes in various variations or varieties & pizza, carrot, donut, broccoli, sandwich, fork, vegetable, salad, apple, oven, toaster, hot dog, food-other, spoon. \\
\hline
20 & Comes in various sizes and shapes, ranging from small to large & book, waterdrops, airplane, plant-other, bird, furniture-other, banana, cabinet, fruit, rug, clothes, tv, desk-stuff. \\
\hline
21 & Found in various landscapes and environments & boat, stone, sheep, moss, mountain, bear, river, water-other, rock, fog, mud, hill, sand. \\
\hline
22 & Equipped with a handle or grip for holding and maneuvering & suitcase, skateboard, hair drier, umbrella, kite, bottle, scissors, tennis racket, wine glass, frisbee, baseball bat, cup, handbag. \\
\hline
23 & Affected by or influenced by the environmental context & bed, cloth, metal, plant-other, chair, textile-other, vegetable, clothes, light, person, food-other, potted plant. \\

\bottomrule
\end{tabular}}
\caption{\textbf{COCO Stuff Properties} (1 of 8)}
\label{tab:coco_property_1}
\end{table*}

\begin{table*}[t]
\centering
\tablestyle{3pt}{1.0}
\resizebox{\columnwidth}{!}{
\begin{tabular}{c|p{7cm}|p{10cm}}
\shline
Idx & Descriptive Property & Categories with this Property \\
\hline
24 & Covered in vegetation or landscaping & cow, tree, grass, bush, dirt, house, mountain, river, snow, ground-other, hill, sand. \\
\hline
25 & Often found in various environments such as bathrooms, bedrooms, kitchens, offices, etc. & sink, vase, hair drier, cupboard, bottle, mat, cabinet, teddy bear, toilet, toothbrush, mirror-stuff, wall-tile. \\
\hline
26 & Varies in color, ranging from brown to gray to earth tones & sky-other, bed, stone, dirt, mud, pavement, wall-stone, ground-other, floor-stone, light, roof, rock. \\
\hline
27 & Specify the primary physical characteristics of each item. & book, bed, motorcycle, potted plant, flower, plant-other, river, vegetable, clothes, keyboard, leaves. \\
\hline
28 & Often decorated with frosting, icing, vibrant colors, logos, text, graphics, patterns, and designs & skateboard, kite, bottle, snowboard, banner, cake, stop sign, frisbee, train, surfboard, truck. \\
\hline
29 & Smooth texture with slight roughness or texture & broccoli, grass, counter, pavement, tie, orange, cake, wall-concrete, cardboard, straw, sand. \\
\hline
30 & Made of transparent or translucent material & net, sky-other, waterdrops, vase, plastic, wall-wood, water-other, window-other, wine glass, fog, light. \\
\hline
31 & Smooth, rounded shape & donut, motorcycle, sports ball, bowl, airplane, road, snowboard, orange, apple, hill. \\
\hline
32 & Often painted or coated for protection or aesthetic purposes & boat, fire hydrant, bench, airplane, parking meter, floor-wood, wall-wood, fence, railing, bridge. \\
\hline
33 & Typically made of metal, plastic, or bamboo & refrigerator, microwave, remote, hair drier, parking meter, stop sign, mouse, toaster, frisbee, toothbrush. \\
\hline
34 & Long and slender shape & broccoli, cat, baseball glove, water-other, teddy bear, zebra, straw, toothbrush, giraffe, handbag. \\
\hline
35 & May have visible patterns, markings, or unique characteristics & stone, clouds, sports ball, wood, wall-other, wall-wood, horse, rock, floor-stone, ceiling-tile. \\
\hline
36 & Comes in various colors, including brown, tan, black, white, beige, gray, and reddish-brown & sheep, baseball glove, wood, wall-brick, couch, cardboard, straw, baseball bat, cup, sand. \\
\hline
37 & Comes in various shapes, sizes, and variations & net, bed, table, skateboard, tent, platform, furniture-other, structural-other, bridge, shelf. \\
\hline
38 & Comes in various sizes, thicknesses, lengths, widths, and depths & wall-panel, branch, bowl, knife, mat, towel, pillow, paper, surfboard, ceiling-tile. \\
\hline
39 & Often found in various culinary settings, including kitchens, restaurants, cafes, and homes & refrigerator, bowl, fork, knife, dining table, scissors, oven, toaster, straw, cup. \\
\hline
40 & Flat or level surface & net, table, pavement, platform, towel, laptop, rug, napkin, paper, playingfield. \\
\hline
41 & Environmental context or surroundings & net, clock, skyscraper, pavement, platform, ground-other, bridge, roof, leaves, mirror-stuff. \\
\hline
42 & Can be used for advertising, promotion, construction, furniture, decoration, landscaping, or functional purposes & stone, gravel, metal, solid-other, wood, furniture-other, rug, rock, banner, wall-tile. \\
\hline
43 & Made from a variety of materials, including natural and synthetic fibers, feathers, down, paper fibers, mineral fiber, metal, dried plant stems, wood pulp, and recycled fibers. & net, cloth, carpet, wood, textile-other, pillow, cardboard, straw, paper, ceiling-tile. \\
\hline
44 & Features doors, windows, and various levels of access & cage, door-stuff, bus, house, skyscraper, window-blind, oven, car, train. \\
\hline
45 & Can be found in various environments, including urban, suburban, and rural areas & bus, house, skyscraper, pavement, road, railroad, building-other, wall-concrete, train. \\
\hline
46 & Rectangular shape & bed, house, umbrella, backpack, tv, cardboard, keyboard, truck, shelf. \\
\hline
47 & Can have different finishes or appearances & carpet, house, metal, solid-other, platform, wall-brick, cabinet, wall-concrete, truck. \\
\hline
48 & Constructed from various materials such as plaster, wood, metal, concrete, steel, fiberglass, glass, brick, asphalt, or stone & boat, house, skyscraper, pavement, platform, road, stairs, ceiling-other, bridge. \\
\hline
49 & Reflective or has a reflective surface & fire hydrant, waterdrops, river, water-other, sea, snow, window-other, light, mirror-stuff. \\
\hline
50 & Comes in a wide range of colors and patterns & cloth, carpet, floor-marble, plastic, flower, blanket, backpack, dog, handbag. \\
\hline
51 & Contains racks, shelves, drawers, or compartments for organization and storage & refrigerator, table, counter, cupboard, furniture-other, cabinet, backpack, oven, shelf. \\
\hline
52 & May feature decorative patterns, designs, or elements & cloth, bowl, wall-other, towel, pillow, dining table, napkin, cup, handbag. \\
\hline
53 & Can vary in size, shape, and weight depending on user preference or purpose & cage, cloth, carpet, plastic, metal, textile-other, river, snowboard, sand. \\
\hline
54 & Can have various shapes and designs & wood, mountain, road, wall-other, stairs, roof, window-other, straw, toothbrush. \\

\bottomrule
\end{tabular}}
\caption{\textbf{COCO Stuff Properties} (2 of 8)}
\label{tab:coco_property_2}
\end{table*}

\begin{table*}[t]
\centering
\tablestyle{3pt}{1.0}
\resizebox{\columnwidth}{!}{
\begin{tabular}{c|p{7cm}|p{10cm}}
\shline

Idx & Descriptive Property & Categories with this Property \\
\hline
55 & Often found in residential and commercial spaces & wall-panel, carpet, flower, floor-wood, window-blind, wall-concrete, window-other, ceiling-tile, floor-marble. \\
\hline
56 & Made of various materials such as glass, ceramic, metal, plastic, and wood & clock, vase, metal, bowl, laptop, tv, car, keyboard. \\
\hline
57 & Features various signs, markings, and signals for traffic control and guidance & bus, pavement, road, parking meter, railroad, stop sign, cardboard, playingfield. \\
\hline
58 & Diverse shapes and sizes & boat, vase, flower, river, vegetable, orange, fruit, leaves. \\
\hline
59 & Covered in fur, which can vary in color, pattern, length, and texture & cow, sheep, cat, bear, teddy bear, dog, horse, giraffe. \\
\hline
60 & Equipped with handles or knobs for opening and closing & refrigerator, door-stuff, vase, counter, cupboard, fork, cabinet, spoon. \\
\hline
61 & Contains seeds or pits, although some varieties are seedless & branch, vase, bush, plant-other, orange, fruit, cake, tree. \\
\hline
62 & Exhibits varying levels of brightness and may have decorative elements or lighting & sky-other, ceiling-other, furniture-other, light, stairs, window-other, banner, bridge. \\
\hline
63 & Irregular shape or irregularly shaped stones & stone, gravel, bush, dirt, solid-other, salad, wall-stone, rock. \\
\hline
64 & Covered or enclosed in a protective material & book, clock, tent, umbrella, pillow, person, traffic light. \\
\hline
65 & Has a rounded head with two ears, two eyes, a snout, and a mouth & sheep, cat, bear, teddy bear, dog, person, giraffe. \\
\hline
66 & Transportation vehicle with varying numbers of wheels & motorcycle, bus, skateboard, airplane, car, truck, bicycle. \\
\hline
67 & Cylindrical or rectangular shape & pizza, suitcase, bowl, bottle, hot dog, train, cup. \\
\hline
68 & Possesses a tail, which can vary in length and appearance & cow, cat, kite, bear, horse, elephant, giraffe. \\
\hline
69 & Comes in a variety of colors, including vibrant shades & skis, suitcase, sports ball, hair drier, laptop, tennis racket, frisbee. \\
\hline
70 & Smooth texture with possible variations in appearance or feel & donut, carrot, floor-wood, wall-wood, wine glass, food-other, paper. \\
\hline
71 & Can be eaten raw, cooked, or used in cooking and baking & donut, banana, orange, fruit, apple, cake, vegetable. \\
\hline
72 & Grows in warm climates, low-growing, dense vegetation growth, grows in clusters, seeds contained in the core, grows underground in soil & carrot, broccoli, bush, moss, banana, orange, apple. \\
\hline
73 & Typically has a stem and petals & branch, broccoli, flower, moss, plant-other, apple, potted plant. \\
\hline
74 & Rectangular shape with a frame and panel design & door-stuff, towel, blanket, window-blind, railing, napkin, ceiling-tile. \\
\hline
75 & Soft and plush texture & carpet, moss, towel, blanket, rug, snow, fog. \\
\hline
76 & Can be natural or man-made & stone, plastic, wall-stone, ground-other, person, playingfield. \\
\hline
77 & Comes in varying sizes, heights, and lengths & grass, vase, curtain, cake, fence, person. \\
\hline
78 & Can have additional accessories or elements & kite, tie, teddy bear, person, bicycle, desk-stuff. \\
\hline
79 & Can have a polished or matte finish & donut, table, bowl, apple, floor-stone, bicycle. \\
\hline
80 & Can be used in various outdoor settings & tent, structural-other, wall-brick, wall-stone, frisbee, bicycle. \\
\hline
81 & Commonly found in various locations such as parking lots, streets, intersections, and designated riding areas & motorcycle, parking meter, stop sign, car, traffic light, truck. \\
\hline
82 & Soft, cushion-like object typically found on beds or sofas, upholstered with fabric or leather, may have textured or padded seat, often accompanied by throw pillows & motorcycle, furniture-other, blanket, pillow, couch, tv. \\
\hline
83 & Comes in various shapes and orientations & clouds, structural-other, building-other, vegetable, food-other, traffic light. \\
\hline
84 & Contains buttons or controls for navigation and additional functions. & microwave, remote, traffic light, cell phone, laptop, keyboard. \\
\hline
85 & Highlight a distinctive shape or structural feature central to the function or identity of the item. & fire hydrant, donut, microwave, wine glass, toilet, spoon. \\
\hline
86 & Quadrupedal body shape & cow, sheep, bear, dog, horse, zebra. \\
\hline
87 & Elongated or flat shape with various features such as curved tips, rounded backs, or raised lips & skis, cow, branch, mouse, frisbee, surfboard. \\
\hline
88 & White or off-white in color, often with white markings or painted in neutral colors & cow, snow, wall-concrete, ceiling-other, toothbrush, ceiling-tile. \\
\hline
89 & Contains compartments, pockets, or levels for organization & book, refrigerator, suitcase, shelf, backpack, handbag. \\
\hline
90 & Common features and additional elements or details & book, textile-other, blanket, curtain, backpack, clothes. \\
\hline
91 & Can be easily folded, draped, or assembled & cloth, umbrella, mat, blanket, cardboard, napkin. \\
\hline
92 & Commonly used for various purposes such as fashion, functionality, cutting, protection, cushioning, cleaning, and decoration. & cloth, mat, textile-other, blanket, scissors, handbag. \\
\bottomrule
\end{tabular}}
\caption{\textbf{COCO Stuff Properties} (3 of 8)}
\label{tab:coco_property_3}
\end{table*}

\begin{table*}[t]
\centering
\tablestyle{3pt}{1.0}
\resizebox{\columnwidth}{!}{
\begin{tabular}{c|p{7cm}|p{10cm}}
\shline
Idx & Descriptive Property & Categories with this Property \\
\hline
93 & Typically made of durable and lightweight materials & skis, suitcase, kite, snowboard, tennis racket, surfboard. \\
\hline
94 & Handheld or portable device & remote, baseball glove, hair drier, cell phone, laptop, mouse. \\
\hline
95 & Can have a smooth or rough surface & stone, skateboard, floor-tile, wood, water-other, mouse. \\
\hline
96 & Commonly silver, metallic, white, black, or stainless steel in color & refrigerator, microwave, remote, fork, scissors, spoon. \\
\hline
97 & Can be served hot or cold, and can be prepared in various ways (steamed, roasted, stir-fried, grilled, boiled) & broccoli, sandwich, salad, snow, hot dog, food-other. \\
\hline
98 & Common features and variations in characteristics among different species & bush, flower, plant-other, vegetable, potted plant, leaves. \\
\hline
99 & Provides habitat and shelter for animals & cage, grass, bush, moss, river, tree. \\
\hline
100 & Flexible or rigid depending on composition & cloth, plastic, floor-tile, textile-other, cardboard, food-other. \\
\hline
101 & Surrounded by other buildings or open spaces & clouds, skyscraper, mountain, sea, railing, playingfield. \\
\hline
102 & Provides privacy, security, and light control & door-stuff, wall-other, window-blind, curtain, fence, window-other. \\
\hline
103 & Composed of various materials such as soil, sand, rocks, organic matter, and debris & gravel, dirt, ground-other, mud, hill, sand. \\
\hline
104 & Horizontal expanse, foundational surface & floor-tile, floor-wood, floor-other, floor-stone, sky-other, floor-marble. \\
\hline
105 & Can provide relaxation, aesthetics, warmth, and comfort & wall-panel, floor-wood, floor-other, pillow, rug, floor-marble. \\
\hline
106 & Consists of a series of steps or treads & house, floor-other, stairs, floor-stone, mud, wall-tile. \\
\hline
107 & Weather or atmospheric phenomena & sea, fruit, snow, fog, sky-other, sand. \\
\hline
108 & Hard and durable composition & stone, grass, metal, wood, salad, leaves. \\
\hline
109 & Features a seat, brakes, gears, leash attachment point, backrest, legs, arms, solid frame, cushioned seat, saddles or bridles & chair, horse, person, surfboard, bicycle. \\
\hline
110 & Can accommodate multiple people, pedestrians, vehicles, or trains & bus, road, bench, railroad, bridge. \\
\hline
111 & Includes various interior features such as seating, storage compartments, and sleeping quarters & boat, bed, bus, house, couch. \\
\hline
112 & Can have various colors, including orange, purple, yellow, white, gray, black, or red & carrot, fog, clouds, banner, truck. \\
\hline
113 & Used for transportation or recreation on water & boat, net, river, water-other, bridge. \\
\hline
114 & Positioned along edges or at ground level & fire hydrant, gravel, road, railroad, railing. \\
\hline
115 & Rectangular shape with a backrest and optional armrests & bed, bench, chair, couch, mouse. \\
\hline
116 & Contains diverse species with distinct physical characteristics, behaviors, and temperaments & cat, bird, bear, sea, dog. \\
\hline
117 & Commonly used in outdoor or recreational settings & tent, umbrella, kite, backpack, hot dog. \\
\hline
118 & Features adjustable temperature settings, chimneys, vents, skylights, tilt function, wind vents, nozzle for directing airflow, windows or vents for ventilation & refrigerator, tent, hair drier, umbrella, roof. \\
\hline
119 & Long and narrow shape with pointed ends or a narrow neck and wider base & vase, knife, bottle, tie, hot dog. \\
\hline
120 & Has a reflective surface and can have a glossy or matte finish & floor-tile, metal, snowboard, keyboard, floor-marble. \\
\hline
121 & Cylindrical or oval-shaped object with a handle or stem & solid-other, tennis racket, wine glass, straw, baseball bat. \\
\hline
122 & Tapered shape with wider ends and narrower middle & carrot, banana, wine glass, baseball bat, cup. \\
\hline
123 & Water bodies with varying characteristics and behaviors & waterdrops, river, water-other, sea, surfboard. \\
\hline
124 & Lightweight & remote, plastic, cell phone, tennis racket, paper. \\
\hline
125 & Consists of liquid or semi-liquid substances & waterdrops, bottle, water-other, mud, sand. \\
\hline
126 & Often used for multiple purposes, such as holding objects, serving, dining, or displaying & table, counter, platform, wine glass, shelf. \\
\hline
127 & Smooth texture with a polished, matte, or glossy finish & laptop, mouse, tv, mirror-stuff, spoon. \\
\hline
128 & Thrives in wet or humid environments & waterdrops, moss, banana, fog, mud. \\
\hline
129 & Food item consisting of layers of ingredients between two slices of bread & sandwich, salad, cake, toaster, hot dog. \\
\bottomrule
\end{tabular}}
\caption{\textbf{COCO Stuff Properties} (4 of 8)}
\label{tab:coco_property_4}
\end{table*}

\begin{table*}[t]
\centering
\tablestyle{3pt}{1.0}
\resizebox{\columnwidth}{!}{
\begin{tabular}{c|p{7cm}|p{10cm}}
\shline
Idx & Descriptive Property & Categories with this Property \\
\hline
130 & Served as a dish or food item & pizza, broccoli, sandwich, salad, cake. \\
\hline
131 & Edible & carrot, broccoli, vegetable, fruit, food-other. \\
\hline
132 & Typically flat or slightly textured, uneven, or curved & pizza, ceiling-tile, ground-other, ceiling-other, mirror-stuff. \\
\hline
133 & Made of various materials such as porcelain, stainless steel, stone, ceramic, plastic, or clay & sink, floor-tile, toilet, potted plant, wall-tile. \\
\hline
134 & Can be connected to a computer or other electronic devices, often accompanied by buttons or a remote control, enables voice calls, messaging, internet browsing, and app usage, can be analog or digital & clock, cell phone, mouse, tv, keyboard. \\
\hline
135 & Can be mounted on walls or placed on furniture & wall-panel, pillow, wall-concrete, tv, mirror-stuff. \\
\hline
136 & Exhibits unique architectural or design elements & skyscraper, solid-other, structural-other, building-other, wall-brick. \\
\hline
137 & Environmentally friendly or sustainable options & plastic, building-other, cardboard, straw, paper. \\
\hline
138 & Smooth or textured surface, depending on the material or type of stone & door-stuff, cupboard, floor-other, floor-stone, shelf. \\
\hline
139 & Provides insulation, protection, and temporary shelter from weather elements & tent, moss, wall-brick, wall-concrete, roof. \\
\hline
140 & Upright orientation with handlebars and pedals for steering and propulsion & person, motorcycle, chair, bicycle. \\
\hline
141 & Frame shape can be diamond or step-through & fence, ceiling-tile, bicycle, floor-wood. \\
\hline
142 & Equipped with common features such as headlights, taillights, turn signals, and mirrors & car, traffic light, motorcycle, truck. \\
\hline
143 & Operates or flies in the sky & cage, sky-other, airplane, kite. \\
\hline
144 & Connected to a plumbing system for water supply & sink, fire hydrant, carpet, toilet. \\
\hline
145 & Handheld utensil for eating or serving & hot dog, fork, bird, spoon. \\
\hline
146 & Domesticated mammal, commonly kept as a pet & cow, sheep, cat, dog. \\
\hline
147 & Stands on four legs for stability & elephant, giraffe, chair, horse. \\
\hline
148 & Often found in groups or herds & cow, sheep, zebra, grass. \\
\hline
149 & Has a long and flexible trunk or neck & elephant, giraffe, bush, tree. \\
\hline
150 & Portable and used for carrying personal items & suitcase, umbrella, backpack, handbag. \\
\hline
151 & Equipped with bindings or laces/straps for securing and adjusting fit & snowboard, skis, baseball glove, tie. \\
\hline
152 & Used in various sports and recreational activities & frisbee, playingfield, baseball glove, sports ball. \\
\hline
153 & Size, weight, and dimensions vary depending on the sport, game, skill level, and age group & frisbee, baseball bat, playingfield, sports ball. \\
\hline
154 & Used for stirring, scooping, eating, holding, and drinking food or beverages & straw, bowl, cup, spoon. \\
\hline
155 & Fluid and constantly in motion & railroad, sea, light, fork. \\
\hline
156 & May have fillings or layers with various ingredients & donut, hot dog, sandwich, cake. \\
\hline
157 & Topped with various condiments and ingredients & salad, hot dog, food-other, pizza. \\
\hline
158 & Furniture item for sitting and hosting other objects & floor-other, dining table, chair, desk-stuff. \\
\hline
159 & Provides cushioning, support, and comfort & pillow, floor-stone, couch, railing. \\
\hline
160 & Features control knobs, digital display, buttons, switches, and additional features & toaster, hair drier, microwave, oven. \\
\hline
161 & Features a handrail or lever for support and safety & toaster, bridge, stairs, railing. \\
\hline
162 & Can be standalone or integrated into larger structures & structural-other, building-other, wall-other, furniture-other. \\
\hline
163 & Enclosed or boundary structure with bars, mesh, or a mesh-like structure & cage, net, fence, tent. \\
\hline
164 & Can be affected by weather conditions & road, snow, grass, clouds. \\
\hline
165 & Made of natural or durable materials such as marble, granite, wood, or laminate & rock, gravel, counter, floor-marble. \\
\hline
166 & Sturdy and durable construction & wall-brick, floor-stone, metal, wall-stone. \\
\hline
167 & Provides a base for various activities and movements & person, ground-other, sports ball. \\
\hline
168 & Wheeled recreational device with rubber tires and spokes & skateboard, surfboard, bicycle. \\
\hline
169 & Variations or types of the described object & skateboard, motorcycle, bicycle. \\
\bottomrule
\end{tabular}}
\caption{\textbf{COCO Stuff Properties} (5 of 8)}
\label{tab:coco_property_5}
\end{table*}

\begin{table*}[t]
\centering
\tablestyle{3pt}{1.0}
\resizebox{\columnwidth}{!}{
\begin{tabular}{c|p{7cm}|p{10cm}}
\shline
Idx & Descriptive Property & Categories with this Property \\
\hline
170 & Variations in type or design & car, truck, bus. \\
\hline
171 & Features windows along the fuselage and wings for flight & bird, airplane, bus. \\
\hline
172 & Equipped with fins, oars, sails, engines, and landing gear & boat, airplane, surfboard. \\
\hline
173 & Operates on designated routes and schedules, connects different locations or serves as a route, operates on roads or designated bus lanes & road, train, bus. \\
\hline
174 & Can be upright or drooping & flower, banner, traffic light. \\
\hline
175 & Equipped with a display screen and user interaction interface & cell phone, laptop, parking meter. \\
\hline
176 & Possesses claws or hooves for walking or running & horse, cat, dog. \\
\hline
177 & Known for its speed, agility, durability, and elegance & cat, horse, floor-marble. \\
\hline
178 & Known for their wool production, meat, intelligence, social behavior, loyalty, and companionship & elephant, sheep, dog. \\
\hline
179 & Large and distinctive ears or horns & cow, elephant, giraffe. \\
\hline
180 & Mammal native to Africa and typically found in grassland or savannah habitats & elephant, giraffe, zebra. \\
\hline
181 & May have a closure mechanism such as a zipper, magnetic closure, or buttons & pillow, baseball glove, handbag. \\
\hline
182 & Features doors, zippered entrance, and hinged lid & tent, cupboard, suitcase. \\
\hline
183 & Handheld tool used for cutting or oral hygiene & knife, scissors, toothbrush. \\
\hline
184 & Consists of two blades joined at a pivot point, typically made of metal, sharp and shiny, straight and elongated blades, often has a serrated edge for sawing, thin and elongated blades with sharp edges for cutting. & knife, scissors, grass. \\
\hline
185 & Shape of the fruit is spherical or curved/elongated & banana, orange, apple. \\
\hline
186 & May have a sweet, sour, or tangy taste & banana, orange, fruit. \\
\hline
187 & Commonly used in cooking, salads, desserts, and beverages & orange, carrot, apple. \\
\hline
188 & Features green leaves & carrot, broccoli, bush. \\
\hline
189 & Often enjoyed on special occasions or celebrations & donut, pizza, cake. \\
\hline
190 & May have decorative elements or details & couch, wall-wood, chair. \\
\hline
191 & Comes in various variations or types & napkin, table, chair. \\
\hline
192 & Typically positioned at waist height or built into cabinetry & dining table, oven, shelf. \\
\hline
193 & May have additional features or attachments & sink, toilet, hair drier. \\
\hline
194 & Kitchen appliance for cooking, baking, heating, or cooling food & refrigerator, microwave, oven. \\
\hline
195 & Used for cleaning or drying purposes & napkin, toothbrush, towel. \\
\hline
196 & Variations in material or type & stone, floor-stone, toothbrush. \\
\hline
197 & Commonly used for packaging and containers & paper, plastic, cardboard. \\
\hline
198 & Overhead or positioned parallel to the floor & ceiling-other, roof, ceiling-tile. \\
\hline
199 & May have growth or formations on its surface & waterdrops, wall-stone, clouds. \\
\hline
200 & Can be woven or solid, can be solid or have gaps between components & curtain, fence, mat. \\
\hline
201 & Loose or compacted, depending on moisture content and usage & sand, gravel, dirt. \\
\hline
202 & Used in interior spaces, especially high-traffic areas & floor-tile, floor-stone, wall-wood. \\
\hline
203 & Exists in different states or forms, such as ice, vapor, fog, or haze & water-other, fog, sky-other. \\
\hline
204 & Natural landform with sloping sides, often steep or rugged, typically elevated and rocky & mountain, hill, river. \\
\hline
205 & Can be prone to cracks, potholes, rot, or decay if not properly treated or maintained & wood, pavement, wall-concrete. \\
\hline
206 & Expressive facial features & person, dog. \\
\bottomrule
\end{tabular}}
\caption{\textbf{COCO Stuff Properties} (6 of 8)}
\label{tab:coco_property_6}
\end{table*}

\begin{table*}[t]
\centering
\tablestyle{3pt}{1.0}
\resizebox{\columnwidth}{!}{
\begin{tabular}{c|p{7cm}|p{10cm}}
\shline
Idx & Descriptive Property & Categories with this Property \\
\hline
207 & May have a hull, deck, superstructure, tail fins, and stabilizers & boat, airplane. \\
\hline
208 & Mode of transportation on tracks & railroad, train. \\
\hline
209 & Can be operated manually or with a motorized system and can be powered by electricity or diesel & train, window-blind. \\
\hline
210 & Emits sounds and vibrations when in motion & train, bird. \\
\hline
211 & Tall and vertical in shape & fire hydrant, tree. \\
\hline
212 & Small to medium-sized and typically quadrupedal & apple, cat. \\
\hline
213 & Often associated with comfort, childhood, and companionship & teddy bear, cat. \\
\hline
214 & Tall, long-necked mammal & giraffe, horse. \\
\hline
215 & May have horns, depending on the breed & cow, sheep. \\
\hline
216 & Mammal with a large, stocky build, distinct trunk, and elongated tusks & elephant, bear. \\
\hline
217 & Short and coarse fur texture & giraffe, zebra. \\
\hline
218 & Sports equipment used for snowboarding & snowboard, skis. \\
\hline
219 & Used for hitting objects in sports & baseball bat, tennis racket. \\
\hline
220 & Often used by players on a baseball field & baseball bat, baseball glove. \\
\hline
221 & Has a protective outer skin or peel & banana, fruit. \\
\hline
222 & Contains juicy and crisp flesh & orange, apple. \\
\hline
223 & Toasts bread slices & toaster, sandwich. \\
\hline
224 & Nutrient-rich and high in vitamins and minerals & carrot, broccoli. \\
\hline
225 & Plumbing fixture used for washing or cleaning and sanitary fixture for human waste disposal & sink, toilet. \\
\hline
226 & Electronic device with a screen for displaying visual content and a camera for capturing photos and videos & cell phone, tv. \\
\hline
227 & Can be opened and closed & laptop, window-other. \\
\hline
228 & Features buttons and controls for various functions & remote, mouse. \\
\hline
229 & Can be used to control various devices & cell phone, remote. \\
\hline
230 & Positioned horizontally, attached to a countertop or wall & sink, microwave. \\
\hline
231 & Covered in leaves or needles & branch, tree. \\
\hline
232 & Freestanding or built-in storage capability & cabinet, cupboard. \\
\hline
233 & Provides insulation and sound absorption & ceiling-tile, carpet. \\
\hline
234 & Formation of condensed water vapor in the atmosphere & fog, clouds. \\
\hline
235 & Made of metal rails supported by wooden or concrete ties & railroad, fence. \\
\hline
236 & Sticky and slippery texture & mud, floor-marble. \\
\hline
237 & Peaks or summits at the highest points & mountain, hill. \\
\hline
238 & Shapeless and intangible & mud, light. \\
\hline
239 & Natural formation or occurrence & snow, rock. \\
\hline
240 & Visible horizon line & sea, sky-other. \\
\hline
241 & Text is bold and in capital letters & stop sign. \\
\hline
242 & Body covered in colorful or patterned feathers & bird. \\
\hline
243 & Different species of elephants have distinct physical characteristics, such as the size of their ears and tusks. & elephant. \\
\hline
244 & Distinctive black and white striped pattern & zebra. \\
\hline
245 & Has strings stretched across the frame to create a mesh pattern & tennis racket. \\
\hline
246 & Bright orange color & orange. \\
\hline
247 & Often associated with vitamin C and citrus fruits & orange. \\
\hline
248 & Consists of labeled keys arranged in rows and columns & keyboard. \\
\bottomrule
\end{tabular}}
\caption{\textbf{COCO Stuff Properties} (7 of 8)}
\label{tab:coco_property_7}
\end{table*}

\begin{table*}[t]
\centering
\tablestyle{3pt}{1.0}
\resizebox{\columnwidth}{!}{
\begin{tabular}{c|p{7cm}|p{10cm}}
\shline
Idx & Descriptive Property & Categories with this Property \\
\hline
249 & Timekeeping device with numbers or Roman numerals and hour, minute, and second hands. & clock. \\
\hline
250 & Soft and cuddly toy in the shape of a bear & teddy bear. \\
\hline
251 & Can have an impact on travel and outdoor activities & fog. \\
\hline
252 & Physical characteristics depend on the specific food-other item. & food-other. \\
\hline
253 & Residential structure for human habitation & house. \\
\hline
254 & Often dressed with vinaigrettes or creamy dressings & salad. \\
\hline
255 & Vast body of saltwater & sea. \\
\hline
256 & Features a loop or hanging tag for easy storage & towel. \\
\bottomrule
\end{tabular}}
\caption{\textbf{COCO Stuff Properties} (8 of 8)}
\label{tab:coco_property_8}
\end{table*}

\begin{table*}[t]
\centering
\tablestyle{3pt}{1.0}
\resizebox{\columnwidth}{!}{
\begin{tabular}{c|p{7cm}|p{10cm}}
\shline
Idx & Descriptive Property & Categories with this Property \\
\hline
1 & Constructed with sturdy and durable materials & swimming pool, shower, escalator, skyscraper, wall, bar, fireplace, stage, building, runway, stairs, stairway, base, road, ceiling, booth, pier, step, hovel, sidewalk, grandstand, path, boat, wardrobe, floor, storage tank, fountain, ship, bridge, tower. \\
\hline
2 & Constructed or made from various materials such as wood, metal, glass, plastic, fabric, vinyl, etc. & buffet, car, lamp, tray, vase, bannister, signboard, chandelier, chair, fan, door, monitor, stool, plaything, bulletin board, hood, screen, seat, blind, sconce, table, canopy, television receiver, conveyer belt, clock, crt screen. \\
\hline
3 & Varies in shape, size, height, length, and design & pole, swimming pool, skyscraper, wall, fence, bannister, bench, plant, column, bed , flag, stool, railing, bottle, radiator, base, bookcase, screen door, apparel, wardrobe, curtain, person, waterfall, cushion, bridge. \\
\hline
4 & Made of various materials, including metal, wood, stone, concrete, plastic, and glass & pole, truck, fence, bench, cradle, box, column, bed , shelf, railing, pool table, cabinet, case, bookcase, coffee table, barrel, sculpture, screen door, van, chest of drawers, earth, desk, house, kitchen island. \\
\hline
5 & Smooth or textured surface & pole, buffet, skyscraper, vase, bannister, ashcan, signboard, bench, box, column, door, stairway, bottle, bookcase, barrel, step, washer, storage tank, chest of drawers, pot, stove, desk, kitchen island. \\
\hline
6 & Typically made of various materials, including metal, ceramic, glass, plastic, and composite materials & minibike, awning, streetlight, computer, microwave, airplane, ashcan, dishwasher, plate, arcade machine, bottle, oven, radiator, ball, bus, refrigerator, washer, bicycle, pot, stove. \\
\hline
7 & Surface texture can vary, ranging from smooth to textured & truck, painting, shower, wall, bag, stage, towel, book, radiator, cabinet, base, ceiling, plaything, pier, sculpture, wardrobe, fountain, bridge, tower. \\
\hline
8 & Orientation can vary, either vertically or horizontally. & painting, shower, skyscraper, building, signboard, bed , flag, sofa, book, screen, seat, screen door, apparel, mirror, sconce, television receiver, cushion, clock, kitchen island. \\
\hline
9 & Typically has a rectangular or circular shape & painting, shower, tray, bar, fireplace, stage, bulletin board, pier, screen door, sidewalk, grandstand, mirror, swivel chair, windowpane, blind, table, cushion, conveyer belt, clock. \\
\hline
10 & Commonly found in various environments such as homes, offices, public spaces, and buildings & swivel chair, book, bannister, awning, poster, swimming pool, chandelier, ashcan, base, case, bulletin board, wall, desk, stairs, clock, door, stairway, screen door. \\
\hline
11 & Features various elements and additional features & bannister, cabinet, blanket, storage tank, sconce, curtain, cradle, streetlight, pier, arcade machine, bed , television receiver, barrel, flag, sofa, armchair. \\
\hline
12 & Comes in various shapes, including rectangular, square, circular, cylindrical, irregular, and cuboid shapes. & buffet, rug, signboard, base, pot, coffee table, desk, house, box, column, fan, step, ottoman, plate, pillow, stool. \\
\hline
13 & Smooth and glossy surface & airplane, poster, minibike, toilet, swimming pool, car, oven, dishwasher, bus, coffee table, bathtub, sink, refrigerator, arcade machine, microwave. \\
\hline
14 & May have decorative patterns or designs & poster, awning, sculpture, blanket, pot, canopy, ball, plate, countertop, fence, glass, pillow, apparel, floor. \\
\hline
15 & Common colors include white, black, red, blue, yellow, gray, metallic finishes, silver, gold, natural wood tones, pastels, and brown. & pole, minibike, oven, car, radiator, cradle, stove, ball, computer, barrel, clock, hood, counter, monitor. \\
\hline
16 & Can be painted or adorned with various colors, finishes, logos, text, or images & airplane, runway, signboard, painting, van, bicycle, sconce, plaything, house, fence, sculpture, ship, grandstand. \\
\hline
17 & Varies in color, ranging from neutral tones to vibrant hues & sky, rug, painting, seat, chair, shower, person, cushion, sofa, armchair, apparel, sea. \\
\hline
18 & Provides scenic views and is surrounded by natural elements or landscapes & field, sky, runway, river, mountain, grass, skyscraper, hill, house, lake, bridge, land. \\
\hline
19 & Flat and smooth surface & runway, windowpane, pool table, road, table, glass, plate, box, tray, crt screen, screen, sidewalk. \\
\hline
20 & Contains drawers, compartments, or shelves for storage & book, buffet, minibike, cabinet, chest of drawers, dishwasher, bookcase, desk, bag, refrigerator, ship, wardrobe. \\
\hline
21 & Rectangular or square shape with straight edges and corners & building, kitchen island, light, car, pool table, booth, bus, wall, countertop, counter, railing. \\
\hline
22 & Smooth texture, often with a polished, glossy, or matte finish & mirror, monitor, television receiver, computer, conveyer belt, countertop, tray, counter, basket, bar, shelf. \\
\hline
23 & Contains various interior features and amenities & oven, booth, bus, house, refrigerator, countertop, hovel, bar, microwave, boat, kitchen island. \\
\hline
24 & Irregular or varying shape & field, rock, swimming pool, sand, cradle, waterfall, lake, hood, hovel, seat, armchair. \\
\hline
25 & Often located in outdoor or public settings & building, pole, signboard, bench, escalator, storage tank, streetlight, fountain, barrel, grandstand. \\

\bottomrule
\end{tabular}}
\caption{\textbf{ADE20K Properties} (1 of 7)}
\label{tab:ade_property_1}
\end{table*}

\begin{table*}[t]
\centering
\tablestyle{3pt}{1.0}
\resizebox{\columnwidth}{!}{
\begin{tabular}{c|p{7cm}|p{10cm}}
\shline
Idx & Descriptive Property & Categories with this Property \\
\hline
26 & Covered or surrounded by vegetation or landscaping & field, sky, swimming pool, mountain, grass, dirt track, hill, lake, hovel, path. \\
\hline
27 & Common colors include white, brown, black, gray, metallic tones, green, and red. & swivel chair, rock, tree, chest of drawers, sand, bulletin board, bookcase, desk, door, bar. \\
\hline
28 & Can have decorative elements & mirror, vase, cabinet, ceiling, chest of drawers, column, door, bridge, fireplace, wardrobe. \\
\hline
29 & Can be opened or closed for ventilation & field, truck, oven, windowpane, shower, ceiling, escalator, door, washer, tent. \\
\hline
30 & Often found in various rooms such as living rooms, bedrooms, offices, or recreational areas & pool table, chest of drawers, bookcase, television receiver, bed , ottoman, sofa, armchair, fireplace, wardrobe. \\
\hline
31 & Found in various environments such as nurseries, bedrooms, childcare facilities, kitchens, bathrooms, stores, beaches, laundry rooms, utility areas, dressing areas, and retail spaces & mirror, bottle, toilet, shower, cradle, sink, countertop, counter, washer, towel. \\
\hline
32 & Comes in various sizes, ranging from small to large & airplane, book, rug, chandelier, pot, ball, animal, barrel, sculpture, monitor. \\
\hline
33 & May have multiple levels or floors & building, house, stairs, fountain, step, ship, sidewalk, pillow, floor. \\
\hline
34 & Texture can vary from smooth to rough or textured & rock, curtain, plant, bulletin board, bed , cushion, flag, apparel, floor. \\
\hline
35 & Rectangular shape & truck, bench, windowpane, van, case, bicycle, bookcase, door, sofa. \\
\hline
36 & Supported by legs or a base for stability and balance & swivel chair, stool, base, table, coffee table, ottoman, tent, fireplace, armchair. \\
\hline
37 & Typically has a rectangular, cylindrical, conical, pyramidal, ridged, arched, or curved shape & mountain, skyscraper, canopy, lamp, bag, vase, basket, tent, tower. \\
\hline
38 & Made of various materials & swivel chair, rug, blanket, apparel, curtain, bag, flag, basket, towel. \\
\hline
39 & Can have a decorative design or elements & building, rock, chandelier, trade name, sconce, coffee table, bar, railing. \\
\hline
40 & Can be found in various environments, including urban, suburban, and rural areas & building, road, skyscraper, bus, house, fence, land, tower. \\
\hline
41 & Elevated or raised horizontal surface & ceiling, table, waterfall, desk, step, land, stage, floor. \\
\hline
42 & Generally flat or level & field, rug, chair, base, ceiling, towel, stool, floor. \\
\hline
43 & Can have additional elements or features & swimming pool, van, chest of drawers, table, coffee table, bed , shelf, kitchen island. \\
\hline
44 & Oriented in various positions, including upright, horizontal, and inclined & pole, ashcan, runway, river, storage tank, cradle, person, bar. \\
\hline
45 & Versatile and multi-functional & buffet, table, plate, food, tray, countertop, counter, basket. \\
\hline
46 & Can be found in various environments or settings & trade name, booth, table, curtain, plate, stairway, bar, stage. \\
\hline
47 & Container shape varies, can be cylindrical, rectangular, oval-shaped, or box-shaped & bottle, ashcan, water, storage tank, bathtub, barrel, vase, washer. \\
\hline
48 & Displays visual information or data & poster, signboard, bulletin board, computer, crt screen, screen, grandstand, monitor. \\
\hline
49 & Positioned or mounted in a specific location or orientation & poster, radiator, dishwasher, sink, traffic light, hood, microwave, fireplace. \\
\hline
50 & Commonly found in kitchen or dining areas & buffet, oven, pot, dishwasher, stove, refrigerator, vase, stool. \\
\hline
51 & Structural construction or architectural structure & building, pole, booth, wall, house, fence, bridge. \\
\hline
52 & Common colors include white, beige, gray, black, brown, or various shades & toilet, curtain, wall, plate, countertop, towel, floor. \\
\hline
53 & Surrounded by outdoor spaces or architectural elements & building, swimming pool, skyscraper, pier, column, flag, railing. \\
\hline
54 & Provides shelter or protection & building, car, booth, canopy, hovel, grandstand, tent. \\
\hline
55 & Made of transparent or translucent material & sky, windowpane, water, blind, glass, vase, screen. \\
\hline
56 & Covered with a variety of materials, such as felt cloth, rugs, mats, carpet, fabric, leather, or synthetic materials & book, pool table, person, seat, pillow, stage, floor. \\
\hline
57 & Environmental context in defining the use and significance of various items. & painting, plaything, cushion, clock, counter, apparel, floor. \\
\hline
58 & Often painted in neutral colors & ceiling, storage tank, streetlight, fan, sidewalk, bed , screen door. \\
\hline
59 & Varies in width, height, and depth & ashcan, river, ceiling, storage tank, waterfall, stairs, stairway. \\
\hline
60 & Marked with signs or indicators & booth, road, escalator, bus, traffic light, sidewalk, path. \\
\hline
61 & Rectangular shape with a vertical orientation & radiator, cabinet, chest of drawers, arcade machine, refrigerator, bed , wardrobe. \\
\hline
62 & Found in various settings or environments & windowpane, blind, radiator, house, conveyer belt, glass, seat. \\
\hline
63 & Orientation can vary, but typically upright or inclined & grass, case, escalator, stairs, stairway, flower, path. \\
\hline
64 & Each item has primary physical characteristics that define its nature & book, river, painting, grass, cushion, flag, apparel. \\
\hline
65 & Common features or characteristics & book, painting, grass, animal, person, flag, sculpture. \\

\bottomrule
\end{tabular}}
\caption{\textbf{ADE20K Properties} (2 of 7)}
\label{tab:ade_property_2}
\end{table*}

\begin{table*}[t]
\centering
\tablestyle{3pt}{1.0}
\resizebox{\columnwidth}{!}{
\begin{tabular}{c|p{7cm}|p{10cm}}
\shline
Idx & Descriptive Property & Categories with this Property \\
\hline
66 & Features a hinged or removable lid for access & case, stove, box, bag, door, ottoman, basket. \\
\hline
67 & Typically oriented horizontally & truck, chandelier, car, stove, stage, monitor, boat. \\
\hline
68 & Soft and cushioned for comfort & cradle, cushion, sofa, ottoman, seat, pillow, stool. \\
\hline
69 & Comes in various colors, including white, off-white, black, stainless steel, brown, reddish-brown, and metallic silver. & dirt track, dishwasher, food, bathtub, refrigerator, microwave, washer. \\
\hline
70 & Often used for communication, observation, support, selling, advertising, decoration, information, or structural purposes & pole, poster, trade name, booth, bulletin board, column, tower. \\
\hline
71 & Comes in various shapes and sizes & case, plaything, food, sculpture, glass, flower, apparel. \\
\hline
72 & Features windows or openings for passengers to look out of & building, airplane, car, skyscraper, bus, house. \\
\hline
73 & Made from a variety of materials and can contain flying objects & sky, animal, pillow, cushion, basket, tent. \\
\hline
74 & Supports furniture, such as tables, chairs, or beds & swivel chair, chair, bookcase, sofa, armchair, floor. \\
\hline
75 & Can be heated or cooled & rug, radiator, blanket, sand, food, floor. \\
\hline
76 & Features hanging or wall-mounted light fixtures & chandelier, sconce, ceiling, curtain, streetlight, bathtub. \\
\hline
77 & Reflective or capable of reflecting light & mirror, buffet, light, river, windowpane, earth. \\
\hline
78 & Smooth and soft texture & rug, water, grass, blanket, sand, food. \\
\hline
79 & Often found in natural or scenic environments & rock, mountain, sand, hill, waterfall, hovel. \\
\hline
80 & Various types of seating options & toilet, chair, pool table, desk, sofa, seat. \\
\hline
81 & Can be wall-mounted, freestanding, or suspended from ceiling & mirror, chandelier, bookcase, fan, crt screen, shelf. \\
\hline
82 & Made of various materials such as granite, marble, wood, laminate, porcelain, ceramic, stainless steel, stone, quartz, acrylic, or fiberglass & rock, toilet, bathtub, sink, countertop, counter. \\
\hline
83 & Used for storage, transportation, or carrying of items & case, box, bag, barrel, basket, wardrobe. \\
\hline
84 & Comes in various sizes and designs & poster, trade name, chest of drawers, lamp, basket, boat. \\
\hline
85 & Typically has a horizontal orientation & buffet, awning, swimming pool, bus, bathtub, barrel. \\
\hline
86 & Features handrails or railings for support and safety & bannister, escalator, stairs, stairway, bridge, railing. \\
\hline
87 & Comes in a variety of colors and patterns & blanket, animal, box, bag, flag, ottoman. \\
\hline
88 & Rectangular in shape & poster, book, blanket, flag, screen, towel. \\
\hline
89 & Can have adjustable settings and additional features & radiator, bicycle, dishwasher, fan, conveyer belt, microwave. \\
\hline
90 & Can bear edible fruits, flowers, or nuts & field, tree, plant, palm, flower. \\
\hline
91 & Found in various environments and habitats & tree, grass, plant, palm, flower. \\
\hline
92 & Linear pathway or route for vehicles, pedestrians, hiking, or walking & dirt track, road, hill, sidewalk, path. \\
\hline
93 & Can have a straight, curved, or spiral shape & river, chair, road, stairs, path. \\
\hline
94 & Horizontal structure or furniture piece & bench, bed , seat, grandstand, shelf. \\
\hline
95 & Comes in various sizes and dimensions & shelf, cabinet, box, tray, towel. \\
\hline
96 & Commonly used in kitchens, living rooms, offices, dining rooms, or restaurants & chair, cabinet, coffee table, tray, shelf. \\
\hline
97 & May have visible features or characteristics & rock, storage tank, person, step, kitchen island. \\
\hline
98 & Available in different styles and variations & rug, painting, person, food, wardrobe. \\
\hline
99 & Comes in different variations or types & book, chair, table, counter, sofa. \\
\hline
100 & Equipped with lighting fixtures, sound equipment, and decorative water features & plant, waterfall, fountain, bar, stage. \\
\hline
101 & Upholstered seat and backrest with a raised or curved backrest for support and comfort. & swivel chair, chair, bathtub, seat, armchair. \\
\hline
102 & Upholstered with various materials such as fabric, leather, or synthetic materials & chair, bicycle, ottoman, sofa, armchair. \\
\hline
103 & Often seen in various locations such as roads, highways, construction sites, parking lots, garages, intersections, pedestrian crossings, buildings, temples, and monuments & truck, car, van, column, traffic light. \\
\hline
104 & Found in bodies of water, such as oceans, rivers, lakes, etc. & water, sand, pier, ship, boat. \\
\hline
105 & May have adjustable features or mechanisms & swivel chair, blind, bookcase, monitor, shelf. \\
\hline
106 & Used to cover and decorate floors & buffet, bannister, rug, bench, pier. \\
\bottomrule
\end{tabular}}
\caption{\textbf{ADE20K Properties} (3 of 7)}
\label{tab:ade_property_3}
\end{table*}

\begin{table*}[t]
\centering
\tablestyle{3pt}{1.0}
\resizebox{\columnwidth}{!}{
\begin{tabular}{c|p{7cm}|p{10cm}}
\shline
Idx & Descriptive Property & Categories with this Property \\
\hline
107 & Used for various purposes, including sports, recreation, and agriculture & field, pool table, dirt track, ball, tent. \\
\hline
108 & Tall architectural structure or element & bannister, skyscraper, column, fence, tower. \\
\hline
109 & May have handles or straps for carrying & case, tray, bathtub, bag, basket. \\
\hline
110 & Often painted or coated for protection or durability & base, clock, bridge, boat, railing. \\
\hline
111 & Typically used for warmth, comfort, drying, wiping, or decorative purposes & base, blanket, cushion, pillow, towel. \\
\hline
112 & Features labels, branding, or markings for identification & bottle, trade name, case, box, clock. \\
\hline
113 & Sleek and streamlined shape & airplane, dirt track, ship, path, boat. \\
\hline
114 & Consists of steps or stairs & escalator, stairs, conveyer belt, step, stairway. \\
\hline
115 & Low in height or seat height & minibike, coffee table, hovel, ottoman, stool. \\
\hline
116 & Provides ambient or focal lighting & light, chandelier, sconce, streetlight, bar. \\
\hline
117 & Includes interactive features such as buttons or controls & plaything, dishwasher, arcade machine, microwave, monitor. \\
\hline
118 & Provides visibility and safety during nighttime & wall, sky, runway, streetlight. \\
\hline
119 & Long, narrow strip of land or pavement, often bordered by curbs or sidewalks, and may have boundaries like fences or hedges. & field, sidewalk, road, runway. \\
\hline
120 & Primary physical characteristics of each item. & screen, screen door, television receiver, bed . \\
\hline
121 & Allows light to pass through while controlling airflow & screen door, windowpane, lamp, canopy. \\
\hline
122 & Comes in earthy colors or tones & hovel, mountain, grass, hill. \\
\hline
123 & Used for transportation purposes & sidewalk, van, minibike, conveyer belt. \\
\hline
124 & Used for washing, cleaning, and waste disposal & person, sink, ashcan, toilet. \\
\hline
125 & Varies in shape, elevation, and geological formations & mountain, land, river, earth. \\
\hline
126 & Exhibits diverse colors, including green, brown, and white & palm, land, lake, earth. \\
\hline
127 & Supports diverse marine life and ecosystems & lake, river, sea, earth. \\
\hline
128 & Features a handle or knob for opening and closing & screen door, refrigerator, pot, door. \\
\hline
129 & Can have various door configurations and options & bus, booth, wardrobe, door. \\
\hline
130 & Often used for light control, privacy, or decorative purposes & lamp, sconce, blind, curtain. \\
\hline
131 & Transportation device with two or four wheels, handlebars for steering, and pedals or a small motor for propulsion. & car, bicycle, van, minibike. \\
\hline
132 & Equipped with headlights, taillights, turn signals, and rearview mirrors & car, truck, minibike, van. \\
\hline
133 & Variations in types or models & bus, car, truck, van. \\
\hline
134 & Can exist in different states: solid, liquid, or gas & water, food, fence, ball. \\
\hline
135 & Can be set up or displayed in various indoor or outdoor environments & column, tent, sculpture, painting. \\
\hline
136 & Body of water, can be natural or artificial, and can be saltwater or freshwater & swimming pool, lake, sea, river. \\
\hline
137 & Moves in a linear direction and typically travels in straight lines & conveyer belt, escalator, light, sea. \\
\hline
138 & Can have varying water conditions, shore types, and bottom compositions & sand, lake, sea, river. \\
\hline
139 & Compact and lightweight design, often handheld or portable & mirror, minibike, plaything, cradle. \\
\hline
140 & May have armrests or additional features for convenience & bench, seat, swivel chair, armchair. \\
\hline
141 & Vertical structure with a light source & lamp, streetlight, traffic light, signboard. \\
\hline
142 & Features steps or ramps for access & tent, swimming pool, stage, railing. \\
\hline
143 & Features a lid or cover for containment & counter, ashcan, pot, hood. \\
\hline
144 & Includes shelves, cabinets, or drawers for storage & counter, oven, bar, kitchen island. \\
\bottomrule
\end{tabular}}
\caption{\textbf{ADE20K Properties} (4 of 7)}
\label{tab:ade_property_4}
\end{table*}

\begin{table*}[t]
\centering
\tablestyle{3pt}{1.0}
\resizebox{\columnwidth}{!}{
\begin{tabular}{c|p{7cm}|p{10cm}}
\shline
Idx & Descriptive Property & Categories with this Property \\
\hline
145 & Connected to plumbing system for water supply and drainage & washer, dishwasher, sink, toilet. \\
\hline
146 & Kitchen appliance for cooking or heating food & stove, refrigerator, microwave, oven. \\
\hline
147 & Can accommodate pedestrians, vehicles, trains, a large number of passengers, spectators, and multiple people & bench, bridge, grandstand, bus. \\
\hline
148 & Provides a platform for various purposes (e.g. aircraft takeoff and landing, docking boats or ships) & stairway, pier, runway, ship. \\
\hline
149 & Rectangular shape with a hinged door and a mesh or perforated panel & oven, screen door, microwave, dishwasher. \\
\hline
150 & Water is typically clear or can be murky, with varying textures and levels of calmness. & waterfall, swimming pool, lake, river. \\
\hline
151 & Horizontal structure that spans over a body of water or a valley & pier, bridge, fountain, canopy. \\
\hline
152 & Equipped with wheels or casters for mobility & animal, swivel chair, van, bicycle. \\
\hline
153 & Large, mobile vehicle used for transportation & bus, airplane, truck, van. \\
\hline
154 & Commonly used in electronic devices and commonly features specific ports and inputs & screen, television receiver, glass, crt screen. \\
\hline
155 & Often adorned with decorative elements or embellishments & chandelier, wall, fountain. \\
\hline
156 & May have distinctive features or architectural elements & building, animal, skyscraper. \\
\hline
157 & Exhibits varying levels of brightness and color throughout the day & sky, light, crt screen. \\
\hline
158 & Living organism & tree, animal, plant. \\
\hline
159 & Can have a single or multiple trunks & tree, palm, plant. \\
\hline
160 & Connects different locations or serves as a route & bus, road, path. \\
\hline
161 & Has a crown of large, fan-shaped leaves at the top & palm, flower, grass. \\
\hline
162 & Natural formation or composition & waterfall, grass, rock. \\
\hline
163 & Worn on the body for protection and adornment & person, blanket, apparel. \\
\hline
164 & Composed of diverse elements such as landmasses, water bodies, soil, rocks, vegetation, etc. & animal, land, earth. \\
\hline
165 & Vertical structure connecting different levels & stairway, stairs, door. \\
\hline
166 & Comes in varying sizes and heights & ottoman, mountain, vase. \\
\hline
167 & Used for storing and containing liquids or gases & water, storage tank, pot. \\
\hline
168 & Three-dimensional or flat surface artistic creation & sculpture, poster, painting. \\
\hline
169 & Rectangular shape with flat surfaces and a flat top surface & stove, computer, shelf. \\
\hline
170 & Consists of multiple tiers or levels for organizing items & grandstand, shelf, bulletin board. \\
\hline
171 & Often features intricate designs or patterns & sconce, rug, vase. \\
\hline
172 & Provides support, cushioning, comfort, and relaxation for sitting, lying down, lounging, resting, or sleeping & pillow, cushion, armchair. \\
\hline
173 & Provides security, privacy, or boundary demarcation & shower, booth, fence. \\
\hline
174 & Illumination or ability to emit light & lamp, light, signboard. \\
\hline
175 & Features control knobs, digital display, switch, or knob for temperature and settings & stove, lamp, oven. \\
\hline
176 & Consists of multiple lights with distinct shapes and colors & lamp, traffic light, light. \\
\hline
177 & Variations in design or type & shower, bathtub, toilet. \\
\hline
178 & May have additional features or amenities & hovel, sink, toilet. \\
\hline
179 & Variations in the type of heating or cooking mechanism & stove, fireplace, oven. \\
\hline
180 & Comes in varying sizes and thicknesses & plate, pillow, countertop. \\
\hline
181 & Used for recreational activities such as swimming, boating, fishing, and water sports & lake, boat, river. \\
\hline
182 & Includes features for attaching papers or notes, and may have payment options or a bookmark attached. & book, arcade machine, bulletin board. \\
\hline
183 & Smooth and rounded shape & streetlight, ball, hill. \\
\bottomrule
\end{tabular}}
\caption{\textbf{ADE20K Properties} (5 of 7)}
\label{tab:ade_property_5}
\end{table*}

\begin{table*}[t]
\centering
\tablestyle{3pt}{1.0}
\resizebox{\columnwidth}{!}{
\begin{tabular}{c|p{7cm}|p{10cm}}
\shline
Idx & Descriptive Property & Categories with this Property \\
\hline
184 & Can be freestanding or attached to a building or structure & awning, tower, kitchen island. \\
\hline
185 & Can be connected to other devices and peripherals & computer, clock, monitor. \\
\hline
186 & May have various functionalities, including hour, minute, and second hands, touch-sensitive capabilities, and functionality determined by the operating system and software. & screen, computer, clock. \\
\hline
187 & May have tail fins, stabilizers, interconnected tubes or fins, hull, deck, and superstructure & radiator, airplane, boat. \\
\hline
188 & Can be equipped with various propulsion methods & ship, airplane, boat. \\
\hline
189 & Can be found in arcades, amusement parks, or entertainment venues & arcade machine, bar, plaything. \\
\hline
190 & May have a pointed or rounded top and often has a screw-on or snap-on cap or cork & pole, bottle, tower. \\
\hline
191 & Variations in type or design & clock, fan, tower. \\
\hline
192 & Provides shade and protection from weather elements & tent, awning, canopy. \\
\hline
193 & Variations in screen sizes and resolutions & screen, television receiver, crt screen. \\
\hline
194 & Often colored in shades of gray or black & ashcan, conveyer belt, crt screen. \\
\hline
195 & Commonly used for cooling or ventilation purposes & waterfall, fan, hood. \\
\hline
196 & Earth's atmosphere is expansive and consists of gases like nitrogen and oxygen. & sky, earth. \\
\hline
197 & Visible horizon line separates the sky from the earth & sky, sea. \\
\hline
198 & Bark texture varies and can be rough and textured. & tree, palm. \\
\hline
199 & Provides shade and shelter for animals and humans & tree, palm. \\
\hline
200 & Can be freestanding or built-in & cabinet, wardrobe. \\
\hline
201 & Possesses limbs or appendages with arms, legs, fingers, and toes. & animal, person. \\
\hline
202 & Displays a range of behaviors and communication methods & animal, person. \\
\hline
203 & Spherical shape & ball, earth. \\
\hline
204 & Natural landform with elevation & mountain, hill. \\
\hline
205 & Common colors include clear, green, brown, or opaque & plant, bottle. \\
\hline
206 & Variations in types and orientations & blind, curtain. \\
\hline
207 & Contains liquid substances & water, bottle. \\
\hline
208 & Reflective or shimmering effect & water, sea. \\
\hline
209 & Reflective or light-reflecting & glass, mirror. \\
\hline
210 & Commonly framed or frameless & mirror, poster. \\
\hline
211 & Equipped with a drain and faucet for water flow & shower, bathtub. \\
\hline
212 & Vertical or horizontal barrier & bannister, railing. \\
\hline
213 & Features a secure enclosure and supportive structure for holding a child & base, cradle. \\
\hline
214 & Primarily composed of loose soil or dirt & dirt track, sand. \\
\hline
215 & Composed of organic ingredients and consists of tiny particles & food, sand. \\
\hline
216 & Typically consists of a basin and a central structure & fountain, sink. \\
\hline
217 & Provides a space for performances, with seating for audience members and possible curtains or backdrops & grandstand, stage. \\
\hline
218 & Used for transportation or recreational activities & bicycle, path. \\
\hline
219 & Provides a means of transition between different levels or areas & stairs, step. \\
\hline
220 & Located at airports or military bases & airplane, runway. \\
\hline
221 & Can have retractable or fixed functionality & case, awning. \\
\hline
222 & Common features include sleeves, collars, buttons, zippers, and pockets & pool table, apparel. \\
\hline
223 & Can have accessories like baskets, lights, or fenders & pool table, bicycle. \\
\bottomrule
\end{tabular}}
\caption{\textbf{ADE20K Properties} (6 of 7)}
\label{tab:ade_property_6}
\end{table*}

\begin{table*}[t]
\centering
\tablestyle{3pt}{1.0}
\resizebox{\columnwidth}{!}{
\begin{tabular}{c|p{7cm}|p{10cm}}
\shline
Idx & Descriptive Property & Categories with this Property \\
\hline
224 & Comes in a variety of vibrant colors and eye-catching graphics & flower, arcade machine. \\
\hline
225 & May have scattered rocks, some of which may be wet or moss-covered & waterfall, hill. \\
\hline
226 & May have visible marks or scars & palm, dirt track. \\
\hline
227 & Electronic device for receiving, displaying, processing, and storing information and audiovisual content & television receiver, computer. \\
\hline
228 & Watercraft used for transportation or leisure & ship, boat. \\
\hline
229 & Rough texture & hovel, dirt track. \\
\hline
230 & Overhead structure or covering & awning, canopy. \\
\hline
231 & Often used or seen in outdoor spaces & awning, canopy. \\
\hline
232 & Flat screen with rectangular shape & television receiver, monitor. \\
\hline
233 & Vertical orientation with a narrow neck and wider base & bottle, vase. \\
\hline
234 & Appliance for cleaning tasks & washer, dishwasher. \\
\hline
235 & Variety of bicycle types or models & bicycle, minibike. \\
\hline
236 & Commonly used for cooking or preparing food & food, pot. \\
\hline
237 & Consumed using utensils or hands & plate, food. \\
\hline
238 & Features a showerhead with additional options such as rotating spray arms, handheld showerhead, or additional jets. & shower, dishwasher. \\
\hline
239 & Rectangular shape in electronic devices or displays & monitor, crt screen. \\
\hline
240 & Roughly spherical shape with a slight bulge at the equator & earth. \\
\hline
241 & Rotates on its axis and revolves around the sun & earth. \\
\hline
242 & Interacts with other celestial bodies in the solar system & earth. \\
\hline
243 & Architectural feature or appliance for containing and displaying fire & fireplace. \\
\hline
244 & Often emits a pleasant fragrance & flower. \\
\hline
245 & Includes a keyboard or touchpad for input & computer. \\
\hline
246 & Located within a larger establishment or standalone & bar. \\
\hline
247 & Equipped with a screen for displaying games and is a standalone gaming device & arcade machine. \\
\hline
248 & Offers a variety of game genres & arcade machine. \\
\hline
249 & Features a loading and unloading door & washer. \\
\hline
250 & Commonly associated with children's play or recreational activities & plaything. \\
\hline
251 & Some waterfalls have pools or basins at the bottom. & waterfall. \\
\hline
252 & Bounces when thrown or dropped & ball. \\
\hline
253 & Often attached to a larger object & hood. \\
\hline
254 & Consists of blades arranged in a radial pattern, powered by a motor & fan. \\
\hline
255 & Brittle and fragile & glass. \\
\hline
256 & Timekeeping device. & clock. \\
\bottomrule
\end{tabular}}
\caption{\textbf{ADE20K Properties} (7 of 7)}
\label{tab:ade_property_7}
\end{table*}

\begin{table*}[t]
\centering
\tablestyle{3pt}{1.0}
\resizebox{0.85\columnwidth}{!}{
\begin{tabular}{c|p{4cm}|p{4cm}|p{5cm}}
\toprule
Class A \& Class B & Shared properties & A only properties & B only properties \\
\midrule
car \& bus & 1. The orientation of the object can vary, including vertical, horizontal, diagonal, or inclined; 

2. Rectangular shape with varying dimensions; 

3. Transportation vehicle with varying numbers of wheels; 

4. Smooth and glossy surface; 

5. Variations in type or design; 

6. Features doors, windows, and various levels of access. & 1. Commonly found in various locations such as parking lots, streets, intersections, and designated riding areas; 

2. Common colors include black, white, silver, and red; 

3. Made of various materials such as glass, ceramic, metal, plastic, and wood; 

4. Equipped with common features such as headlights, taillights, turn signals, and mirrors. & 1. Made of various materials, including metal, wood, plastic, glass, ceramic, and composite materials; 

2. Includes various interior features such as seating, storage compartments, and sleeping quarters; 

3. Features various signs, markings, and signals for traffic control and guidance; 

4. Features windows along the fuselage and wings for flight; 

5. Can be found in various environments, including urban, suburban, and rural areas; 

6. Operates on designated routes and schedules, connects different locations or serves as a route, operates on roads or designated bus lanes; 

7. Can accommodate multiple people, pedestrians, vehicles, or trains. \\
\hline
stop sign \& parking meter & 1. Commonly found in various locations such as parking lots, streets, intersections, and designated riding areas; 

2. Rectangular shape with varying dimensions; 

3. Features various signs, markings, and signals for traffic control and guidance; 

4. Typically made of metal, plastic, or bamboo. & 1. The orientation of the object can vary, including vertical, horizontal, diagonal, or inclined; 

2. Often decorated with frosting, icing, vibrant colors, logos, text, graphics, patterns, and designs; 

3. Text is bold and in capital letters. & 1. Comes in various sizes, shapes, designs, and variations; 

2. Often painted or coated for protection or aesthetic purposes; 

3. Vertical structure; 

4. Smooth or textured surface; 

5. Equipped with a display screen and user interaction interface. \\
\hline
dog \& sheep & 1. Has a rounded head with two ears, two eyes, a snout, and a mouth; 

2. Covered in fur, which can vary in color, pattern, length, and texture; 

3. Quadrupedal body shape; 

4. Known for their wool production, meat, intelligence, social behavior, loyalty, and companionship; 

5. Domesticated mammal, commonly kept as a pet. & 1. Comes in a wide range of colors and patterns; 

2. Found in various environments; 

3. Contains diverse species with distinct physical characteristics, behaviors, and temperaments; 

4. Possesses claws or hooves for walking or running; 

5. Expressive facial features. & 1. Often found in groups or herds; 

2. Found in various landscapes and environments; 

3. May have horns, depending on the breed; 

4. Comes in various colors, including brown, tan, black, white, beige, gray, and reddish-brown. \\
\hline
sheep \& cow & 1. Often found in groups or herds; 

2. Covered in fur, which can vary in color, pattern, length, and texture; 

3. Quadrupedal body shape; 

4. Domesticated mammal, commonly kept as a pet; 

5. May have horns, depending on the breed. & 1. Has a rounded head with two ears, two eyes, a snout, and a mouth; 

2. Found in various landscapes and environments; 

3. Known for their wool production, meat, intelligence, social behavior, loyalty, and companionship; 

4. Comes in various colors, including brown, tan, black, white, beige, gray, and reddish-brown. & 1. Covered in vegetation or landscaping; 

2. Elongated or flat shape with various features such as curved tips, rounded backs, or raised lips; 

3. Possesses a tail, which can vary in length and appearance; 

4. Large and distinctive ears or horns; 

5. White or off-white in color, often with white markings or painted in neutral colors. \\
\hline
sheep \& bear & 1. Has a rounded head with two ears, two eyes, a snout, and a mouth; 

2. Covered in fur, which can vary in color, pattern, length, and texture; 

3. Found in various landscapes and environments; 

4. Quadrupedal body shape. & 1. Often found in groups or herds; 

2. Known for their wool production, meat, intelligence, social behavior, loyalty, and companionship; 

3. Domesticated mammal, commonly kept as a pet; 

4. May have horns, depending on the breed; 

5. Comes in various colors, including brown, tan, black, white, beige, gray, and reddish-brown. & 1. Possesses a tail, which can vary in length and appearance; 

2. Contains diverse species with distinct physical characteristics, behaviors, and temperaments; 

3. Mammal with a large, stocky build, distinct trunk, and elongated tusks. \\
\bottomrule
\end{tabular}}
\caption{\textbf{COCO Stuff similar categories} (1 of 6)}
\label{tab:coco_shared_properties_1}
\end{table*}

\begin{table*}[t]
\centering
\tablestyle{3pt}{1.0}
\resizebox{0.85\columnwidth}{!}{
\begin{tabular}{c|p{4cm}|p{4cm}|p{5cm}}
\toprule
baseball bat \& cup & 1. Made of various materials, including metal, wood, plastic, glass, ceramic, and composite materials; 

2. Tapered shape with wider ends and narrower middle; 

3. Smooth and glossy surface; 

4. Equipped with a handle or grip for holding and maneuvering; 

5. Comes in various colors, including brown, tan, black, white, beige, gray, and reddish-brown. & 1. Often used by players on a baseball field; 

2. Used for hitting objects in sports; 

3. Size, weight, and dimensions vary depending on the sport, game, skill level, and age group; 

4. Cylindrical or oval-shaped object with a handle or stem. & 1. Comes in various sizes, shapes, designs, and variations; 

2. May feature decorative patterns, designs, or elements; 

3. Cylindrical or rectangular shape; 

4. Used for stirring, scooping, eating, holding, and drinking food or beverages; 

5. Often found in various culinary settings, including kitchens, restaurants, cafes, and homes. \\
\hline
cup \& scissors & 1. Made of various materials, including metal, wood, plastic, glass, ceramic, and composite materials; 

2. Comes in various sizes, shapes, designs, and variations; 

3. Smooth and glossy surface; 

4. Equipped with a handle or grip for holding and maneuvering; 

5. Often found in various culinary settings, including kitchens, restaurants, cafes, and homes. & 1. May feature decorative patterns, designs, or elements; 

2. Tapered shape with wider ends and narrower middle; 

3. Cylindrical or rectangular shape; 

4. Used for stirring, scooping, eating, holding, and drinking food or beverages; 

5. Comes in various colors, including brown, tan, black, white, beige, gray, and reddish-brown. & 1. Handheld tool used for cutting or oral hygiene; 

2. Commonly used for various purposes such as fashion, functionality, cutting, protection, cushioning, cleaning, and decoration; 

3. Commonly silver, metallic, white, black, or stainless steel in color; 

4. Consists of two blades joined at a pivot point, typically made of metal, sharp and shiny, straight and elongated blades, often has a serrated edge for sawing, thin and elongated blades with sharp edges for cutting. \\
\hline
fork \& spoon & 1. Made of various materials, including metal, wood, plastic, glass, ceramic, and composite materials; 

2. Comes in various variations or varieties; 

3. Commonly silver, metallic, white, black, or stainless steel in color; 

4. Equipped with handles or knobs for opening and closing; 

5. Handheld utensil for eating or serving. & 1. Smooth and glossy surface; 

2. Often found in various culinary settings, including kitchens, restaurants, cafes, and homes; 

3. Fluid and constantly in motion. & 1. Comes in various sizes, shapes, designs, and variations; 

2. Smooth texture with a polished, matte, or glossy finish; 

3. Used for stirring, scooping, eating, holding, and drinking food or beverages; 

4. Highlight a distinctive shape or structural feature central to the function or identity of the item. \\
\hline
apple \& orange & 1. Commonly used in cooking, salads, desserts, and beverages; 

2. Can be eaten raw, cooked, or used in cooking and baking; 

3. Grows in warm climates, low-growing, dense vegetation growth, grows in clusters, seeds contained in the core, grows underground in soil; 

4. Contains juicy and crisp flesh; 

5. Shape of the fruit is spherical or curved/elongated; 

6. Smooth, rounded shape. & 1. Small to medium-sized and typically quadrupedal; 

2. Shows a spectrum of colors and shades; 

3. Found in various environments; 

4. Comes in various variations or varieties; 

5. Typically has a stem and petals; 

6. Can have a polished or matte finish. & 1. Diverse shapes and sizes; 

2. Often associated with vitamin C and citrus fruits; 

3. Contains seeds or pits, although some varieties are seedless; 

4. Bright orange color; 

5. May have a sweet, sour, or tangy taste; 

6. Smooth texture with slight roughness or texture. \\
\hline
sandwich \& hot dog & 1. Food item consisting of layers of ingredients between two slices of bread; 

2. Can be served hot or cold, and can be prepared in various ways (steamed, roasted, stir-fried, grilled, boiled); 

3. Comes in various variations or varieties; 

4. May have fillings or layers with various ingredients. & 1. The shape of the object varies, but it is typically rectangular or has a rectangular shape; 

2. Toasts bread slices; 

3. Served as a dish or food item. & 1. Cylindrical or rectangular shape; 

2. Long and narrow shape with pointed ends or a narrow neck and wider base; 

3. Commonly used in outdoor or recreational settings; 

4. Topped with various condiments and ingredients; 

5. Handheld utensil for eating or serving. \\
\bottomrule
\end{tabular}}
\caption{\textbf{COCO Stuff similar categories} (2 of 6)}
\label{tab:coco_shared_properties_2}
\end{table*}

\begin{table*}[t]
\centering
\tablestyle{3pt}{1.0}
\resizebox{0.85\columnwidth}{!}{
\begin{tabular}{c|p{4cm}|p{4cm}|p{5cm}}
\toprule
sandwich \& salad & 1. Food item consisting of layers of ingredients between two slices of bread; 

2. Can be served hot or cold, and can be prepared in various ways (steamed, roasted, stir-fried, grilled, boiled); 

3. Comes in various variations or varieties; 

4. Served as a dish or food item. & 1. The shape of the object varies, but it is typically rectangular or has a rectangular shape; 

2. May have fillings or layers with various ingredients; 

3. Toasts bread slices. & 1. Shows a spectrum of colors and shades; 

2. Irregular shape or irregularly shaped stones; 

3. Hard and durable composition; 

4. Topped with various condiments and ingredients; 

5. Often dressed with vinaigrettes or creamy dressings. \\
\hline
broccoli \& carrot & 1. Comes in various variations or varieties; 

2. Grows in warm climates, low-growing, dense vegetation growth, grows in clusters, seeds contained in the core, grows underground in soil; 

3. Edible; 

4. Nutrient-rich and high in vitamins and minerals; 

5. Features green leaves. & 1. Can be served hot or cold, and can be prepared in various ways (steamed, roasted, stir-fried, grilled, boiled); 

2. Typically has a stem and petals; 

3. Long and slender shape; 

4. Served as a dish or food item; 

5. Smooth texture with slight roughness or texture. & 1. Commonly used in cooking, salads, desserts, and beverages; 

2. Tapered shape with wider ends and narrower middle; 

3. Can have various colors, including orange, purple, yellow, white, gray, black, or red; 

4. Smooth texture with possible variations in appearance or feel. \\
\hline
potted plant \& bed & 1. The orientation of the object can vary, including vertical, horizontal, diagonal, or inclined; 

2. Texture varies or can range from smooth to rough or different textures; 

3. Specify the primary physical characteristics of each item; 

4. Comes in various sizes, shapes, designs, and variations; 

5. Affected by or influenced by the environmental context. & 1. Shows a spectrum of colors and shades; 

2. Common features and variations in characteristics among different species; 

3. Made of various materials such as porcelain, stainless steel, stone, ceramic, plastic, or clay; 

4. Typically has a stem and petals. & 1. Varies in color, ranging from brown to gray to earth tones; 

2. Rectangular shape; 

3. Includes various interior features such as seating, storage compartments, and sleeping quarters; 

4. Comes in various shapes, sizes, and variations; 

5. Rectangular shape with a backrest and optional armrests; 

6. Made of various materials. \\
\hline
potted plant \& leaves & 1. The orientation of the object can vary, including vertical, horizontal, diagonal, or inclined; 

2. Specify the primary physical characteristics of each item; 

3. Shows a spectrum of colors and shades; 

4. Common features and variations in characteristics among different species. & 1. Texture varies or can range from smooth to rough or different textures; 

2. Comes in various sizes, shapes, designs, and variations; 

3. Made of various materials such as porcelain, stainless steel, stone, ceramic, plastic, or clay; 

4. Typically has a stem and petals; 

5. Affected by or influenced by the environmental context. & 1. Environmental context or surroundings; 

2. Diverse shapes and sizes; 

3. Hard and durable composition. \\
\bottomrule
\end{tabular}}
\caption{\textbf{COCO Stuff similar categories} (3 of 6)}
\label{tab:coco_shared_properties_3}
\end{table*}

\begin{table*}[t]
\centering
\tablestyle{3pt}{1.0}
\resizebox{0.85\columnwidth}{!}{
\begin{tabular}{c|p{4cm}|p{4cm}|p{5cm}}
\toprule
potted plant \& plant-other & 1. The orientation of the object can vary, including vertical, horizontal, diagonal, or inclined; 

2. Texture varies or can range from smooth to rough or different textures; 

3. Specify the primary physical characteristics of each item; 

4. Shows a spectrum of colors and shades; 

5. Common features and variations in characteristics among different species; 

6. Typically has a stem and petals; 

7. Affected by or influenced by the environmental context. & 1. Comes in various sizes, shapes, designs, and variations; 

2. Made of various materials such as porcelain, stainless steel, stone, ceramic, plastic, or clay. & 1. Found in various environments; 

2. Comes in various sizes and shapes, ranging from small to large; 

3. Contains seeds or pits, although some varieties are seedless. \\
\hline
toilet \& sink & 1. Connected to a plumbing system for water supply; 

2. Smooth and glossy surface; 

3. Often found in various environments such as bathrooms, bedrooms, kitchens, offices, etc; 

4. Made of various materials such as porcelain, stainless steel, stone, ceramic, plastic, or clay; 

5. May have additional features or attachments; 

6. Plumbing fixture used for washing or cleaning and sanitary fixture for human waste disposal. & 1. Common colors include white, off-white, beige, gray, brown, green, various shades, clear, opaque, darker shades, metallic tones, black, neutral tones, vibrant patterns, tan, pastel shades; 

2. Highlight a distinctive shape or structural feature central to the function or identity of the item. & 1. The shape of the object varies, but it is typically rectangular or has a rectangular shape; 

2. Positioned horizontally, attached to a countertop or wall. \\
\hline
remote \& cell phone & 1. Rectangular shape with varying dimensions; 

2. Contains buttons or controls for navigation and additional functions; 

3. Lightweight; 

4. Handheld or portable device; 

5. Smooth or textured surface; 

6. Can be used to control various devices. & 1. Location or context of usage; 

2. Typically made of metal, plastic, or bamboo; 

3. Commonly silver, metallic, white, black, or stainless steel in color; 

4. Features buttons and controls for various functions. & 1. Made of various materials, including metal, wood, plastic, glass, ceramic, and composite materials; 

2. Comes in various sizes, shapes, designs, and variations; 

3. Can be connected to a computer or other electronic devices, often accompanied by buttons or a remote control, enables voice calls, messaging, internet browsing, and app usage, can be analog or digital; 

4. Common colors include black, white, silver, and red; 

5. Electronic device with a screen for displaying visual content and a camera for capturing photos and videos; 

6. Equipped with a display screen and user interaction interface. \\
\bottomrule
\end{tabular}}
\caption{\textbf{COCO Stuff similar categories} (4 of 6)}
\label{tab:coco_shared_properties_4}
\end{table*}

\begin{table*}[t]
\centering
\tablestyle{3pt}{1.0}
\resizebox{0.85\columnwidth}{!}{
\begin{tabular}{c|p{4cm}|p{4cm}|p{5cm}}
\toprule

microwave \& refrigerator & 1. Rectangular shape with varying dimensions; 

2. Typically made of metal, plastic, or bamboo; 

3. Kitchen appliance for cooking, baking, heating, or cooling food; 

4. Smooth and glossy surface; 

5. Commonly silver, metallic, white, black, or stainless steel in color. & 1. Contains buttons or controls for navigation and additional functions; 

2. Positioned horizontally, attached to a countertop or wall; 

3. Features control knobs, digital display, buttons, switches, and additional features; 

4. Highlight a distinctive shape or structural feature central to the function or identity of the item. & 1. Contains compartments, pockets, or levels for organization; 

2. Features adjustable temperature settings, chimneys, vents, skylights, tilt function, wind vents, nozzle for directing airflow, windows or vents for ventilation; 

3. Equipped with handles or knobs for opening and closing; 

4. Often found in various culinary settings, including kitchens, restaurants, cafes, and homes; 

5. Contains racks, shelves, drawers, or compartments for organization and storage. \\
\hline
book \& clothes & 1. Texture varies or can range from smooth to rough or different textures; 

2. Specify the primary physical characteristics of each item; 

3. Comes in various sizes, shapes, designs, and variations; 

4. Comes in various sizes and shapes, ranging from small to large; 

5. Common features and additional elements or details. & 1. The orientation of the object can vary, including vertical, horizontal, diagonal, or inclined; 

2. Rectangular shape with varying dimensions; 

3. Location or context of usage; 

4. Contains compartments, pockets, or levels for organization; 

5. Covered or enclosed in a protective material. & 1. Can have various patterns or designs; 

2. Affected by or influenced by the environmental context; 

3. Made from a variety of materials. \\
\hline
branch \& plant-other & 1. The orientation of the object can vary, including vertical, horizontal, diagonal, or inclined; 

2. Texture varies or can range from smooth to rough or different textures; 

3. Shows a spectrum of colors and shades; 

4. Found in various environments; 

5. Contains seeds or pits, although some varieties are seedless; 

6. Typically has a stem and petals. & 1. Elongated or flat shape with various features such as curved tips, rounded backs, or raised lips; 

2. Covered in leaves or needles; 

3. Comes in various sizes, thicknesses, lengths, widths, and depths; 

4. Made of various materials. & 1. Specify the primary physical characteristics of each item; 

2. Comes in various sizes and shapes, ranging from small to large; 

3. Common features and variations in characteristics among different species; 

4. Affected by or influenced by the environmental context. \\
\hline
cabinet \& cupboard & 1. Rectangular shape with varying dimensions; 

2. Freestanding or built-in storage capability; 

3. Often found in various environments such as bathrooms, bedrooms, kitchens, offices, etc; 

4. Equipped with handles or knobs for opening and closing; 

5. Contains racks, shelves, drawers, or compartments for organization and storage. & 1. Made of various materials, including metal, wood, plastic, glass, ceramic, and composite materials; 

2. Comes in various sizes and shapes, ranging from small to large; 

3. Can have different finishes or appearances; 

4. Smooth or textured surface. & 1. Comes in various sizes, shapes, designs, and variations; 

2. Common colors include white, off-white, beige, gray, brown, green, various shades, clear, opaque, darker shades, metallic tones, black, neutral tones, vibrant patterns, tan, pastel shades; 

3. Features doors, zippered entrance, and hinged lid; 

4. Smooth or textured surface, depending on the material or type of stone; 

5. Made of various materials. \\
\bottomrule
\end{tabular}}
\caption{\textbf{COCO Stuff similar categories} (5 of 6)}
\label{tab:coco_shared_properties_5}
\end{table*}

\begin{table*}[t]
\centering
\tablestyle{3pt}{1.0}
\resizebox{0.85\columnwidth}{!}{
\begin{tabular}{c|p{4cm}|p{4cm}|p{5cm}}
\toprule
dirt \& gravel & 1. Texture varies or can range from smooth to rough or different textures; 

2. Found in various environments; 

3. Composed of various materials such as soil, sand, rocks, organic matter, and debris; 

4. Loose or compacted, depending on moisture content and usage; 

5. Irregular shape or irregularly shaped stones. & 1. Varies in color, ranging from brown to gray to earth tones; 

2. Covered in vegetation or landscaping. & 1. Made of natural or durable materials such as marble, granite, wood, or laminate; 

2. Positioned along edges or at ground level; 

3. Common colors include white, off-white, beige, gray, brown, green, various shades, clear, opaque, darker shades, metallic tones, black, neutral tones, vibrant patterns, tan, pastel shades; 

4. Can be used for advertising, promotion, construction, furniture, decoration, landscaping, or functional purposes. \\
\hline
hill \& mountain & 1. Covered in vegetation or landscaping; 

2. Shows a spectrum of colors and shades; 

3. Found in various landscapes and environments; 

4. Natural landform with sloping sides, often steep or rugged, typically elevated and rocky; 

5. Peaks or summits at the highest points. & 1. Composed of various materials such as soil, sand, rocks, organic matter, and debris; 

2. Smooth, rounded shape. & 1. Surrounded by other buildings or open spaces; 

2. Can have various shapes and designs. \\
\hline
railroad \& road & 1. Features various signs, markings, and signals for traffic control and guidance; 

2. Positioned along edges or at ground level; 

3. Can be found in various environments, including urban, suburban, and rural areas; 

4. Can accommodate multiple people, pedestrians, vehicles, or trains. & 1. Mode of transportation on tracks; 

2. Fluid and constantly in motion; 

3. Made of metal rails supported by wooden or concrete ties. & 1. Can be affected by weather conditions; 

2. Constructed from various materials such as plaster, wood, metal, concrete, steel, fiberglass, glass, brick, asphalt, or stone; 

3. Operates on designated routes and schedules, connects different locations or serves as a route, operates on roads or designated bus lanes; 

4. Can have various shapes and designs; 

5. Smooth, rounded shape. \\
\bottomrule
\end{tabular}}
\caption{\textbf{COCO Stuff similar categories} (6 of 6)}
\label{tab:coco_shared_properties_6}
\end{table*}

\begin{table*}[t]
\centering
\tablestyle{3pt}{1.0}
\resizebox{0.85\columnwidth}{!}{
\begin{tabular}{c|p{4cm}|p{4cm}|p{5cm}}
\toprule
Class A \& Class B & Shared properties & Class A only properties & Class B only properties \\
\midrule
building \& skyscraper & 1. Surrounded by outdoor spaces or architectural elements; 

2. Can be found in various environments, including urban, suburban, and rural areas; 

3. Constructed with sturdy and durable materials; 

4. May have distinctive features or architectural elements; 

5. Orientation can vary, either vertically or horizontally; 

6. Features windows or openings for passengers to look out of. & 1. Can have a decorative design or elements; 

2. Rectangular or square shape with straight edges and corners; 

3. Often located in outdoor or public settings; 

4. Provides shelter or protection; 

5. Structural construction or architectural structure; 

6. May have multiple levels or floors. & 1. Varies in shape, size, height, length, and design; 

2. Provides scenic views and is surrounded by natural elements or landscapes; 

3. Tall architectural structure or element; 

4. Typically has a rectangular, cylindrical, conical, pyramidal, ridged, arched, or curved shape; 

5. Smooth or textured surface. \\
\hline
tree \& plant & 1. Found in various environments and habitats; 

2. Living organism; 

3. Can bear edible fruits, flowers, or nuts; 

4. Can have a single or multiple trunks. & 1. Common colors include white, brown, black, gray, metallic tones, green, and red; 

2. Provides shade and shelter for animals and humans; 

3. Bark texture varies and can be rough and textured. & 1. Varies in shape, size, height, length, and design; 

2. Common colors include clear, green, brown, or opaque; 

3. Equipped with lighting fixtures, sound equipment, and decorative water features; 

4. Texture can vary from smooth to rough or textured. \\
\hline
tree \& palm & 1. Provides shade and shelter for animals and humans; 

2. Found in various environments and habitats; 

3. Can bear edible fruits, flowers, or nuts; 

4. Bark texture varies and can be rough and textured; 

5. Can have a single or multiple trunks. & 1. Common colors include white, brown, black, gray, metallic tones, green, and red; 

2. Living organism. & 1. Has a crown of large, fan-shaped leaves at the top; 

2. Exhibits diverse colors, including green, brown, and white; 

3. May have visible marks or scars. \\
\hline
road \& sidewalk & 1. Marked with signs or indicators; 

2. Constructed with sturdy and durable materials; 

3. Flat and smooth surface; 

4. Linear pathway or route for vehicles, pedestrians, hiking, or walking; 

5. Long, narrow strip of land or pavement, often bordered by curbs or sidewalks, and may have boundaries like fences or hedges. & 1. Can be found in various environments, including urban, suburban, and rural areas; 

2. Connects different locations or serves as a route; 

3. Can have a straight, curved, or spiral shape. & 1. Typically has a rectangular or circular shape; 

2. Often painted in neutral colors; 

3. Used for transportation purposes; 

4. May have multiple levels or floors. \\
\hline
road \& path & 1. Marked with signs or indicators; 

2. Constructed with sturdy and durable materials; 

3. Linear pathway or route for vehicles, pedestrians, hiking, or walking; 

4. Connects different locations or serves as a route; 

5. Can have a straight, curved, or spiral shape. & 1. Can be found in various environments, including urban, suburban, and rural areas; 

2. Flat and smooth surface; 

3. Long, narrow strip of land or pavement, often bordered by curbs or sidewalks, and may have boundaries like fences or hedges. & 1. Covered or surrounded by vegetation or landscaping; 

2. Sleek and streamlined shape; 

3. Orientation can vary, but typically upright or inclined; 

4. Used for transportation or recreational activities. \\
\bottomrule
\end{tabular}}
\caption{\textbf{ADE20K similar categories} (1 of 6)}
\label{tab:ade_shared_properties_1}
\end{table*}

\begin{table*}[t]
\centering
\tablestyle{3pt}{1.0}
\resizebox{0.85\columnwidth}{!}{
\begin{tabular}{c|p{4cm}|p{4cm}|p{5cm}}
\toprule
cabinet \& wardrobe & 1. Surface texture can vary, ranging from smooth to textured; 

2. Contains drawers, compartments, or shelves for storage; 

3. Can be freestanding or built-in; 

4. Can have decorative elements; 

5. Rectangular shape with a vertical orientation. & 1. Made of various materials, including metal, wood, stone, concrete, plastic, and glass; 

2. Comes in various sizes and dimensions; 

3. Features various elements and additional features; 

4. Commonly used in kitchens, living rooms, offices, dining rooms, or restaurants. & 1. Varies in shape, size, height, length, and design; 

2. Constructed with sturdy and durable materials; 

3. Often found in various rooms such as living rooms, bedrooms, offices, or recreational areas; 

4. Can have various door configurations and options; 

5. Available in different styles and variations; 

6. Used for storage, transportation, or carrying of items. \\
\hline
mountain \& hill & 1. Provides scenic views and is surrounded by natural elements or landscapes; 

2. Covered or surrounded by vegetation or landscaping; 

3. Often found in natural or scenic environments; 

4. Comes in earthy colors or tones; 

5. Natural landform with elevation. & 1. Comes in varying sizes and heights; 

2. Typically has a rectangular, cylindrical, conical, pyramidal, ridged, arched, or curved shape; 

3. Varies in shape, elevation, and geological formations. & 1. Smooth and rounded shape; 

2. Linear pathway or route for vehicles, pedestrians, hiking, or walking; 

3. May have scattered rocks, some of which may be wet or moss-covered. \\
\hline
painting \& sculpture & 1. Surface texture can vary, ranging from smooth to textured; 

2. Common features or characteristics; 

3. Can be painted or adorned with various colors, finishes, logos, text, or images; 

4. Can be set up or displayed in various indoor or outdoor environments; 

5. Three-dimensional or flat surface artistic creation. & 1. Typically has a rectangular or circular shape; 

2. Varies in color, ranging from neutral tones to vibrant hues; 

3. Each item has primary physical characteristics that define its nature; 

4. Environmental context in defining the use and significance of various items;

5. Orientation can vary, either vertically or horizontally; 

6. Available in different styles and variations. & 1. Made of various materials, including metal, wood, stone, concrete, plastic, and glass; 

2. May have decorative patterns or designs; 

3. Comes in various sizes, ranging from small to large; 

4. Comes in various shapes and sizes. \\
\hline
cushion \& apparel & 1. Varies in shape, size, height, length, and design; 

2. Varies in color, ranging from neutral tones to vibrant hues; 

3. Each item has primary physical characteristics that define its nature; 

4. Environmental context in defining the use and significance of various items; 

5. Texture can vary from smooth to rough or textured; 

6. Orientation can vary, either vertically or horizontally. & 1. Typically has a rectangular or circular shape; 

2. Provides support, cushioning, comfort, and relaxation for sitting, lying down, lounging, resting, or sleeping; 

3. Made from a variety of materials and can contain flying objects; 

4. Typically used for warmth, comfort, drying, wiping, or decorative purposes; 

5. Soft and cushioned for comfort. & 1. May have decorative patterns or designs; 

2. Worn on the body for protection and adornment; 

3. Made of various materials; 

4. Comes in various shapes and sizes; 

5. Common features include sleeves, collars, buttons, zippers, and pockets. \\
\hline
chest of drawers \& bookcase & 1. Made of various materials, including metal, wood, stone, concrete, plastic, and glass; 

2. Common colors include white, brown, black, gray, metallic tones, green, and red; 

3. Contains drawers, compartments, or shelves for storage; 

4. Often found in various rooms such as living rooms, bedrooms, offices, or recreational areas; 

5. Smooth or textured surface. & 1. Can have decorative elements; 

2. Comes in various sizes and designs; 

3. Rectangular shape with a vertical orientation; 

4. Can have additional elements or features. & 1. Varies in shape, size, height, length, and design; 

2. May have adjustable features or mechanisms; 

3. Supports furniture, such as tables, chairs, or beds; 

4. Rectangular shape; 

5. Can be wall-mounted, freestanding, or suspended from ceiling. \\
\bottomrule
\end{tabular}}
\caption{\textbf{ADE20K similar categories} (2 of 6)}
\label{tab:ade_shared_properties_2}
\end{table*}

\begin{table*}[t]
\centering
\tablestyle{3pt}{1.0}
\resizebox{0.85\columnwidth}{!}{
\begin{tabular}{c|p{4cm}|p{4cm}|p{5cm}}
\toprule
counter \& countertop & 1. Smooth texture, often with a polished, glossy, or matte finish; 

2. Versatile and multi-functional; 

3. Made of various materials such as granite, marble, wood, laminate, porcelain, ceramic, stainless steel, stone, quartz, acrylic, or fiberglass; 

4. Rectangular or square shape with straight edges and corners; 

5. Found in various environments such as nurseries, bedrooms, childcare facilities, kitchens, bathrooms, stores, beaches, laundry rooms, utility areas, dressing areas, and retail spaces. & 1. Includes shelves, cabinets, or drawers for storage; 

2. Environmental context in defining the use and significance of various items; 

3. Common colors include white, black, red, blue, yellow, gray, metallic finishes, silver, gold, natural wood tones, pastels, and brown; 

4. Features a lid or cover for containment; 

5. Comes in different variations or types. & 1. Contains various interior features and amenities; 

2. May have decorative patterns or designs; 

3. Common colors include white, beige, gray, black, brown, or various shades; 

4. Comes in varying sizes and thicknesses. \\
\hline
sink \& toilet & 1. Made of various materials such as granite, marble, wood, laminate, porcelain, ceramic, stainless steel, stone, quartz, acrylic, or fiberglass; 

2. Smooth and glossy surface; 

3. Connected to plumbing system for water supply and drainage; 

4. Found in various environments such as nurseries, bedrooms, childcare facilities, kitchens, bathrooms, stores, beaches, laundry rooms, utility areas, dressing areas, and retail spaces; 

5. Used for washing, cleaning, and waste disposal; 

6. May have additional features or amenities. & 1. Positioned or mounted in a specific location or orientation; 

2. Typically consists of a basin and a central structure. & 1. Common colors include white, beige, gray, black, brown, or various shades; 

2. Variations in design or type; 

3. Various types of seating options. \\
\hline
refrigerator \& oven & 1. Contains various interior features and amenities; 

2. Commonly found in kitchen or dining areas; 

3. Smooth and glossy surface; 

4. Kitchen appliance for cooking or heating food; 

5. Typically made of various materials, including metal, ceramic, glass, plastic, and composite materials. & 1. Contains drawers, compartments, or shelves for storage; 

2. Comes in various colors, including white, off-white, black, stainless steel, brown, reddish-brown, and metallic silver; 

3. Features a handle or knob for opening and closing; 

4. Rectangular shape with a vertical orientation. & 1. Can be opened or closed for ventilation; 

2. Features control knobs, digital display, switch, or knob for temperature and settings; 

3. Includes shelves, cabinets, or drawers for storage; 

4. Common colors include white, black, red, blue, yellow, gray, metallic finishes, silver, gold, natural wood tones, pastels, and brown; 

5. Variations in the type of heating or cooking mechanism; 

6. Rectangular shape with a hinged door and a mesh or perforated panel. \\
\hline
refrigerator \& microwave & 1. Contains various interior features and amenities; 

2. Smooth and glossy surface; 

3. Comes in various colors, including white, off-white, black, stainless steel, brown, reddish-brown, and metallic silver; 

4. Kitchen appliance for cooking or heating food; 

5. Typically made of various materials, including metal, ceramic, glass, plastic, and composite materials. & 1. Commonly found in kitchen or dining areas; 

2. Contains drawers, compartments, or shelves for storage; 

3. Features a handle or knob for opening and closing; 

4. Rectangular shape with a vertical orientation. & 1. Positioned or mounted in a specific location or orientation; 

2. Includes interactive features such as buttons or controls; 

3. Can have adjustable settings and additional features; 

4. Rectangular shape with a hinged door and a mesh or perforated panel. \\
\bottomrule
\end{tabular}}
\caption{\textbf{ADE20K similar categories} (3 of 6)}
\label{tab:ade_shared_properties_3}
\end{table*}

\begin{table*}[t]
\centering
\tablestyle{3pt}{1.0}
\resizebox{0.85\columnwidth}{!}{
\begin{tabular}{c|p{4cm}|p{4cm}|p{5cm}}
\toprule

stairs \& stairway & 1. Features handrails or railings for support and safety; 

2. Constructed with sturdy and durable materials; 

3. Vertical structure connecting different levels; 

4. Varies in width, height, and depth; 

5. Consists of steps or stairs; 

6. Orientation can vary, but typically upright or inclined; 

7. Commonly found in various environments such as homes, offices, public spaces, and buildings. & 1. Provides a means of transition between different levels or areas; 

2. Can have a straight, curved, or spiral shape; 

3. May have multiple levels or floors. & 1. Provides a platform for various purposes (eg aircraft takeoff and landing, docking boats or ships); 

2. Smooth or textured surface; 

3. Can be found in various environments or settings. \\
\hline
river \& lake & 1. Provides scenic views and is surrounded by natural elements or landscapes; 

2. Used for recreational activities such as swimming, boating, fishing, and water sports; 

3. Water is typically clear or can be murky, with varying textures and levels of calmness; 

4. Supports diverse marine life and ecosystems; 

5. Body of water, can be natural or artificial, and can be saltwater or freshwater; 

6. Can have varying water conditions, shore types, and bottom compositions. & 1. Each item has primary physical characteristics that define its nature; 

2. Oriented in various positions, including upright, horizontal, and inclined; 

3. Varies in width, height, and depth; 

4. Reflective or capable of reflecting light; 

5. Varies in shape, elevation, and geological formations; 

6. Can have a straight, curved, or spiral shape. & 1. Covered or surrounded by vegetation or landscaping; 

2. Irregular or varying shape; 

3. Exhibits diverse colors, including green, brown, and white. \\
\hline
stove \& oven & 1. Commonly found in kitchen or dining areas; 

2. Features control knobs, digital display, switch, or knob for temperature and settings; 

3. Kitchen appliance for cooking or heating food; 

4. Common colors include white, black, red, blue, yellow, gray, metallic finishes, silver, gold, natural wood tones, pastels, and brown; 

5. Typically made of various materials, including metal, ceramic, glass, plastic, and composite materials; 

6. Variations in the type of heating or cooking mechanism. & 1. Typically oriented horizontally; 

2. Features a hinged or removable lid for access; 

3. Rectangular shape with flat surfaces and a flat top surface; 

4. Smooth or textured surface. & 1. Contains various interior features and amenities; 

2. Can be opened or closed for ventilation; 

3. Includes shelves, cabinets, or drawers for storage; 

4. Smooth and glossy surface; 

5. Rectangular shape with a hinged door and a mesh or perforated panel. \\
\hline
boat \& ship & 1. Watercraft used for transportation or leisure; 

2. Constructed with sturdy and durable materials; 

3. Sleek and streamlined shape; 

4. Can be equipped with various propulsion methods; 

5. Found in bodies of water, such as oceans, rivers, lakes, etc. & 1. Contains various interior features and amenities; 

2. Typically oriented horizontally; 

3. Used for recreational activities such as swimming, boating, fishing, and water sports; 

4. Often painted or coated for protection or durability; 

5. Comes in various sizes and designs; 

6. May have tail fins, stabilizers, interconnected tubes or fins, hull, deck, and superstructure. & 1. Contains drawers, compartments, or shelves for storage; 

2. Provides a platform for various purposes (eg aircraft takeoff and landing, docking boats or ships); 

3. Can be painted or adorned with various colors, finishes, logos, text, or images; 

4. May have multiple levels or floors. \\
\bottomrule
\end{tabular}}
\caption{\textbf{ADE20K similar categories} (4 of 6)}
\label{tab:ade_shared_properties_4}
\end{table*}

\begin{table*}[t]
\centering
\tablestyle{3pt}{1.0}
\resizebox{0.85\columnwidth}{!}{
\begin{tabular}{c|p{4cm}|p{4cm}|p{5cm}}
\toprule
truck \& van & 1. Made of various materials, including metal, wood, stone, concrete, plastic, and glass; 

2. Often seen in various locations such as roads, highways, construction sites, parking lots, garages, intersections, pedestrian crossings, buildings, temples, and monuments; 

3. Equipped with headlights, taillights, turn signals, and rearview mirrors; 

4. Variations in types or models; 

5. Rectangular shape; 

6. Large, mobile vehicle used for transportation. & 1. Surface texture can vary, ranging from smooth to textured; 

2. Can be opened or closed for ventilation; 

3. Typically oriented horizontally. & 1. Transportation device with two or four wheels, handlebars for steering, and pedals or a small motor for propulsion; 

2. Used for transportation purposes; 

3. Can be painted or adorned with various colors, finishes, logos, text, or images; 

4. Equipped with wheels or casters for mobility; 

5. Can have additional elements or features. \\
\hline
basket \& bag & 1. May have handles or straps for carrying; 

2. Typically has a rectangular, cylindrical, conical, pyramidal, ridged, arched, or curved shape; 

3. Features a hinged or removable lid for access; 

4. Made of various materials; 

5. Used for storage, transportation, or carrying of items. & 1. Smooth texture, often with a polished, glossy, or matte finish; 

2. Versatile and multi-functional; 

3. Made from a variety of materials and can contain flying objects; 

4. Comes in various sizes and designs. & 1. Surface texture can vary, ranging from smooth to textured; 

2. Comes in a variety of colors and patterns; 

3. Contains drawers, compartments, or shelves for storage. \\
\hline
oven \& microwave & 1. Contains various interior features and amenities; 

2. Smooth and glossy surface; 

3. Kitchen appliance for cooking or heating food; 

4. Typically made of various materials, including metal, ceramic, glass, plastic, and composite materials; 

5. Rectangular shape with a hinged door and a mesh or perforated panel. & 1. Can be opened or closed for ventilation; 

2. Commonly found in kitchen or dining areas; 

3. Features control knobs, digital display, switch, or knob for temperature and settings; 

4. Includes shelves, cabinets, or drawers for storage; 

5. Common colors include white, black, red, blue, yellow, gray, metallic finishes, silver, gold, natural wood tones, pastels, and brown; 

6. Variations in the type of heating or cooking mechanism. & 1. Positioned or mounted in a specific location or orientation; 

2. Comes in various colors, including white, off-white, black, stainless steel, brown, reddish-brown, and metallic silver; 

3. Includes interactive features such as buttons or controls; 

4. Can have adjustable settings and additional features. \\
\hline
microwave \& dishwasher & 1. Positioned or mounted in a specific location or orientation; 

2. Smooth and glossy surface; 

3. Comes in various colors, including white, off-white, black, stainless steel, brown, reddish-brown, and metallic silver; 

4. Includes interactive features such as buttons or controls; 

5. Typically made of various materials, including metal, ceramic, glass, plastic, and composite materials; 

6. Can have adjustable settings and additional features; 

7. Rectangular shape with a hinged door and a mesh or perforated panel. & 1. Contains various interior features and amenities; 

2. Kitchen appliance for cooking or heating food. & 1. Commonly found in kitchen or dining areas; 

2. Features a showerhead with additional options such as rotating spray arms, handheld showerhead, or additional jets; 

3. Contains drawers, compartments, or shelves for storage; 

4. Appliance for cleaning tasks; 

5. Connected to plumbing system for water supply and drainage. \\
\bottomrule
\end{tabular}}
\caption{\textbf{ADE20K similar categories} (5 of 6)}
\label{tab:ade_shared_properties_5}
\end{table*}

\begin{table*}[t]
\centering
\tablestyle{3pt}{1.0}
\resizebox{0.85\columnwidth}{!}{
\begin{tabular}{c|p{4cm}|p{4cm}|p{5cm}}
\toprule
screen \& crt screen & 1. Constructed or made from various materials such as wood, metal, glass, plastic, fabric, vinyl, etc; 

2. Displays visual information or data; 

3. Flat and smooth surface; 

4. Variations in screen sizes and resolutions; 

5. Commonly used in electronic devices and commonly features specific ports and inputs. & 1. Made of transparent or translucent material; 

2. Rectangular in shape; 

3. Orientation can vary, either vertically or horizontally; 

4. Primary physical characteristics of each item; 

5. May have various functionalities, including hour, minute, and second hands, touch-sensitive capabilities, and functionality determined by the operating system and software. & 1. Exhibits varying levels of brightness and color throughout the day; 

2. Rectangular shape in electronic devices or displays; 

3. Can be wall-mounted, freestanding, or suspended from ceiling; 

4. Often colored in shades of gray or black. \\
\bottomrule
\end{tabular}}
\caption{\textbf{ADE20K similar categories} (6 of 6)}
\label{tab:ade_shared_properties_6}
\end{table*}
\end{document}

%% file: 00_abstract.tex
\begin{abstract}

We introduce \textbf{ProLab}, a novel approach using \textbf{pro}perty-level \textbf{lab}el space for creating strong interpretable segmentation models. Instead of relying solely on category-specific annotations, ProLab uses descriptive properties grounded in common sense knowledge for supervising segmentation models.  It is based on two core designs. First, we employ Large Language Models (LLMs) and carefully crafted prompts to generate descriptions of all involved categories that carry meaningful common sense knowledge and follow a structured format. Second, we introduce a description embedding model preserving semantic correlation across descriptions and then cluster them into a set of descriptive properties (\eg, 256) using K-Means. These properties are based on interpretable common sense knowledge consistent with theories of human recognition.  We empirically show that our approach makes segmentation models perform stronger on five classic benchmarks (\eg, ADE20K, COCO-Stuff, Pascal Context, Cityscapes and BDD). Our method also shows better scalability with extended training steps than category-level supervision. Our interpretable segmentation framework also emerges with the generalization ability to segment out-of-domain or unknown categories using in-domain descriptive properties. Code is available at  \url{https://github.com/lambert-x/ProLab}.

\end{abstract}

%% file: 01_intro.tex
\begin{figure}[t]
\centering
\includegraphics[width=\linewidth]{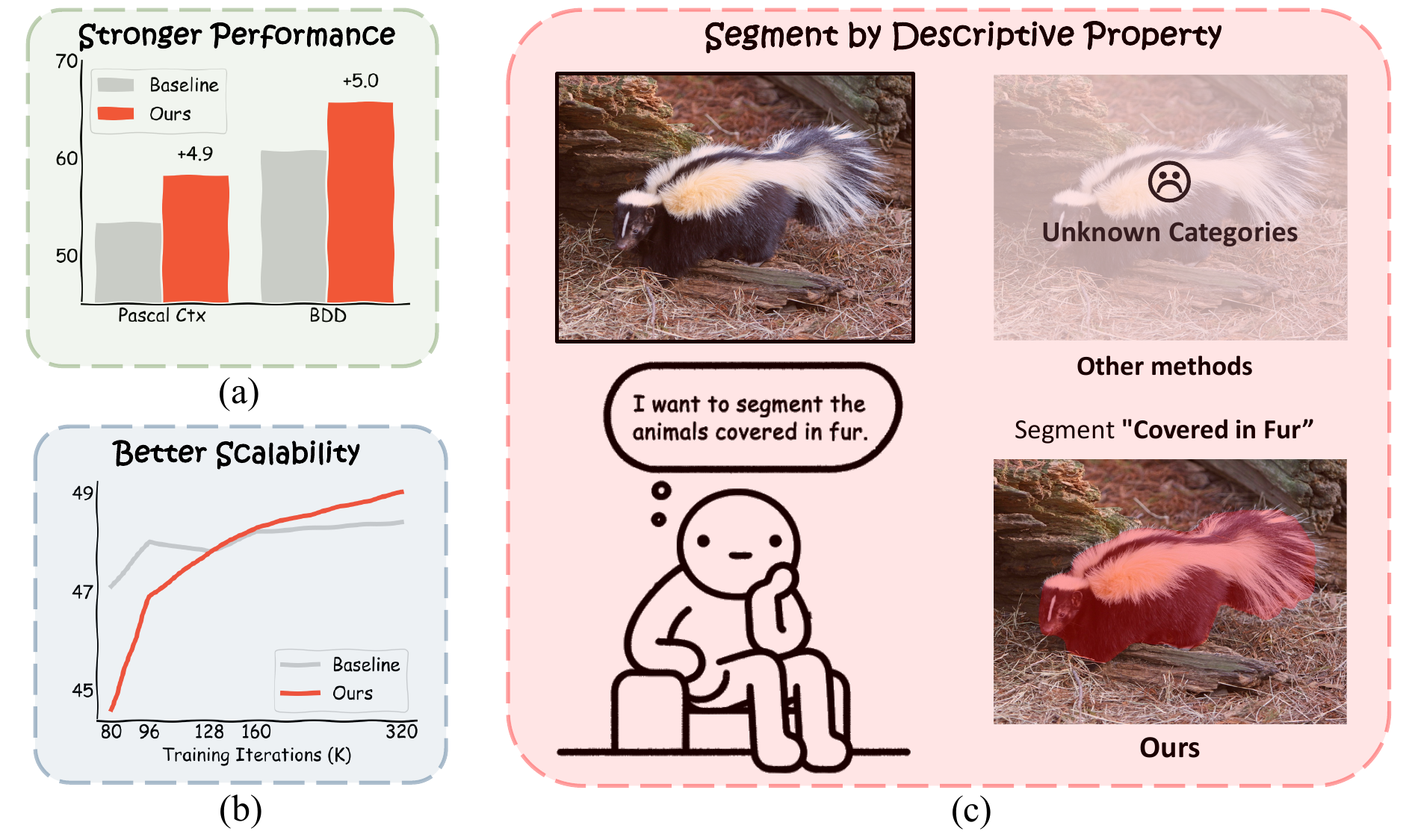}
\caption{\textbf{Advantages of ProLab.} Compared to classic category-level label space, ProLab improves segmentation models in three aspects: (a) \textbf{stronger performance} on classic segmentation benchmarks (b) \textbf{better scalability} with extended training steps (c) ability to \textbf{segment by descriptive properties}, which could generalize to out-of-domain categories or even unknown categories.}
\label{fig:teaser}
\end{figure}

\section{Introduction}
\label{sec:intro}

Semantic segmentation is widely used in many real-world applications such as autonomous driving~\cite{cordts2016cityscapes, geiger2013kitti, yu2020bdd100k}, scene understanding ~\cite{zhou2017ade, lin2014coco, everingham2010pascal, mottaghi2014pascalcontext}, and medical image analysis~\cite{roth2015deeporgan, liang2018dynamic, havaei2017brain}. Seminal works in this domain include ~\cite{long2015fully, ronneberger2015u, chen2017deeplab, xiao2018unified, xie2021segformer}, all of which have significantly advanced the field with their innovative architectures and strategies.

Despite their advanced design, models like DeepLab~\cite{chen2017deeplab}, UperNet~\cite{xiao2018unified}, SegFormer~\cite{xie2021segformer}, and Vision Perceiver~\cite{chen2023vision} use a one-hot label space for categories, lacking inter-category semantic correlations. Attempts to address this, such as manual category merging~\cite{lambert2020mseg} or modeling hierarchical label relationships~\cite{li2022deep}, often result in performance drops and scalability challenges, exacerbated by expanding data and semantic spaces.
On the other hand, recent works~\cite{li2022language, zhou2023lmseg} have addressed label space issues by leveraging language embeddings from CLIP~\cite{radford2021clip} for constructing label spaces. However, methods that use CLIP to model inter-class embeddings often struggle with human interpretability. This is primarily because CLIP, despite its capabilities, lacks an extensive common sense knowledge base. Additionally, CLIP encounters challenges due to its reliance on image-text paired data, which inherently suffers from the long-tail distribution issue,  limiting the model's ability to less common or more nuanced scenarios.

To address these challenges, we developed \textbf{ProLab}, an innovative approach that creates a \textbf{pro}perty-level \textbf{lab}el space for semantic segmentation. This label space, derived from the rich common sense knowledge base of Large Language Models (LLMs), is filled with descriptive properties. It aligns segmentation models to a nuanced and human-interpretable common sense semantic space, leading to stronger performance and better generalization ability, shown in \cref{fig:teaser}(a) and \cref{fig:teaser}(b). Moreover, ProLab models interpretable semantic correlations between categories and is scalable and adaptable to expanding data volumes.

ProLab enables models to recognize objects based on a set of interpretable properties. For instance, our model can generate pixel-wise logits for properties like ``paws have claws and pads for walking" or ``round eyes." For general segmentation benchmarks evaluation, our method compares these property-level logits to the original label space which can be somehow in line with the human-reasoning process~\cite{davis2015commonsense,knowlton1993learning}. It might, for example, assign the category ``dog" based on activated properties like ``paws have claws and pads for walking" and ``seen in parks", showcasing its ability to align property-based recognition with conventional category-level identification.

\noindent Our contribution can be summarized as follows: 

\begin{itemize}
        \item  We propose a novel method, \textbf{ProLab}, building an interpretable label space for semantic segmentation by retrieving common sense knowledge from pure language models instead of by using CLIP text encoders.
        \item \textbf{ProLab} consistently shows stronger performance than classic category-level supervision on five benchmarks: ADE20K~\cite{zhou2017ade}, COCO-Stuff~\cite{lin2014coco}, Pascal Context~\cite{lin2014coco},  Cityscapes~\cite{cordts2016cityscapes}, and BDD~\cite{yu2020bdd100k}.
        \item \textbf{ProLab} shows better scalability with extended training steps without having performance saturation.
        \item \textbf{ProLab} qualitatively exhibits strong generalization capabilities to segment out-of-domain categories with in-domain descriptive properties.
\end{itemize}

%% file: 02_related.tex
\section{Related Work}
\label{sec:related}

\subsection{Open-vocabulary recognition}
Open-vocabulary recognition aims to address visual recognition problems in an open world by extending the semantic space to unlimited vocabularies. To address this problem, a universal trend is to leverage pre-trained vision-language models such as CLIP~\cite{radford2021clip}, where the language modules are trained to be visually aligned such that they can be mapped to open vocabularies.

Recent works such as~\cite{gu2021open} address open-vocabulary object detection and subsequent works extend the problem to various segmentation tasks with more or less similar approaches~\cite{li2022language,ghiasi2022scaling,ding2023maskclip,chen2023open,zou2023generalized,zhang2023simple,xu2023odise,chen2023semantic,li2023semantic}. A critical difference between prior works and this paper is that our method focuses on the construction of semantic space using LLM knowledge instead of vision-language pre-training.

\subsection{Language-supervised image segmentation}
Besides open-vocabulary recognition, recent works also consider language-supervised dense prediction without using mask annotations~\cite{zhou2022maskclip,xu2022groupvit,xu2022groupvit,mukhoti2023open}. While enjoying similar open-vocabulary recognition capabilities using vision-language models, these models further explore the emerging dense prediction properties from language supervision. 
It should be mentioned that these works are inherently related to earlier works on the emerging localization from network activation~\cite{zhou2016learning} and weakly-supervised learning~\cite{bilen2016weakly,durand2017wildcat}.

\subsection{Referring expression grounding}
Another important language model application for vision tasks is referring expression grounding. Referring expression grounding aims to locate a target region in an image according to the referring language expression. There exists a rich literature that focuses on the design of vision-language interaction modules~\cite{hu2016natural,hu2016segmentation,ye2019cross,kamath2021mdetr,yang2022lavt} with common grounding benchmarks such as RefCOCO~\cite{kazemzadeh2014referitgame} and PhraseCut~\cite{wu2020phrasecut}. More recent works feature the use of well pre-trained vision-language models~\cite{li2022grounded,liu2023grounding} and LLMs~\cite{peng2023kosmos,wang2023visionllm,lai2023lisa,bai2023qwen,chen2023minigpt,xiao2024palm2vadapter}.

Our work is partly related to referring expression grounding in the sense that segmentation with interpretable properties can be broadly viewed as an inverse process going from regions to language expression. With moderate changes, it is possible to also retrieve these regions with free-form language similar to referring expression grounding.

\subsection{Semantic space construction}

The use of language knowledge for semantic space construction is possibly the most relevant area to this work. This area is rooted from many seminal early works in visual recognition, including but are not limited to the following:

\paragraph{Hierarchy and graph.} Semantic hierarchy~\cite{dekel2004large,marszalek2007semantic,amit2007uncovering,zweig2007exploiting,fergus2010semantic,li2022deep} and relational/knowledge graph~\cite{deng2014large,wang2018zero,zhu2021semantic} has a long history~\cite{tousch2012semantic} for large-scale visual recognition~\cite{deng2009imagenet}. In these cases, one could formulate the recognition task as a structured prediction task, or leverage the hierarchy and graph as structured priors besides likelihood. Hierarchy and relational/knowledge graph are essentially special cases of the LLM encoded knowledge.

\paragraph{Attribute learning.} Learning with attributes~\cite{farhadi2010attribute,sharmanska2012augmented,russakovsky2012attribute,akata2013label,yu2013designing} is a task in which one focuses on learning the properties or ``visual adjectives'' beyond just taxonomy. A direct application of attribute learning is zero-shot recognition~\cite{palatucci2009zero,farhadi2010attribute,yu2013designing} and our method can be broadly viewed as a form of attribute learning.

\paragraph{Multi-dataset training.} Constructing aligned label space is an important step for multi-dataset training since different dataset may use different vocabularies to represent the same visual concept~\cite{lambert2020mseg, xiao2022rvc, kim2022learning,zhou2023lmseg}. Lambert et al.~\cite{lambert2020mseg} show that aligning the label space is key to the domain generalization capability of trained models. Kim et al.~\cite{kim2022learning} proposes a label unification approach to systematically resolve the label definition conflicts, while recent works also resort to language models to automate this process~\cite{zhou2023lmseg}. Our method can be used for similar purposes using LLM.

%% file: 03_method.tex
\section{Method}
\label{sec:method}

\begin{figure*}[!t]
\centering
\includegraphics[width=\textwidth]{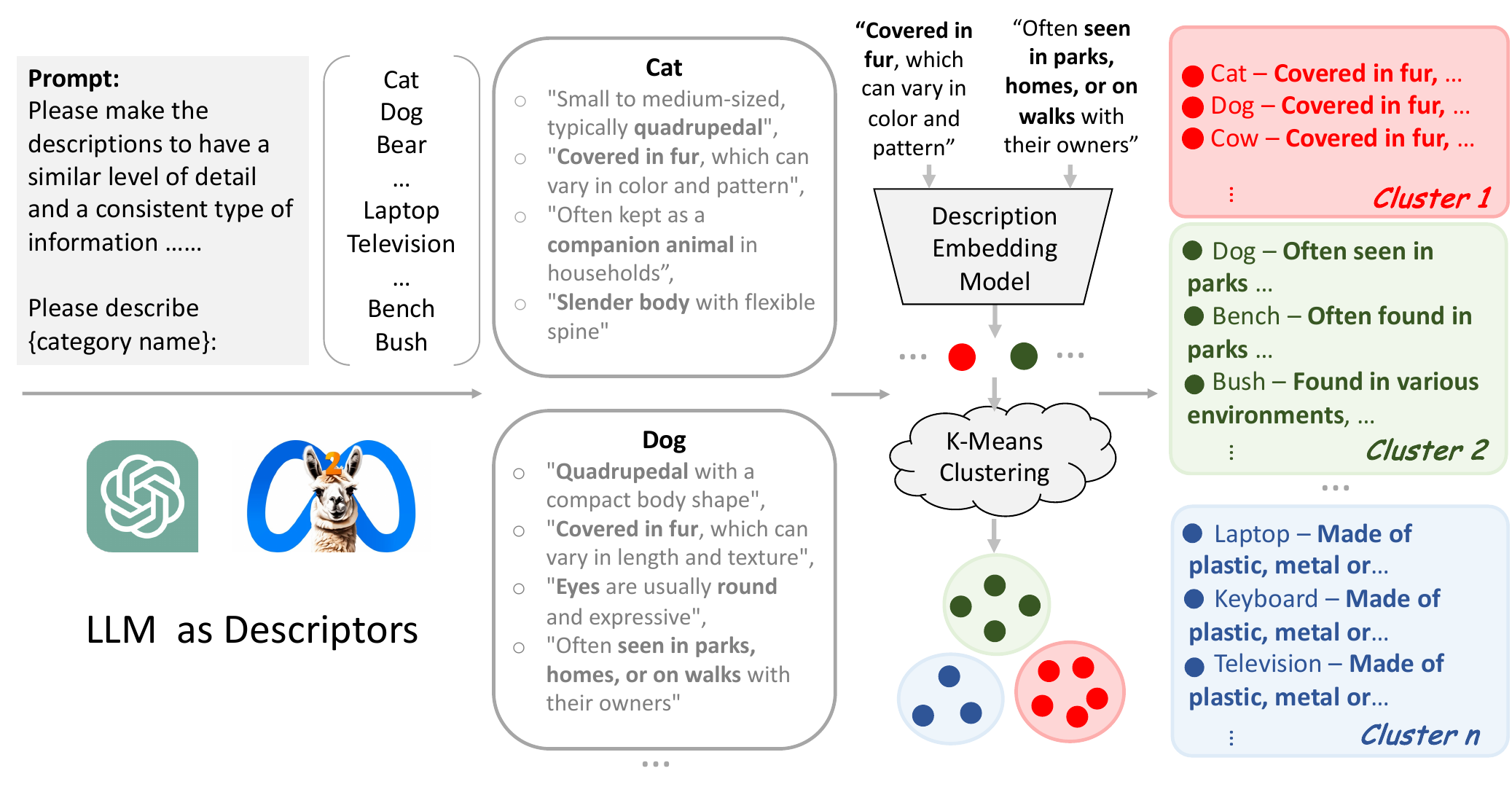}

\caption{\textbf{Build a semantic space of descriptive properties.} ProLab firstly employs a Large Language Model to extract common sense knowledge pertinent to all involved categories, utilizing crafted prompts to ensure a structured format. Subsequently, a description embedding model is used to encode these descriptions, preserving semantic correlations. 
Finally, the description embeddings are grouped into a series of unique descriptive properties through K-Means clustering.
}

\label{fig:method_llm_as_descriptors}
\end{figure*}

\def\cR{{\cal R}}

Conventionally, a semantic segmentation model $f$  process an RGB image $x\in {\cal R}^{3\times H\times W}$ as input, generating pixel-wise predictions $p = f(x)\in {\cal R}^{N\times H\times W}$, where $N$ signifies the number of categories in line with the label space $\{C_1, ... C_N\}$ of the designated training dataset(s). This is represented as:

$$
y \in \mathbbm{1}_{C_i}(x) :=
\begin{cases}
1 &\text{if } x \in C_i, \\
0 &\text{if } x \notin C_i.
\end{cases},\quad i \in \{1, 2, 3, \ldots, N\}
$$

However, this traditional one-hot label space fails to capture inter-class correlations, resulting in models lacking out-of-domain generalization ability. Our approach, in contrast, employs LLMs (e.g., GPT-3.5) to transform this one-hot category-level label space into a multi-hot property-level label space for supervision. Initially, LLMs function as descriptors to provide a set of descriptions regarding the properties of each distinct category (as detailed in \S\ref{sec:method_LLM_as_desc}). These descriptions are encoded into embeddings by a sentence embedding model and subsequently clustered into a series of interpretable properties $\{P_1, P_2, P_3 ... P_M\}$ (as detailed in \S\ref{sec:method_build_space}).  This is represented as:

$$
y \in \mathbb{1}_{P_j}(x):=\left\{\begin{array}{lll}
1 & \text { if } x \in C_i, C_i \in P_j & \:\:\quad i \in\{1,2,3, \ldots, N\}, \\
0 & \text { if } x \notin C_i & ,\quad j \in\{1,2,3, \ldots, M\}
\end{array}\right.
$$

\subsection{Property Knowledge Retrieval from LLM}
\label{sec:method_LLM_as_desc}

\paragraph{Large language models.} Our approach leverages LLMs as descriptive tools, illusrated in \Cref{fig:method_llm_as_descriptors}. They offer a spectrum of detailed descriptions related to the characteristics of various categories within a label space. These models, particularly state-of-the-art ones like GPT-3.5, are adept at generating rich, meaningful descriptions at the property level. These descriptions not only resonate with human understanding but also serve as benchmarks for distinguishing between different categories. 
The capacity of Large Language Models (LLMs) to express common sense knowledge about shape, texture, and other general properties, akin to human recognition processes, renders them a vital element in our approach. This ability significantly augments the interpretability and practicality of label spaces in our method.

    \begin{figure}[t]
        \centering

        \includegraphics[width=\linewidth]{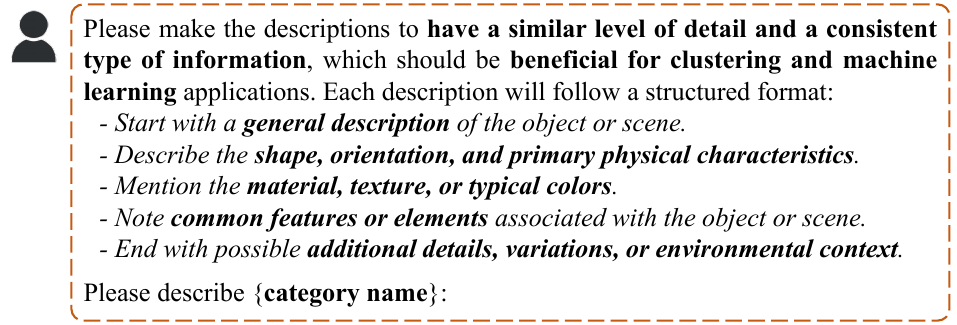}
        \caption{\textbf{Prompt for descriptions.} This crafted-prompt ensures precise guidance for Large Language Models in retrieving consistent and reliable property descriptions.}
        \label{fig:Prompt}
    \end{figure}

\paragraph{Prompt.}
Crafting an effective prompt is essential for getting high-quality, detailed descriptions from Large Language Models (LLMs), which are instrumental in reconstructing the semantic space with rich property-level information. As illustrated in \Cref{fig:Prompt}, we deploy precise instructions to guide the LLM in characterizing categories with uniform detail and consistent information types. 

This approach clusters properties into groups with meaningful semantics. The LLM provides descriptions across a broad spectrum of attributes, including shape, orientation, and primary physical features like material, texture, and characteristic colors. It also covers common elements associated with the object or scene. This structured prompting generates comprehensive descriptions that enhance the quality and usability of the semantic space.

\subsection{Build Semantic Space with Descriptions}
\label{sec:method_build_space}

\paragraph{Description embedding model.} Our approach employs sentence embedding models, such as Sentence Transformers~\cite{reimers2019sentence} or BGE-Sentence~\cite{bge_embedding}, to transform descriptions into a representational space suitable for supervision. These models, purely based on language, are pre-trained exclusively on textual data. Their training involves the use of contrastive loss, which effectively constructs an embedding space capable of modeling semantic similarities between sentences. These similarities are quantified using cosine similarity scores, allowing the embedding space to accurately preserve semantic correlations among various property-level descriptions. This methodology ensures that the essence and nuanced differences of each description are effectively captured and represented in the model, facilitating a more robust and semantically rich supervision process.

\paragraph{Cluster description embeddings.}
\label{sec:method_clustering}
Given that many properties are shared across multiple categories, it becomes necessary to group descriptions into clusters when they refer to identical or similar properties. To achieve this, we employ K-Means clustering~\cite{lloyd1982least} to cluster  the embeddings generated by the description embedding model into a set of generalized properties. This clustering is performed while maintaining a controlled semantic distance between the properties, ensuring that they remain interpretable and relevant to humans. By aggregating similar descriptions, we streamline the semantic space, making it more practical and user-friendly for both computational processes and human understanding.

\begin{figure*}[!t]
\centering
\includegraphics[width=\linewidth]{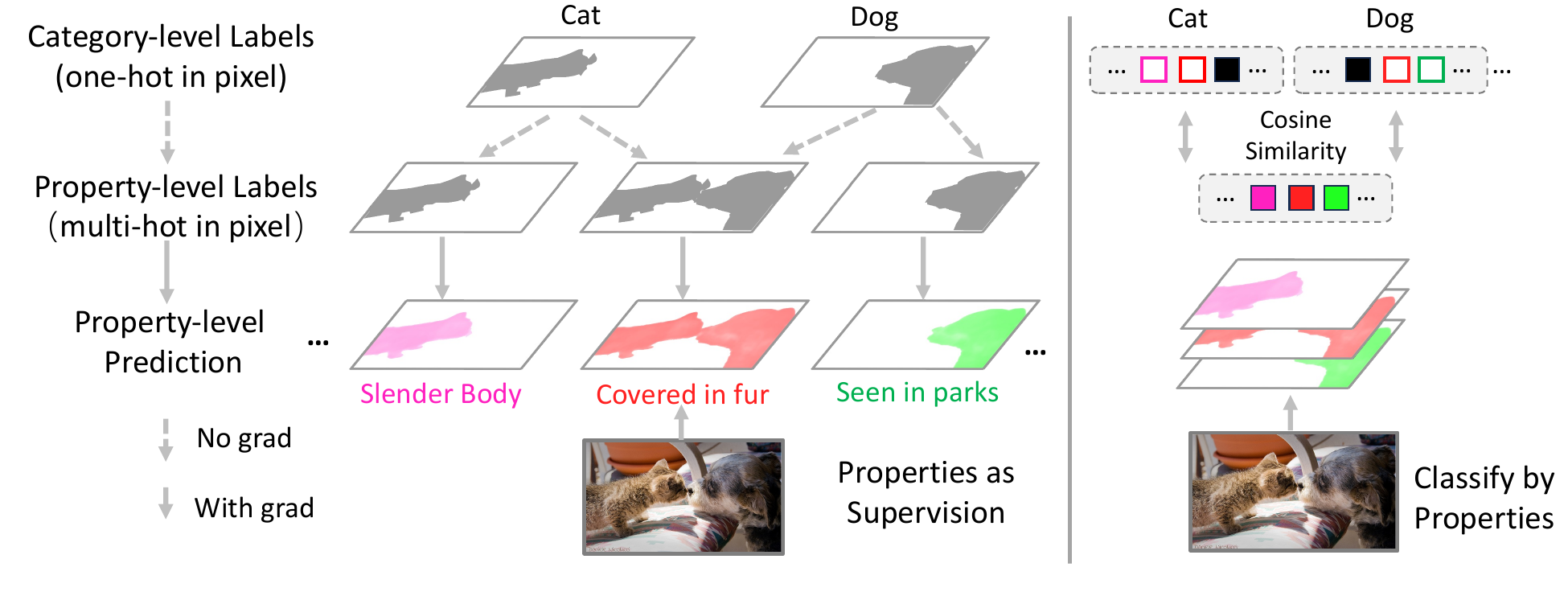}
\caption{\textbf{Supervise and classify with properties}. \textbf{Left}: the training procedure where descriptive properties are used for model supervision. \textbf{Right}: the inference procedure of categorizing items within the original category-level label space.
}
\label{fig:descriptions_as_supervisions}
\end{figure*}

\subsection{Supervise and Classify with Properties}

\paragraph{Properties as supervision.} 
In contrast to the category-level label space that a pixel is only labeled with one category, this property-level label space is a multi-label classification problem. As shown in \Cref{fig:descriptions_as_supervisions}, the pixels which are originally labeled as ``cat", now correspond to all the clustered properties of the ``cat" descriptions, including ``slender body" and ``covered in fur". And for the pixels labeled with ``dog", they are now labeled with all the dog's properties including ``covered with fur" and "seen or found in various environments". For each pixel \(i\), our model predicts an embedding
$\mathbf{e}_i$ at the pixel. Let \( \mathbf{E} \in \mathbb{R}^{d \times k} \) be the property embedding bank. Then, the property-level logits for pixel \(i\) can be calculated:
$\mathbf{z}_i = \sigma(\mathbf{e}_i \mathbf{E})$, where \(\sigma(\cdot)\) denotes the sigmoid function. Let \( \mathbf{y}_j \) be the multi-hot label vector for the class \(j\) of the label space. The cosine similarity is calculated as:
$\text{sim}(\mathbf{y}_j, \mathbf{z}_i) = \frac{\mathbf{y}_j \cdot \mathbf{z}_i}{\|\mathbf{y}_j\| \|\mathbf{z}_i\|}$. With this multi-hot property-level label space, our model has stronger performance, better scalability, and emerges generalization ability to out-of-domain and even unknown categories.

\paragraph{Classify by properties.}
Our model is still adept at categorizing pixels into their original categories by leveraging property-level logits, as depicted in \Cref{fig:descriptions_as_supervisions}. Specifically, this is achieved by assigning each pixel to the category whose property-level label exhibits the highest cosine similarity with that pixel:
$c_i = \arg\max_{j} \left( \text{sim}(\mathbf{y}_{j}, \mathbf{z}_{i}) \right)$.  This method effectively harnesses the nuanced information contained within the property-level descriptions, ensuring that pixel categorization aligns accurately with the most relevant semantic characteristics.

%% file: 04_experiment.tex
\section{Experiments}
\label{sec:exp:main}

\subsection{Implementation Details}

\paragraph{Segmenatation model.}
We use ViT-Adapter~\cite{chen2023vision} with UperNet~\cite{xiao2018unified} as the segmentation framework which is a state-of-the-art method for the semantic segmentation task. Unless otherwise specified, ViT-Base serves as the standard vision backbone across all experiments. The output feature dimension is determined by the dimension of language embeddings, which is set to 384 or 768, which matches the dimension of description embeddings.

\paragraph{Large Language Models.}
In our experiments, we employ GPT-3.5~\cite{brown2020gpt3} and LLAMA2-7B~\cite{touvron2023llama} as our primary large language models. By default, we utilize GPT-3.5 for descriptive properties retrieval. To ensure the extraction of descriptive properties at a consistent level for different categories, we craft a prompt to accurately guide the LLMs in retrieving relevant and uniform property description, as detailed in \cref{fig:Prompt}.

\paragraph{Embedding model.} For the language embedding model, we choose Sentence Transformers ~\cite{reimers2019sentence} and BGE-Sentence~\cite{bge_embedding}. The output feature dimension is set to 384 or 768. All pretrained weights are imported from HuggineFace~\cite{wolf2019huggingface}.

\paragraph{Training details.}
The ViT backbone in our model is initialized with the DeiT-Base weights \cite{touvron2021deit}, which have been pre-trained on ImageNet-1K \cite{deng2009imagenet}. For ADE, COCO-Stuff, and Pascal Context, the input resolution is set to 512x512, while for Cityscapes and BDD, a resolution of 768x768 is used. The optimizer of choice is AdamW, with LR configured at 6e-5. Our models are trained on 8-GPU machines, with a total batch size of 16. For supervising the property-level labels, cosine similarity loss is utilized. A one-hot training stage is adopted for warm-up when training large models. Further details are provided in the appendix.

\paragraph{Evaluation details.}
The models are evaluated using the classic single-scale test setting. The primary metric is the Mean Intersection over Union (mIoU). For classic evaluation, a dictionary of multi-hot property-level labels is employed to calculate cosine similarity scores with the property-level logits. The category with the highest cosine similarity score is determined as the predicted label. Further details are provided in the paper's appendix.

\begin{figure}[t]
    \centering
    \includegraphics[width=\linewidth]{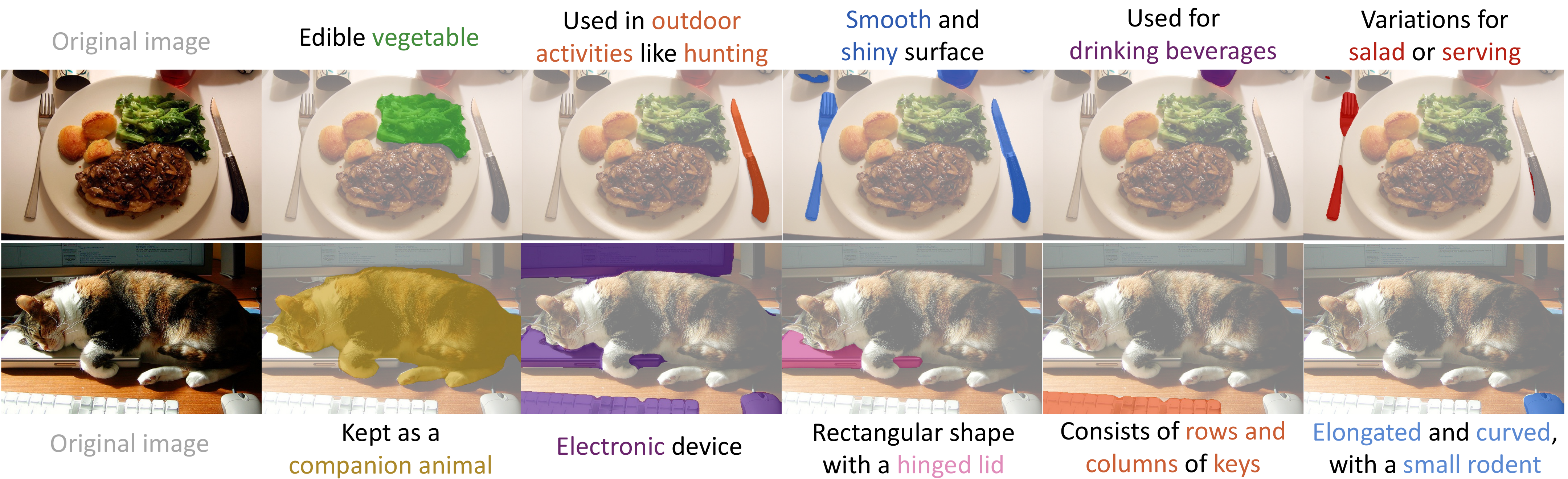}

    \caption{\textbf{Segmentation with interpretable properties}.  Our ProLab model enables property-level segmentation using descriptive prompts, enhancing interpretability and mirroring human-like understanding.}
    \label{fig:interpretable_prop}
\end{figure}

\subsection{Segmentation with Interpretable Properties}
\label{sec:exp_interpretable}

Our method utilizes property-level supervision to generate activation maps reflecting distinct descriptive properties, which are then correlated with traditional category-level labels based on property-level logits similarity. \Cref{fig:interpretable_prop} displays the model's predictions related to different properties in two example images. Each row shows activated areas corresponding to three distinct interpretable properties. In 1st row, ``Broccoli'' is identified through the ``edible vegetable'' property. However, ``knife''and ``fork'' require a nuanced approach for categorization. They are grouped under the ``smooth and shiny surface'' property. The ``knife'' can be distinguished by its association with ``activities like hunting.'' Consequently, our model segments areas that align with both properties as ``knife'', while employing additional properties to isolate the region representing ``fork''.

The second row demonstrates the model's ability to segment by diverse interpretable properties like ``rectangular shape with a hinged lid,'' ``companion animal,'' and ``electronic device.'' Here, items such as keyboards, mice, laptops, and monitors are segmented under the ``electronic device'' property, which aligns with human cognition. This approach underlines the effectiveness of property-level segmentation for more accurate and understandable categorization, mirroring human-like decision-making in identifying objects.

\subsection{Stronger Performance \& Better Scalability}

\paragraph{Stronger performance.}
\label{sec:exp_stronger_models}
To comprehensively evaluate our method, We conduct extensive experiments on five classic semantic segmentation datasets: three natural scene datasets (ADE20K~\cite{zhou2017ade}, COCO-Stuff~\cite{lin2014coco}, Pascal Context~\cite{lin2014coco}), and two self-driving datasets (Cityscapes~\cite{cordts2016cityscapes}, BDD~\cite{yu2020bdd100k}). We utilized ViT-Adapter, a state-of-the-art segmentation framework, as our baseline to evaluate the efficacy of our property-level label space.  As detailed in \cref{tab:combined_table}, our method consistently outperformed the baseline across all five datasets, often by a substantial margin. Notably, it achieved significant enhancements in the Pascal Context and BDD datasets, with an increase to 58.2 (\textcolor{Green}{+4.9}) mIoU and 65.7 (\textcolor{Green}{+5.0}) mIoU.

We also make improvements on remaining datasets (ADE20K, COCO-Stuff, and Cityscapes). These results validate the effectiveness of our approach, demonstrating that our label space, constructed with descriptive properties, builds a better representation space. This space adeptly encodes semantic correlations across different categories, proving its superiority over traditional methods.

\begin{table}[t]
\centering
\tablestyle{1pt}{1.5}
\scriptsize
\begin{tabular}{lcc|ccc|cc}
\multicolumn{3}{c|}{} & \multicolumn{3}{c|}{\textbf{Natural Scene }} & \multicolumn{2}{c}{\textbf{Self Driving }} \\
\hline
Label Space & Backbone & Framework & ADE20K & COCO-S. & Pascal Ctx.  & Cityscapes & BDD \\
\shline
Categories & DeiT-B & ViT-Adapter & 48.4 & 43.1 & 53.3 & 79.9 & 60.7 \\
Properties & DeiT-B & ViT-Adapter & \textbf{49.0}\textbf{\textcolor{Green}{(+0.6)}} & \textbf{45.4}\textbf{\textcolor{Green}{(+2.3)}} & \textbf{58.2}\textbf{\textcolor{Green}{(+4.9)}} & \textbf{81.4}\textbf{\textcolor{Green}{(+1.5)}} & \textbf{65.7}\textbf{\textcolor{Green}{(+5.0)}} \\
\end{tabular}
\caption{\textbf{Categories v.s. Properties}. We adopt the state-of-the-art segmentation method ViT-Adapter~\cite{chen2023vision,xiao2018unified} as our baseline.Our ProLab consistently shows stronger performance on three natural scene datasets and two self-driving datasets.}
\label{tab:combined_table}
\end{table}

\begin{table}[t]
\begin{minipage}{0.5\linewidth}
\centering
 \includegraphics[width=\linewidth]{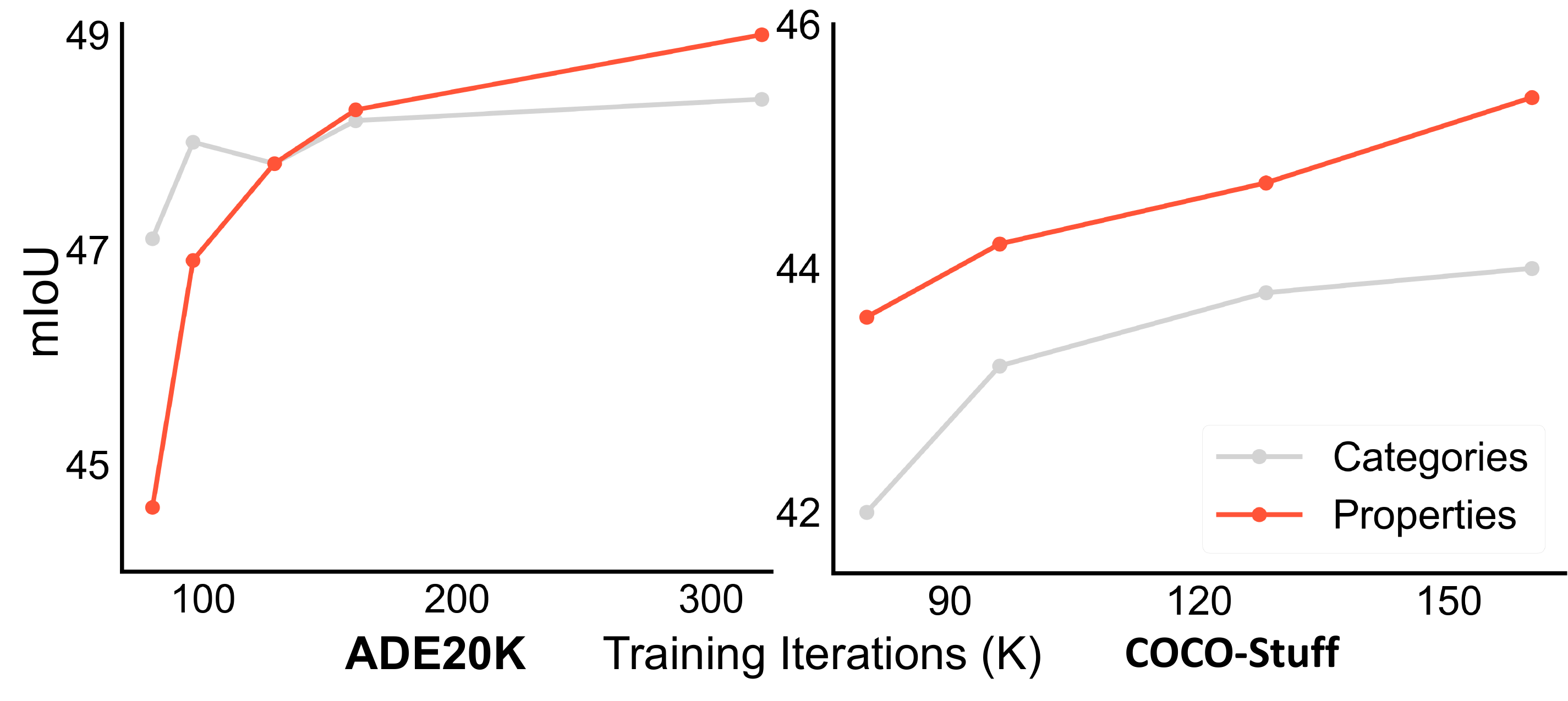}
        \captionof{figure}{\textbf{Better scalability.} The demonstrated results validate that our property-level semantic space effectively reduces the tendency of models to overfit, even with extended training durations.}
    \label{fig:exp_scale}
    \end{minipage}
\begin{minipage}{0.48\linewidth}
\tablestyle{1.0pt}{1.8}
\scriptsize
\begin{tabular}{cc|ccc}
Method     & Backbone                & ADE20K              & Citysc.          & BDD                 \\ \shline
DeepLabv3+ & \multicolumn{1}{c|}{ResNet50} & 42.7       & 79.5                & 63.6                \\
\rowcolor{gray!15}\multicolumn{2}{c|}{\textbf{w/ ProLab}}         & \textbf{43.6} & \textbf{79.9} & \textbf{64.7} \\ \hline
Segformer  & MIT-B1                  & 41.5                & 78.8                & 61.7                \\
\rowcolor{gray!15}\multicolumn{2}{c|}{\textbf{w/ ProLab}}         & \textbf{42.3} & \textbf{79.9} & \textbf{62.1}
\end{tabular}
\caption{\textbf{Generalizable to other methods.} ProLab demonstrates generalizable effectiveness by making consistent improvements segmentation frameworks (\ie, DeepLabv3+ \cite{chen2017deeplab}, SegFormer \cite{xie2021segformer}). }
\label{tab:generalizable_to_other}
\end{minipage}
\end{table}
    
\paragraph{Better scalability.}
\label{sec:exp_scaling}
Our proposed property-label based method demonstrates significant potential for scalability with extended training steps. 
The evidence supporting this is illustrated in \cref{fig:exp_scale}, where we observe performance improvements in our model (depicted by the red lines) that surpass those of the baseline model (represented by the gray lines) when the models trained. This trend is evident on both the ADE20K and COCO-Stuff datasets.
\begin{table}[t]
	\centering
	\renewcommand\arraystretch{1.0}
    \tablestyle{1pt}{1.0}
    \begin{tabular}{*l^c^c^H^c^H^c^H^H}
        Method & Framework & Backbone Pre-train & Extra Pre-train & Crop Size & Iters & ADE20K & \#Param +MS \\
        \shline
        Swin-L~\cite{liu2021swin} & Mask2Former & IN-22K, sup & - & 640 & 160k & 56.1 & 57.3 & 215M\\
        Swin-L-FaPN~\cite{huang2021fapn} & Mask2Former & IN-22K, sup & -& 640 & 160k & 56.4 & 57.7 & 217M\\
        SeMask-Swin-L\cite{jain2021semask} & Mask2Former & IN-22K, sup & -& 640 & 160k & 57.0 & 58.2 & -\\
        HorNet-L~\cite{rao2022hornet} & Mask2Former & IN-22K, sup & -& 640 & 160k & 57.5 & 57.9 & -\\

        ViT-Adapter-L~\cite{chen2023vision} & Mask2Former & IN-22K, sup & - & 640 & 160k & 56.8 & 57.7 & 438M \\
        \hline
	    BEiT-L~\cite{beit} & UperNet & IN-22K, BEiT & - & 640 & 160k & 56.7 & 57.0 & 441M\\
	
	    ViT-Adapter-L~\cite{chen2023vision} &
	    UperNet & IN-22K, BEiT & - & 640 & 160k & 58.0 & 58.4 & 451M\\
	    BEiTv2-L~\cite{beitv2} & UperNet & IN-22K, BEiTv2 & - & 512 & 160k & 57.5 & 58.0 & 441M\\
	     
	    ViT-Adapter-L~\cite{chen2023vision} &
	    UperNet & IN-22K, BEiTv2 & - & 512 & 160k & 58.0 & 58.5  & 451M\\
    \rowcolor{green!12} 
    ProLab (Ours) & UperNet & IN-22K, BEiT & - & 640 & 160k & 58.2 & - & - \\
    \rowcolor{green!12} 
    ProLab (Ours) & UperNet & IN-22K, BEiTv2 & - & 896 & 160k & \textbf{58.7} & - & - \\
	   \rowstyle{\color{hr}}SwinV2-G~\cite{liu2022swin} & UperNet & IN-22K, sup & Ext-70M, sup & 896 & 160k & 59.3 & 59.9 & 3.0B \\
	
    \end{tabular}
    \caption{\textbf{Comparison with state-of-the-art methods} on the ADE20K \textit{val} set. ProLab achieves state-of-the-art performance. We report \textbf{mIoU} with single-scale inference.  SwinV2-G~\cite{liu2022swin} is \textcolor{gray}{grayed} as its backbone is in a much larger scale (\ie, 3B). }
    \label{tab:sota_ade}
\end{table}

\subsection{Generalizablity to Other Segmentation Models}

To evaluate the generalizability of our property-level label space,  we arm two other classic segmentation methods (\ie, DeepLabv3+\cite{chen2017deeplab} and Segformer\cite{xie2021segformer}) with ProLab. As detailed in \Cref{tab:generalizable_to_other}, ProLab consistently improves the performance with both DeepLabv3+ and Segformer on three classic datasets (\ie, ADE, Cityscapes and BDD), demonstrating strong generalizability with ProLab.

\subsection{Comparison with State-of-the-Art Methods}
To validate the versatility of our approach across different backbone architectures, especially those with advanced pretraining, we evaluated our method using larger backbones pretrained with state-of-the-art methods \cite{beit, beitv2}. As indicated in \Cref{tab:sota_ade}, our approach enhances the performance of the ViT-Adapter-L model to achieve a new state-of-the-art performance of 58.2 mIoU on the ADE20K validation set using BeiT-Large, which further increases to 58.7 mIoU when using a higher input resolution of 896 with BeiTv2. It's worth noting that while SwinV2-G~\cite{liu2022swin}, implemented in the UperNet framework, achieves slightly higher performance, its model size is approximately five times larger than ours.

\begin{table}[t]
\centering
\subfloat[\textbf{Description embedding model.}\label{tab:choice_of_LM}]{
\tablestyle{1pt}{2}
\scriptsize
\begin{tabular}{ccc}
Model. & Embed. Len. & mIoU \\
\shline
Sent. TR-Small & 384 & 47.7 \\
Sent. TR-Base & 768 &  47.8\\
BGE-Small & 384 & 47.9\\

\rowcolor{gray!15} BGE-Base & 768 & \textbf{48.3} \\
\end{tabular}
}
\tablestyle{1pt}{1.6}
\scriptsize
\subfloat[\textbf{Description cluster number.}\label{tab:size_clip_models}]{
\begin{tabular}{ccc}
\# Cluster & mIoU \\
\shline
None & 30.2\\
64 & 47.8\\
128 & 48.0\\
\rowcolor{gray!15} 256 & \textbf{48.3}\\
512 & 47.6\\
\end{tabular}
}
\subfloat[\textbf{Prompts with text encoders.}\label{tab:prompts_for_text_encoders}]{
\tablestyle{1.0pt}{1.32}
\scriptsize
\begin{tabular}{lccc}
\multirow{2}{*}{Prompts} & Text & Embed. & \multirow{2}{*}{mIoU} \\
& Encoder & Space& \\ 
\shline
CLIP Official  & CLIP-B & Category & 48.6 \\
Ours Descr. & CLIP-B & Category & 48.6 \\
\hline
Ours Descr. & T5-B & Category &  42.0 \\
Ours Descr. & BERT-B & Category & 40.8 \\
\hline
Ours Descr. & BGE-B & Category &  47.7\\
\rowcolor{gray!15} Ours Descr. & BGE-B & Property
& \textbf{49.0} \\
\end{tabular}
}

\subfloat[\textbf{Description logit temperature. } \label{tab:logit_temp}]{
\tablestyle{1.2pt}{1.3}
\scriptsize
\begin{tabular}{cc}
Logit Temp. & mIoU \\ \shline
0.02      & 47.4 \\
\rowcolor{gray!15}
0.04                  & \textbf{47.7} \\
0.07         &  47.2\\  
\end{tabular}
}
\hfill
\subfloat[\textbf{Loss function for multi-hot labels. }
\label{tab:loss_function}]{
\tablestyle{1.0pt}{1.3}
\scriptsize
\begin{tabular}{*c^c}
Loss Function     & mIoU \\ \shline
Binary Cross Entropy     & 47.4       \\
Cosine Simi. (w/o Sigmoid) & 47.3   \\   
\rowcolor{gray!15} Cosine Simi. (w/. Sigmoid) & \textbf{47.7}   \\   
\end{tabular}
}
\hfill
\subfloat[\textbf{Description format.}\label{tab:prompt_description}]{
\tablestyle{1.0pt}{1.3}
\scriptsize
\begin{tabular}{*c^c^c}
LLM & Desc. Format   & mIoU \\
\shline
LLAMA2-7B & Naive~\cite{menon2022visual} & 47.0\\
GPT-3.5 & Naive~\cite{menon2022visual} & 47.5\\
\rowcolor{gray!15} GPT-3.5 & Aligned & \textbf{47.7} 
\end{tabular}
}
\caption{\textbf{Ablation studies} on the ADE20k dataset, mIoU scores are reported.  Best performance settings are marked in \colorbox{baselinecolor}{gray}. }
\end{table}

\subsection{Ablation Studies}
\label{sec:exp_ablation}

\paragraph{Description embedding model.} 
In our study, we explore two widely used language embedding models—Sentence TR~\cite{reimers2019sentence} and BGE~\cite{zhang2023retrieve}—to encode descriptive properties into a language semantic representation space. As detailed in \Cref{tab:choice_of_LM}, we experiment with two variants of these models, featuring embedding dimensions of 384 and 768. Our findings indicate a clear preference for BGE language embedding models, which seem to more effectively capture the essence of the sentences. Notably, larger embedding models outperform their smaller counterparts for both BGEs and Sentence TRs. Based on these results, we choose BGE as our language embedding model for subsequent experiments, aiming to leverage its enhanced performance capabilities.

\paragraph{Number of clusters.}
We also ablate the number of clusters for clustering embedding descriptions in \cref{tab:size_clip_models}.
Clustering has shown to be a critical component in our methodology, as evidenced by the markedly poor performance in the first row (without clustering).
Among different cluster numbers, models with 64 and 512 clusters show slight performance drops, hovering around 47.8 mIoU and 47.6 mIoU. The model using 128 clusters performs well, with only a small gap compared to the best cluster number of 256 (48.0 \vs 48.3).
These suggest that ProLab is not highly sensitive to cluster numbers, but they should be within a reasonable range. 
The optimal number of clusters can vary for different datasets. 1/6 to 1/8 of the number of descriptions usually works well based on our experiments in ~\cref{tab:combined_table}.

\paragraph{Prompts with text encoders.}
We conduct a comprehensive ablation study of the embedding space and text encoders. As outlined in \cref{tab:prompts_for_text_encoders}, we experiment with different kinds of language models: the CLIP text encoder, classic LMs (such as BERT and T5), and sentence embedding model (BGE). The first two rows of our experiment aim to determine if our prompt results in improvements when used with CLIP text encoders. The 3rd and 4th rows reveal that pure language models like BERT and T5 are less effective at encoding descriptions into embeddings compared to the CLIP text encoder and also underperform relative to sentence embedding models. This is possibly because general-purpose language models like BERT and T5 struggle to capture semantic relationships between sentences due to the absence of contrastive training. In contrast, the BGE sentence embedding model demonstrates reasonable performance when encoding descriptions into a category-level embedding space and achieves the best result of 49.0  when the embedding space is constructed at the property level.

\paragraph{Description logit temperature.} We explore the impact of sigmoid temperature on model performance, considering its role in rescaling the cosine similarity scores in property-level (ranging from -1 to 1) and smoothing gradients for better convergence. In our experiments detailed in \cref{tab:logit_temp}, we test three different temperatures: 0.02, 0.04, and 0.07. The results indicate that a temperature setting of 0.04 leads to the highest performance, achieving 47.6 mIoU. However, the temperatures of 0.02 and 0.07 also yield strong results (47.4 mIoU and 47.2 mIoU, respectively), showing only slight variations in performance. Based on these results, we use a temperature of 0.04 by default.

\paragraph{Loss function for multi-hot labels.}
We explore both binary cross-entropy loss and cosine similarity loss for our multi-hot property-level labels.
Interestingly, utilizing the cosine similarity loss allows the model to achieve comparable performance to binary cross-entropy loss (47.3 vs. 47.4).
Moreover, when we apply the sigmoid function to the output logits, the cosine similarity loss outperforms binary cross-entropy loss (47.6 vs. 47.4). 
This improvement is attributed to multi-hot labels being binary (0 or 1), making sigmoid-rescaled logits more suitable for model optimization than directly output logits.

\paragraph{Description format.} We evaluate the effectiveness of different description prompts with various Large Language Models (LLMs). Our baseline prompt, inspired by~\cite{menon2022visual}, poses a question to the LLM: \textit{``What are useful features for distinguishing a {category name} in a photo?''} As detailed in \cref{tab:prompt_description}, we experiment with GPT-3.5 and LLAMA2-7B for generating description prompts. The first two rows show that GPT-3.5 delivers common sense knowledge of higher quality compared to LLAMA2-7B. Using our specifically crafted prompt (termed as "Aligned") led to a further increase of 0.2 mIoU compared to the models trained with "Naive" prompts. These findings indicate that our tailored prompts are more effective at extracting property-level common sense knowledge.

%% file: 10_conclusion.tex
\section{Discussion}
\label{sec:discussion}

\subsection*{Emerged Generalization Ability}
\begin{figure}[t]
    \centering
    \includegraphics[width=\linewidth]{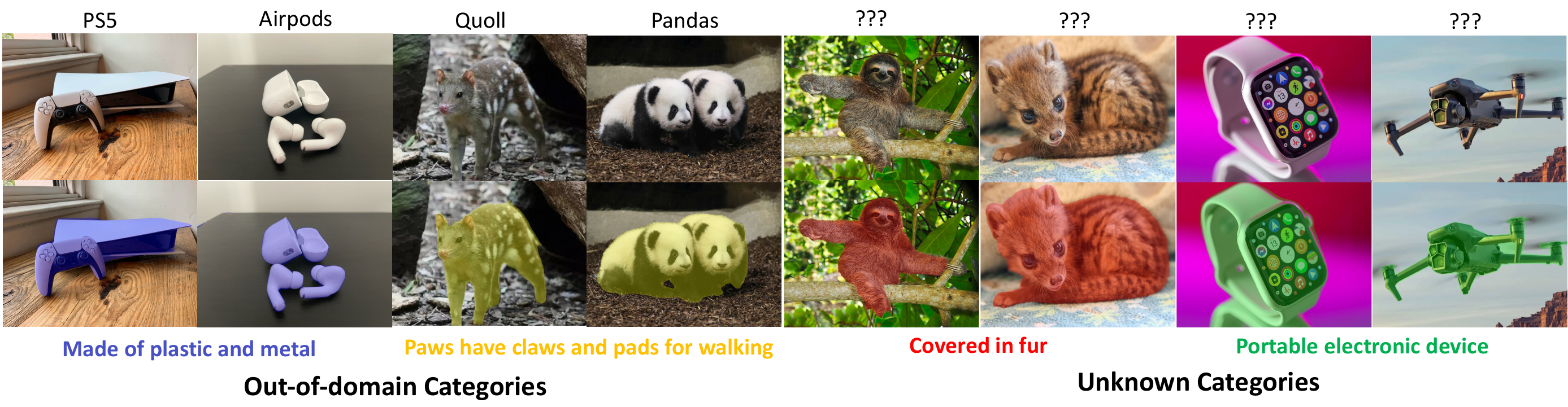}
\caption{\textbf{Emerged generalization ability.} Our model demonstrates the ability to segment categories that are outside its training domain, including those it has never encountered, as well as unknown categories, using significant descriptive properties. These properties draw on human-interpretable, common sense knowledge, showcasing the model's adaptability and depth of understanding.}
    \label{fig:emerged_generalization_ability}
\end{figure}

As highlighted in \Cref{sec:exp_interpretable}, our approach enables segmentation based on descriptive properties, such as shape, texture, and other common-sense knowledge. This capability leads to an emergent generalization ability, allowing our model to segment objects even if their categories are not included in the training set, by identifying their properties. \Cref{fig:emerged_generalization_ability} presents four examples: ``PS5'', ``Airpods'', ``Quoll'', and ``Pandas''. Despite these categories not being part of the training dataset, our model successfully segments them based on properties like ``made of plastic and metal'' and ``paws with claws and pads for walking''.

Additionally, in instances where category names are unknown, while humans can still segment objects using common sense properties, most deep learning models fail without specific category information. In contrast, our method demonstrates the ability to generalize to unknown categories using properties such as ``covered in fur'' and ``electronic device''. This somehow mirrors human reasoning processes, showcasing our model's advanced capability to recognize and segment objects beyond its trained categories.

\begin{table*}[t]
\centering
\tablestyle{3pt}{1.1}
\begin{tabular}{*l|^c|^l^c^c^c^c} %
Method                            & PT. Data (Scale) & Training Data    & A-847         & A-150         & PC-59         & PAS-20          \\
\shline
SPNet\cite{xian2019spnet}         & ImageNet (1.28M)  & Pascal VOC & -             & -             & 24.3          & 18.3            \\
ZS3Net\cite{bucher2019zs3net}     & ImageNet (1.28M)  & Pascal VOC & -             & -             & 19.4          & 38.3            \\
LSeg\cite{li2022language}         & ImageNet (1.28M)  & Pascal VOC & -             & -             & -             & 47.4            \\
ZegFormer\cite{ding2022zegformer} & CLIP-WIT (400M)  & COCO-Stuff       & 5.6 & 18.0 & 45.5 & 89.5            \\
LSeg+\cite{ghiasi2021open}        & ALIGN (1.8B) & COCO-Stuff       & 3.8           & 18.0          & 46.5          & -               \\

SimBaseline\cite{xu2021simple}    & CLIP-WIT (400M)  & COCO-Stuff & {7.0} & {20.5} & {47.7} & 88.4 \\

MaskCLIP\cite{ding2023maskclip}    & CLIP-WIT (400M)  & COCO-Stuff & 8.2 & 23.7 & 45.9 & - \\ 
ODISE\cite{xu2023odise}    & CLIP-WIT (400M)  & COCO-Stuff & \textbf{11.1} & \textbf{29.9} & \textbf{57.3} & - \\ 
\hline
\textbf{ProLab}                   & ImageNet (1.28M)  & COCO-Stuff 
& 2.1 & 15.6 & 44.9 & \textbf{90.6}\\
\rowstyle{\color{gray}}\textbf{ProLab$^{\dagger}$}       & ImageNet (1.28M)  & COCO-Stuff           & 3.5 & 23.1 & 57.7 & 92.5
\end{tabular}
\caption{
    \label{tab:open_voc_benchmark}
    \textbf{Open-vocabulary semantic segmentation.} 
  $\dagger$: Linear probing  with property-level logits (only 1 fully-connected layer is trainable). Without large-scale pre-training on image-text pairs for alignment, our method still shows strong open-vocabulary capability. Moreover, with a minimal linear projection (property-level  $\rightarrow$ class-level), ProLab could beat a lot of methods with a large margin.
}
\end{table*}

In \Cref{tab:open_voc_benchmark}, we quantitatively evaluate our method with the open vocabulary setting. Without pre-alignment on image-text pairs, our method still shows comparable performance on 4 benchmarks (\ie, ADE-847, ADE-150, Pascal Context, and Pascal VOC). Moreover, with simple linear probing on property-level logits, ProLab significantly outperforms competing methods by a wide margin, showcasing the efficacy of our property-level label space. This underscores ProLab's promising open-vocabulary capabilities and is worthy for further exploration.

\section{Conclusion}
\label{sec:conclusion}

The quest for an interpretable semantic space modeling inter-class semantic correlations has been a long-standing goal in computer vision. Previous methods, such as manual label merging~\cite{lambert2020mseg}, hierarchical label spaces~\cite{li2022deep}, or CLIP text encoders~\cite{zhou2023lmseg, li2022language}, have fallen short. Addressing this, we propose \textbf{ProLab}, a semantic segmentation method leveraging a \textbf{pro}perty-level \textbf{lab}el space, drawing from the common sense knowledge base of Large Language Models (LLMs), aligning with human reasoning to enhance interpretability and relevance.

Empirically, ProLab shows superior performance and scalability across classic benchmarks, adeptly segmenting out-of-domain and unknown categories based on in-domain descriptive properties. This paves the way for future research to improve segmentation models beyond traditional category-level supervision, aiming for a holistic understanding of scenes and objects that mirrors human perception.